\documentclass[letterpaper,11pt]{article}

\newcommand*{\affaddr}[1]{#1} 
\newcommand*{\affmark}[1][*]{\textsuperscript{#1}}

\usepackage[numbers]{natbib}
\usepackage{hyperref} 
\usepackage{amsmath}
\usepackage{amssymb} 
\usepackage{amsthm}
\usepackage{geometry} 
\usepackage{enumitem}
\usepackage{graphicx}
\usepackage{float}  
\usepackage{subcaption} 
\usepackage{color}  
\usepackage{algorithm}
\usepackage{algorithmic}
\usepackage[normalem]{ulem} 
\usepackage[export]{adjustbox}
\usepackage{setspace}
\usepackage[flushleft]{threeparttable}
\usepackage{booktabs}
\usepackage{multirow}
\usepackage{lscape}
\usepackage{tablefootnote}

\providecommand{\keywords}[1]
{
  \small	
  \textbf{\textit{Keywords---}} #1
}

\def\grad{\nabla}

\def\bc{\mathbf{c}}

\def\br{\mathbf{r}}

\def\bt{\mathbf{t}}

\def\bx{\mathbf{x}}  
\def\by{\mathbf{y}}
\def\bz{\mathbf{z}}

\def\bH{\mathbf{H}}
\def\bI{\mathbf{I}}
\def\bJ{\mathbf{J}}

\def\th{{\boldsymbol{\theta}}}

\def\b{{\boldsymbol{\beta}}}

\def\cC{\mathcal{C}}
\def\cD{\mathcal{D}}
\def\cE{\mathcal{E}}
\def\cF{\mathcal{F}}

\def\cH{\mathcal{H}}
\def\cI{\mathcal{I}}

\def\cL{\mathcal{L}}

\def\cN{\mathcal{N}}
\def\cO{\mathcal{O}}

\def\cQ{\mathcal{Q}}

\def\cX{\mathcal{X}}

\def\mR{\mathbb{R}}
\def\mS{\mathbb{S}}

\def\smskip{\smallskip}

\def\texitem#1{\par\smskip\noindent\hangindent 25pt
               \hbox to 25pt {\hss #1 ~}\ignorespaces}

\def\norm#1{\|#1\|}

\newcommand{\BEAS}{\begin{eqnarray*}}
\newcommand{\EEAS}{\end{eqnarray*}}
\newcommand{\BEA}{\begin{eqnarray}}
\newcommand{\EEA}{\end{eqnarray}}
\newcommand{\BEQ}{\begin{eqnarray}}
\newcommand{\EEQ}{\end{eqnarray}}
\newcommand{\BIT}{\begin{itemize}}
\newcommand{\EIT}{\end{itemize}}
\newcommand{\BNUM}{\begin{enumerate}}
\newcommand{\ENUM}{\end{enumerate}}

\newcommand{\BA}{\begin{array}}
\newcommand{\EA}{\end{array}}

\newcommand{\reals}{\mathbb{R}}
\newcommand{\integers}{\mathbb{Z}}

\newcommand{\Tr}{\mathop{\bf Tr}}

\def\blue#1{\textcolor{blue}{#1}}

\newif\ifpagenumbering
\pagenumberingtrue

\pagenumberingfalse

\newsavebox{\theorembox}
\newsavebox{\lemmabox}
\newsavebox{\defnbox}
\newsavebox{\assbox}
\savebox{\theorembox}{\noindent\bf Theorem}
\savebox{\lemmabox}{\noindent\bf Lemma}
\savebox{\defnbox}{\noindent\bf Definition}
\savebox{\assbox}{\noindent\bf Assumption}

\DeclareMathOperator*{\cov}{Cov}
\DeclareMathOperator*{\var}{Var}

\DeclareMathOperator*{\argmin}{argmin}

\DeclareMathOperator*{\sgn}{sgn}
\DeclareMathOperator*{\diag}{diag}

\def\fprod#1{\left\langle#1\right\rangle}
\def\prox#1{\mathbf{prox}_{#1}}
\newcommand{\mb}[1]{\mathbf{#1}}

\newtheorem{theorem}{Theorem}[section]
\newtheorem{corollary}{Corollary}[section]
\newtheorem{lemma}{Lemma}[section]
\newtheorem{remark}{Remark}
\newtheorem{definition}{Definition}[section]

\begin{document}
\date{}
\title{\vspace{-1cm} \Large\bfseries On the Theoretical Guarantees for Parameter Estimation of Gaussian Random Field Models: \\ A Sparse Precision Matrix Approach}

%

\author{%
Sam Davanloo Tajbakhsh\affmark[1]\thanks{davanloo.1@osu.edu}, Necdet Serhat Aybat\affmark[2]\thanks{nsa10@psu.edu}, Enrique Del Castillo\affmark[3]\thanks{exd13@psu.edu}\\
\\
\affaddr{\affmark[1]Department of Integrated Systems Engineering \\ The Ohio State University}\\
\\
\affaddr{\affmark[2]Department of Industrial and Manufacturing Engineering}\\
\affaddr{\affmark[3]Department of Industrial and Manufacturing Engineering and Department of Statistics \\ The Pennsylvania State University}\\
}

\maketitle

\begin{abstract}
Iterative methods for fitting a Gaussian Random Field (GRF) model via maximum likelihood (ML) estimation requires solving a nonconvex optimization problem. The problem is aggravated for anisotropic GRFs where the number of covariance function parameters increases with the dimension. Even evaluation of the likelihood function requires $\cO(n^3)$ floating point operations, where $n$ denotes the number of data locations.
In this paper, we propose a new two-stage procedure to estimate the parameters of second-order stationary GRFs. First, a \emph{convex} likelihood problem regularized with a weighted $\ell_1$-norm, utilizing the available distance information between observation locations, is solved to fit a sparse {\em precision} (inverse covariance) matrix to the observed data. Second, the parameters of the covariance function are estimated by solving a least squares problem. Theoretical error bounds for the solutions of stage I and II problems are provided, and their tightness are investigated.
\end{abstract}

\keywords{Convex Optimization, Gaussian Markov Random Fields, Kernel Methods, Hyperparameter Optimization, Covariance Selection, Spatial Statistics.}

\section{Introduction}\label{sec:intro}

Gaussian Random Field (GRF) models, also known as Gaussian Process (GP) models, are widely used in several fields, e.g., Machine Learning, Geostatistics, Computer Experiments (metamodeling) and Industrial Metrology. Traditional methods for fitting a GRF model to given
sample data rely on computing the maximum likelihood estimate (MLE) of the parameters of an assumed spatial covariance function belonging to a known parametric family. As it is
well-known in Spatial Statistics~\citep{WarnesRipley1987}, the log-likelihood function for the covariance parameters of a GRF is \emph{non-concave}, which leads to numerical difficulties in solving the optimization problem for MLE, yielding suboptimal estimates that do not possess the desirable properties of MLE.
Even though parametric GRF covariance matrices for \emph{isotropic} processes require estimation of a small number of parameters, finding the MLEs is challenging due to nonconvexity of the negative loglikelihood function. Furthermore, each evaluation of the negative loglikelihood function requires $O(n^3)$ operations due to covariance matrix inversions (where $n$ is the number of distinct data locations). Since $n$ is typically large in GRF modeling, the computational issues due to large $n$ is called the ``big $n$ problem"~\citep{BanerjeeEtal2008}. The problem is much more significant for \emph{anisotropic} processes where the number of parameters scales with the dimension.

To overcome these difficulties, 
we propose a new method, Sparse Precision matrix Selection (SPS), for fitting a GRF model {and 
establish the recovery guarantees and theoretical error bounds for the proposed estimator.} 
In the \emph{first stage} of the SPS method, a sparse precision (inverse covariance) matrix is fitted to the GRF data observations by solving a {\em convex} likelihood problem regularized with a weighted $\ell_1$-norm, utilizing the available 
distance information among observation locations. 
This precision matrix is not parameterized in any form and constitutes a Gaussian Markov Random Field (GMRF) approximation to the GRF. 
{The}  first-stage problem is solved using a {variant} of the Alternating Direction Method of Multipliers (ADMM) 
with a \emph{linear} convergence rate. Suppose the covariance function has $q$ parameters $(q\ll n)$. In the \emph{second stage}, 
these parameters are estimated 
via a least-squares~(LS) problem in $\reals^q$, resulting in more consistent estimates than 
the suboptimal solutions of the non-convex MLE problem. Although the second stage 
 LS problem is non-convex in general, it is still numerically much easier to solve when compared to the non-convex MLE problem. In particular, the solution to the second stage LS problem can be computed via a line search in the range parameter for isotropic GRFs. 
 Empirical evidence suggests that the first stage optimization ``zooms-in" to a region in the covariance parameter space that is close to the true parameter values, 
 alleviating the non-convexity to a certain degree. 
We next provide some preliminaries, including a brief review of
other state-of-the-art methods.

\subsection*{Preliminaries}\label{sec:prelim}
Let $\mathcal{X} \subseteq \mathbb{R}^d$ and 
$y(\mb{x})$
denote
the value of a latent GRF $f:\cX\rightarrow\reals$
observed with additive Gaussian noise at location $\mb{x}\in\cX$:
$y(\mb{x})=f(\mb{x})+\epsilon$,
where $f(\mb{x})$ has a mean function $m_f(\mb{x})$ and
covariance function $c_f(\mb{x},\mb{x}')=\mbox{cov}\left(f(\mb{x}),f(\mb{x}')\right)$ for all $\mb{x}, \mb{x}' \in\cX$, {and $\epsilon\sim \cN(0,{\theta^*_0})$  models the ``nugget" error, assumed independent of $f(\mb{x})$}. We  assume the training data $\mathcal{D} = \{(\mb{x}_i,y^{(r)}_{i}): i=1,...,n,\ r=1,...,N\}$ contains $N$ realizations of the GRF at each of $n$ distinct locations in $\mathcal{D}^x\triangleq\{\mb{x}_i\}_{i=1}^n\subset\cX$. Let $\mb{y}^{(r)}=[y_i^{(r)}]_{i=1}^n\in\reals^n$ denote the vector of $r$-th realization values for locations in $\cD^x$. Given a new location $\mb{x}_0\in\cX$, the goal in GRF modeling is to predict $f_0\triangleq f(\mb{x}_0)$. 
We assume that the GRF has a constant mean equal to zero, i.e., $m_f(\mb{x})=0$. Since any countable collection of observations from a GRF follows a multivariate normal distribution, the joint distribution of $(\mb{y}^\top,f_0)^\top$ is given by
$\textstyle \left( \begin{array}{c} \mb{y}^{(r)} \\ f_0 \end{array} \right) \sim \cN \left( \mb{0}_{n+1}, \begin{bmatrix} C_f+\theta^*_0\mb{I}_n & \mb{c}_0 \\ \mb{c}_0^\top & c_{00} \end{bmatrix} \right)$, 
for all $r=1,\ldots,N$, where $c_{00}=c_f(\mb{x}_0,\mb{x}_0)$, $\mb{c}_0^\top = \left[c_f(\mb{x}_1,\mb{x}_0),...,c_f(\mb{x}_n,\mb{x}_0)\right]$, and the covariance matrix $C_f\in\reals^{n\times n}$ is formed such that its $(i,j)^{th}$ element is equal to $c_f(\mb{x}_i,\mb{x}_j)$. Therefore, the conditional distribution of $f_0$ given $\{\mb{y}^{(r)}\}_{r=1}^N$, i.e., the predictive distribution of $f_0$ denoted by $p(\cdot~|~\{\mb{y}^{(r)}\}_{r=1}^N)$, is given as \vspace*{-2mm}
\begin{small}
\begin{equation} \label{eq:predDist}
p(f_0~|~\{\mb{y}^{(r)}\}_{r=1}^N) = \cN\left( \mb{c}_0^\top(C_f+\theta^*_0\mb{I}_n)^{-1}\sum_{r=1}^N\mb{y}^{(r)}/N,\ c_{00}-\mb{c}_0^\top(C_f+\theta^*_0\mb{I}_n)^{-1}\mb{c}_0\right).\vspace*{-2mm}
\end{equation}
\end{small}

\noindent The mean of this predictive distribution is a point estimate (known as the {\em Kriging} estimate in Geostatistics) and its variance measures the uncertainty of this prediction.

It is clear from \eqref{eq:predDist} 
that the prediction performance 
can be made significantly robust by correctly estimating the unknown covariance function $c_f$, which is typically assumed to belong
to some parametric family 
$\{c_f(\mb{x},\mb{x}',\boldsymbol{\theta}_f):\ \boldsymbol{\theta}_f=[\th_\rho^\top,\theta_v]^\top\in\Theta_f\}$, where $\Theta_f =\{(\th_\rho,\theta_v)\in\reals^q\times\reals:\ \th_\rho\in\Theta_\rho,~\theta_v\geq 0\}$ is a set that contains the true parameters $\boldsymbol{\theta}_f^*$ of the $f$-process -- this practice {is common in both} the {Geostatistics and in the Machine Learning literature, e.g., {\cite{CressieBook,RasmussenWilliamsBook}}}. Let $c_f(\bx,\bx',\th_f)\triangleq\theta_v r(\mb{x},\mb{x}',\boldsymbol{\theta}_{\rho})$, where $r(\mb{x},\mb{x}',\boldsymbol{\theta}_{\rho})$ is a parametric correlation function, and $\th_\rho$ 
and $\theta_v$ denote the spatial correlation and variance parameters, {respectively}. For \emph{isotropic} correlation functions 
{we have} $q=1$; {for instance}, the Squared-Exponential (SE) $\exp\left(\frac{-\norm{\mb{x}-\mb{x}'}^2}{\th_{\rho}^2}\right)$ and {the} Matern-$\tfrac{3}{2}$ {function}, $\left(1+\frac{\sqrt{3}\norm{\mb{x}-\mb{x}'}}{\th_{\rho}}\right)\exp\left(\frac{-\sqrt{3}\norm{\mb{x}-\mb{x}'}}{\th_{\rho}}\right)$.
In the isotropic setting, $\th_\rho\in\reals$ is the range parameter, and $\Theta_\rho=\reals_{+}$. In second-order stationary \emph{anisotropic} random fields, the correlation between two points is a function of the vector connecting the two locations rather than the distance, e.g., 
{the anisotropic squared exponential correlation function \vspace*{-3mm}
\begin{equation}\label{eq:anisCorrFunc}
r(\bx,\bx',\th_\rho)=\exp\Big(-(\bx-\bx')^\top M(\th_\rho)(\bx-\bx')\Big), \vspace*{-2mm}
\end{equation}
where $M(\th_\rho)\in\mS^d$ is a symmetric matrix, e.g., $q=d$, $M(\th_\rho)=\diag(\th_{\rho}^{-2})$ and $\Theta_\rho=\reals^d_{+}$. } A
covariance function is called valid if it leads to
a positive definite covariance matrix for any finite set of fixed locations $\{\mb{x}_i\}_{i=1}^n\subset \mathcal{X}$. Let $\boldsymbol{\theta}^*=[{\boldsymbol{\theta}_f^*}^\top,\theta_0^*]^\top\in\Theta$ denote the unknown true parameters of the $y$-process, where $\Theta\triangleq\Theta_f\times\reals_+$. Hence, $c(\mb{x},\mb{x}',\boldsymbol{\theta}^*)\triangleq c_f(\mb{x},\mb{x}',\boldsymbol{\theta}_f^*)+\theta_0^*\delta(\mb{x},\mb{x}')$ denotes the covariance function of the $y$-process. Here, 
$\delta(\mb{x},\mb{x}')=1$ if $\mb{x}=\mb{x}'$ {and equals} 0 otherwise.

Given a set of locations $\cD^x=\{\bx_i\}_{i=1}^n$, and $\boldsymbol{\theta}=[{\boldsymbol{\theta}_f}^\top,\theta_0]^\top\in\Theta$, let $C_f(\boldsymbol{\theta}_f)\in\reals^{n\times n}$ be such that its $(i,j)^{th}$ element is $c_f(\mb{x}_i,\mb{x}_j,\boldsymbol{\theta}_f)$, and define $C(\boldsymbol{\theta})\triangleq C_f(\boldsymbol{\theta}_f)+\theta_0\mb{I}_n$, i.e., $(i,j)^{th}$ element is equal to $c(\mb{x}_i,\mb{x}_j,\boldsymbol{\theta})$. Hence, $C_f(\boldsymbol{\theta}_f^*)$ and $C^*\triangleq C(\boldsymbol{\theta}^*)$ denote true covariance matrices of the $f$-process and $y$-process, resp., corresponding to locations $\{\mb{x}_i\}_{i=1}^n$. The log marginal likelihood function $\ell(\boldsymbol{\theta}~|~\cD)\triangleq \frac{1}{N}\sum_{r=1}^N\log p\left(\mb{y}^{(r)}|~\boldsymbol{\theta},\cD^x\right)$ is written as \vspace*{-2mm}
{\small
\begin{equation*}
\ell(\boldsymbol{\theta}~|~\cD) = -\tfrac{1}{2}\log\det(C(\boldsymbol{\theta}))-\tfrac{1}{2N}\sum_{r=1}^N{\mb{y}^{(r)}}^\top C(\boldsymbol{\theta})^{-1}\mb{y}^{(r)}-\tfrac{n}{2}\log(2\pi).\vspace*{-3mm}
\end{equation*}
}

\noindent Let $S=\frac{1}{N}\sum_{r=1}^N{\mb{y}^{(r)}\mb{y}^{(r)}}^\top$. Given $X,Y\in\reals^{n\times n}$, let $\fprod{X,Y}=\Tr(X^\top Y)$. Hence, finding the MLE of the $y$-process parameters requires solving 
{\small
\begin{equation}\label{eq:likelihoodOpt}
\hat{\boldsymbol{\theta}}_{MLE} = \argmin_{\boldsymbol{\theta}\in\Theta} \fprod{S,C(\boldsymbol{\theta})^{-1}}+\log\det(C(\boldsymbol{\theta})) \vspace*{-4mm}
\end{equation}
}

\noindent over a set $\Theta=\Theta_f\times\reals_+$ containing the true unknown parameters $\boldsymbol{\theta}^*=[{\boldsymbol{\theta}_f^*}^\top,\theta_0^*]^\top$.

The log-likelihood function $\ell(\boldsymbol{\theta}|\cD)$ is \emph{not} concave in $\th$ for many important parametric families of covariance functions.
Therefore, the MLE problem in \eqref{eq:likelihoodOpt} is \emph{non-convex}, which
causes standard optimization routines to be trapped in local minima \citep{MardiaWatkins1989,RasmussenWilliamsBook}. 
The main result of this paper given in Theorem~\ref{thm:secondstage}, and empirical evidence from our numerical experiments 
{indicate} the reason why 
{our} \emph{two-step} SPS approach works 
better compared to well-known one-step \emph{non-convex} log-likelihood maximization approaches. SPS defers dealing with the non-convexity issue to a later stage, and first obtains a regularized log-likelihood estimate of the precision matrix solving a convex problem. At the second stage, a non-convex least-squares problem is solved, 
{in which the} global minimum is ``close" to the true values; moreover, the objective is strongly convex in a considerably ``large" neighborhood of the global minimum -- 
see Figure~\ref{fig:second_phase}.

Several other methods have been proposed in the literature to deal with the ``Big $n$" problem in GRF 
{estimation}. These approaches can be broadly classified in six main classes: \textbf{1)} \emph{Likelihood approximation methods} approximate the likelihood function in the spectral domain~\citep{Fuentes2007, SteinBook}, or approximate it as a product of conditional densities~\citep{Vecchia1988, SteinEtal2004}; \textbf{2)} \emph{Covariance tapering} provides a sparse covariance matrix in which the long range (usually weak) covariance elements are set to zero. Sparse matrix routines are then used to efficiently find the inverse and determinant of the resulting matrix~\citep{FurrerEtal2006}; \textbf{3)} \emph{Low-rank process approximation methods} are based on a truncated basis  expansion of the underlying GRF which results in reducing the computational complexity from $O(n^3)$ to $O(p^3)$, where $p$ is the number of basis functions used to approximate the process~\citep{Higdon2002, CressieJohannesson2008, BanerjeeEtal2008, Nychka2012}; \textbf{4)} \emph{Sampling-based stochastic approximation} draws 
$m$ sample data points ($m\ll n$) at each iteration, and the model parameters are updated according to a stochastic approximation technique until the convergence is achieved~\citep{LiangEtal2013}; \textbf{5)} \emph{Localized GRFs} split the input domain into different segments, and the covariance parameters are estimated via ML locally on each segment~\citep{GramacyApley2013} -- this approach requires further formulation to avoid discontinuities on the predicted surface over the full domain~\citep{Park2011}; and finally \textbf{6)} \emph{Markov random field approximations}, related to our proposed SPS method, will be discussed in more detail in Section~\ref{sec:motivation}. There are also methods in the intersection of two classes: \cite{SangHuang2012} combined low-rank process approximation with covariance tapering; \cite{snelson2007} proposed a mix of likelihood approximation and localized GRF; {\cite{hensman13gaussian} used stochastic variational inference for GRF models and extended their approach to non-Gaussian and latent variable models; \cite{rodner12large} considered using GRF models for multi-class classification under the big $n$ scenario and used parametrized histogram intersection kernels.}

The rest 
is organized as follows: in Section~\ref{sec:motivation}, we motivate the proposed method. In Sections \ref{sec:covSelection} and \ref{sec:statProp} we discuss the two-stage SPS method in detail and prove the statistical properties of the SPS estimator. 
From a computational perspective, it is shown that the first stage has linear convergence rate, and that the second stage problem is strongly convex around the estimator, which can be solved efficiently via a line search for isotropic GRFs. Next, in Section~\ref{sec:numericals}, we assess the prediction performance of the proposed method comparing it to alternative methods
on both synthetic and real data. Finally, in Section~\ref{sec:concludings} we conclude by providing some summarizing remarks and directions for further research.

\section{Motivation for the proposed SPS method} \label{sec:motivation}
\vspace*{-0.25cm}
Let $\boldsymbol{\theta}^*\in\Theta$ be the true {covariance} parameters, and $C^*=C(\boldsymbol{\theta}^*)\in\reals^{n\times n}$ be the true covariance matrix of the $y$-process corresponding to given locations $\cD^x=\{\bx_i\}_{i=1}^n$. The proposed method can be motivated by providing four interrelated remarks: \textbf{a)} the precision matrix $P^*\triangleq {C^*}^{-1}$ of a stationary 
GRF  
can be approximated with a sparse matrix; \textbf{b)} powerful convex optimization algorithms exist to solve Sparse Covariance Selection (SCS) problems 
to find a sparse approximation to $P^*$; \textbf{c)} the past and recent work on directly approximating a GRF with a GMRF also involves determining a sparse precision matrix; \textbf{d)} the available distance information among given locations can be incorporated into the GRF estimation. 

\noindent \textbf{a)~first motivation:}  
Our 
method is 
motivated by the 
observation that 
$P^*={C^*}^{-1}$ of a stationary 
GRF can be approximated by a sparse matrix. The {\it off-diagonal} element $P^*_{ij}$, i.e., $i\neq j$, 
is 
determined by the conditional covariance (\emph{partial covariance}) between 
$y(\mb{x}_i)$ and $y(\mb{x}_j)$ given the rest of the variables; indeed, for any $i\neq j$, 
$|P^*_{ij}|\rightarrow 0$ as $\cov\big(y(\mb{x}_i),~y(\mb{x}_j)~|~\{y(\mb{x}_k)\}_{k\neq i,j}\big)\rightarrow 0$ because for $i\neq j$ we have 
{\small
\begin{equation*}
P^*_{ij}=\frac{-\cov\big(y(\mb{x}_i),~y(\mb{x}_j)~|~\{y(\mb{x}_k)\}_{k\neq i,j}\big)}
{\var(y(\mb{x}_i)|\{y(\mb{x}_k)\}_{k\neq i,j})\var(y(\mb{x}_j)|\{y(\mb{x}_k)\}_{k\neq i,j})-\cov(y(\mb{x}_i),y(\mb{x}_j)|\{y(\mb{x}_k)\}_{k\neq i,j})^2}.
\end{equation*}
}%

\begin{figure}[!h]
  \centering
  \includegraphics[width=0.7\columnwidth]{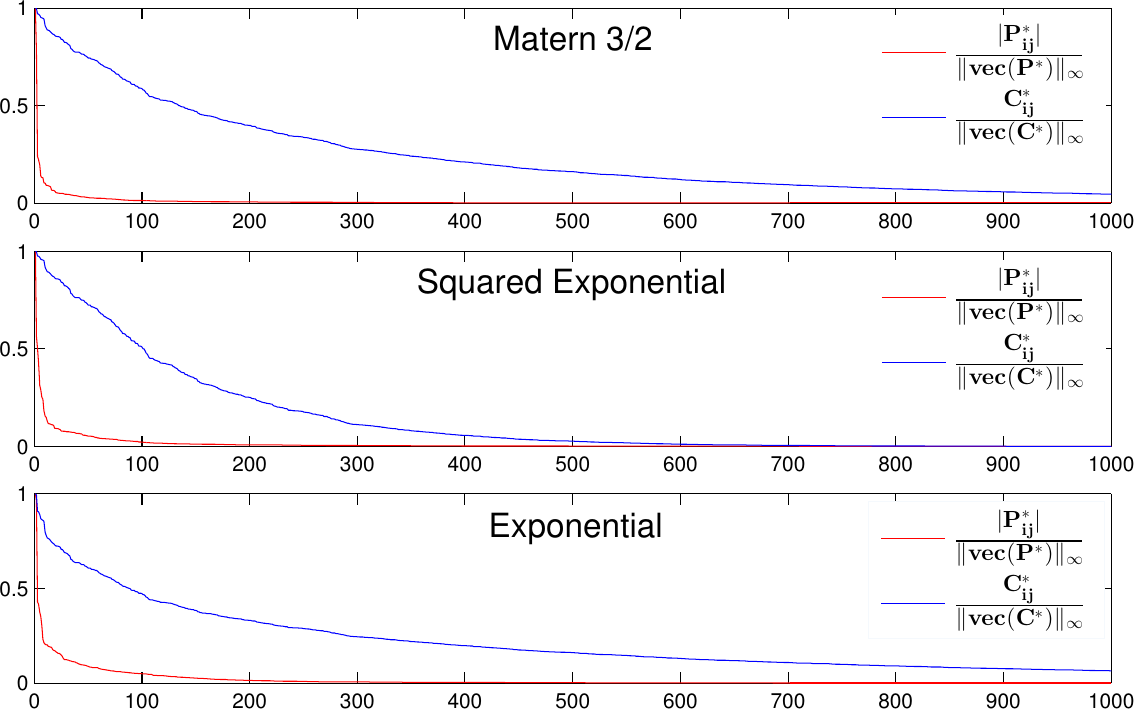}
   \caption{{\scriptsize 
   Decaying behavior of elements of the Precision and Covariance matrices for GRFs. 
   Scaled magnitudes of the largest 1000 
   off-diagonal elements of the precision and covariance matrices (scaled by their maximums) are plotted after sorting in descending order. 
   The underlying GRF was evaluated over a set of 100 points {generated uniformly at random from} $\cX=\{\mb{x}\in\reals^2: -50 \leq \mb{x}\leq 50\}$ for three different covariance functions with range and variance parameters equal to 10, and 1, resp.
   }} \label{fig:covSparsity}
\end{figure}
In particular, \emph{conditionally independent} variables lead to a zero entry in the precision matrix \citep{WhittakerBook}. This is 
why sparse precision matrices are common in graphical models and Bayesian networks~
\citep{WhittakerBook} when most of the variable pairs are conditionally independent. 
The fact that the precision matrix of a GRF is close to sparse is related to the interesting behavior of the so-called {\em screen effect} in a spatial process (\cite{CressieBook}, p. 133;~\cite{JournelHuijbregtsBook}, p. 346). The \emph{screen effect} is complete in $\mathbb{R}^1$, i.e., for given three data points on a line, the two outer points are conditionally independent (in time series models, the partial (auto)correlation function ``cuts off" after lag $k$ for a Markovian AR($k$) process -- see~\cite{BoxJenkins}). However, for a GRF in $\mathbb{R}^d$ with $ d>1$, the screen effect is \emph{not} complete; hence, the corresponding precision matrix is not sparse for any {\em finite} set of variables pertaining to the GRF.

In addition, existing results from numerical linear algebra demonstrate that if the elements of a matrix {show} a decay property, then {the elements of} its inverse also {show} a similar behavior -- see~\cite{benzi16,jaffard1990proprietes}. In particular, consider the two decay {classes} 
defined in~\cite{jaffard1990proprietes}:
\begin{definition}\label{def:jaffard}
Given $\{\mb{x}_i\}_{i=1}^n\subset\cX$ and a metric $d:\cX\times\cX\rightarrow\reals_+$, a matrix $A\in\reals^{n\times n}$ belongs to the class $\cE_\gamma$ for some $\gamma>0$ if for all $\gamma'<\gamma$ there exists a constant $K_{\gamma'}$ such that
\begin{equation}
|A_{ij}|\leq K_{\gamma'}\ \exp\big(-\gamma' d(\mb{x}_i,\mb{x}_j)\big),\quad \forall 1\leq i,j\leq n.
\end{equation}
Moreover, $A$ belongs to the class $\cQ_\gamma$ for some $\gamma>1$ if there exists a constant $K$ such that
\begin{equation}
|A_{ij}|\leq K\ \big(1+d(\mb{x}_i,\mb{x}_j)\big)^{-\gamma},\quad \forall 1\leq i,j\leq n.
\end{equation}
\end{definition}
\begin{theorem}
\label{thm:decay}
Given $\{\mb{x}_i\}_{i=1}^n\subset\cX$ and a metric $d:\cX\times\cX\rightarrow\reals_+$, let $A\in\reals^{n\times n}$ be an invertible matrix. If $A\in\cE_\gamma$ for some $\gamma>0$, then $A^{-1}\in\cE_{\gamma'}$ for some $\gamma'>0$. Moreover, if $A\in\cQ_\gamma$ for some $\gamma>0$, then $A^{-1}\in\cQ_{\gamma}$.\vspace*{-3mm}
\end{theorem}
\begin{proof}
See Proposition~2 and Proposition~3 in~\citep{jaffard1990proprietes}.
\end{proof}
This fast decay structure in the precision matrix makes it a ``\emph{compressible signal}''~\citep{candes2006compressive}, which is not sparse; however, due to quick decay in its entries when sorted by their magnitude, 
it can be well-approximated by a sparse matrix -- see Corollary~\ref{cor:probBound4JaffardCov}. 
{For} all 
stationary GRFs tested, we 
observed that for a finite set of {locations, the} magnitudes of the off-diagonal elements of the {\em precision} matrix decay 
to 0 
 much \emph{faster} than 
 {the elements of} the covariance matrix. 
To illustrate {this behavior}, we compared the decay in covariance and precision elements in Figure~\ref{fig:covSparsity} for data generated from GRFs with Matern-$\tfrac{3}{2}$, Squared Exponential, and Exponential covariance functions. Clearly, the precision matrix can be \emph{better} approximated {than the covariance matrix} by {using} a sparse matrix. 

\begin{table}[!h]
{\footnotesize
\parbox{.50\linewidth}{
$\%$ of elements s.t. $|\tilde{P}_{ij}|>\epsilon$
\centering
\begin{tabular}{lccc}
    \hline
    \hline
    $\epsilon\setminus n$ & 10 & 100 & 1000 \\
    \hline
    0.1   & 10.49 & 0.29 & 0.00 \\
    0.01  & 26.94 & 2.52 & 0.03 \\
    0.001 & 46.46 & 9.34 & 0.28 \\
    \hline
\end{tabular}
}
\hspace{-1cm}
\parbox{.50\linewidth}{
$\%$ of elements s.t. $|\tilde{C}_{ij}|>\epsilon$
\centering
\begin{tabular}{lccc}
    \hline
    \hline
    $\epsilon \setminus n$ & 10 & 100 & 1000 \\
    \hline
    0.1   & 16.82 & 13.00 & 12.94 \\
    0.01  & 36.84 & 32.31 & 32.18 \\
    0.001 & 56.86 & 53.04 & 52.86 \\
    \hline
\end{tabular}
}
}
\caption{{\scriptsize Effect of increasing density in a fixed spatial domain on the near-sparsity of precision \textbf{(left)} and covariance \textbf{(right)} matrices for 
Matern GRF ($\nu=3/2$) 
over $100 \times 100$ fixed spatial domain. 
}}
\label{tbl:sparsityAsFuncN}
\vspace*{-0.2cm}
\end{table}

By fixing the domain of the process and increasing $n$ (increasing the density of the data points), the \emph{screen effect} becomes stronger, i.e., off-diagonal entries decay to 0 faster. Hence, the precision matrices can be better approximated with 
a sparse matrix as $n$ increases in a fixed domain. To illustrate this phenomenon numerically, we calculate the precision matrices corresponding to 
a Matern GRF ($\nu=3/2$) with variance and range parameters equal to 1 and 10, resp., for $n\in\{10, 100, 1000\}$ over a \emph{fixed} square domain 
$\cX=\{\mb{x}\in\reals^2: -50 \leq \mb{x}\leq 50\}$. Then, as a measure of near-sparsity, we computed the percentage of scaled off-diagonal elements in the precision matrix greater in absolute value than certain threshold $\epsilon\in\{0.1, 0.01, 0.001\}$, i.e., {\small $\mbox{card}\left(\{(i,j):\ |\tilde{\mbox{P}}_{ij}|>\epsilon,\ 1 \leq i \neq j\leq n\}\right)/(n^2-n)$}, where {\small $\tilde{P}_{ij}=P^*_{ij}/\max\{|P^*_{ij}|:\ 1 \leq i \neq j\leq n\}$}. For comparison, we report the same quantities for the covariance matrices 
in Table~\ref{tbl:sparsityAsFuncN}. This shows the effect of {\em infill asymptotics}~\citep{CressieBook} on the \emph{screen effect}:
in a fixed domain as $n$ increases, precision matrices get closer to sparse matrices, while the covariance matrices are {much less sensitive} to increasing density.\\ 

\noindent \textbf{b)~second motivation:} Our work is also motivated by the recent optimization literature on the Sparse Covariance Selection~(SCS) problem~\citep{Dempster1972} in \eqref{eq:CSConvex} -- compare it with \eqref{eq:likelihoodOpt}. 
Given a sample covariance matrix $S\in\reals^{n\times n}$ of a zero-mean multivariate Gaussian 
$\mb{y}\in\reals^n$, 
\cite{dAspremont2008} proposed to estimate the corresponding precision matrix by solving a regularized \emph{maximum likelihood problem}:
$\min_{P \succ 0} \fprod{S,P} - \mbox{log} \; \mbox{det}(P) + \alpha~\mbox{card}(P)$,
where 
$\mbox{card}(P)$ denotes the cardinality of non-zero elements of $P$, 
$P \succ 0$ denotes the cone of symmetric, positive definite (PD) matrices. This problem is combinatorially hard due to 
the cardinality operator in the objective function. 
Since the $\ell_1$-norm, defined as $\norm{P}_1\triangleq\sum_{1\leq i,j\leq n}|P_{ij}|$, is the tightest convex envelope of $\mbox{card}(.)$, a convex approximation 
problem can be formulated as \vspace*{-3mm}
{\small
\begin{equation}\label{eq:CSConvex}
\min_{P \succ 0} \quad \fprod{S,P} - \mbox{log} \; \mbox{det}(P) + \alpha \norm{P}_1. \vspace*{-4mm}
\end{equation}
}
\vspace*{-3mm}

\noindent 
The growth of interest in SCS in the last decade is mainly due to development of first-order algorithms that can efficiently deal with large-scale $\ell_1$-regularized convex problems~{\citep{Yuan2012, friedman2008, Mazumder2012, Honorio2013, Hsieh_NIPS2013, lu2010adaptive, lu2009smooth, Scheinberg_NIPS2010}}.

\noindent \textbf{c)~third motivation:} Further motivation 
comes from 
{prior} work on approximating a GRF with a Gaussian Markov Random Field (GMRF) to obtain computational gains using sparsity.
A GRF process on a \emph{lattice} 
is a GMRF under the conditional independence assumption, i.e., 
a variable is conditionally independent of the other variables on the lattice given its ``neighbors"
\citep{RueHeldBook}. While the index set 
is countable for the lattice data, the index set $\mathcal{X}$ for a GRF is \emph{uncountable}; hence, in general GMRF models cannot represent GRFs {\em exactly}. For a very special class, \cite{LindgrenRueLindstrom2011} recently established
that the Matern GRFs are Markovian; in particular, when the smoothing parameter $\nu$ is such that $\nu-d/2\in\integers_+$, where $d$ is the dimension of the input space --
{see~\cite{LindgrenRueLindstrom2011,Simpson2012}}
for using this idea in the approximation of anisotropic and non-stationary GRFs.
Rather than using a triangulation of the input space as proposed by \cite{LindgrenRueLindstrom2011}, or assuming a lattice process, we let the data determine the near-conditional independence pattern between variables through the precision matrix estimated via a weighted $\ell_1$-regularization similar to that used in the SCS problem.

\begin{figure}[H]
    \hspace*{-1.6cm}
  \includegraphics[width=1.2\columnwidth]{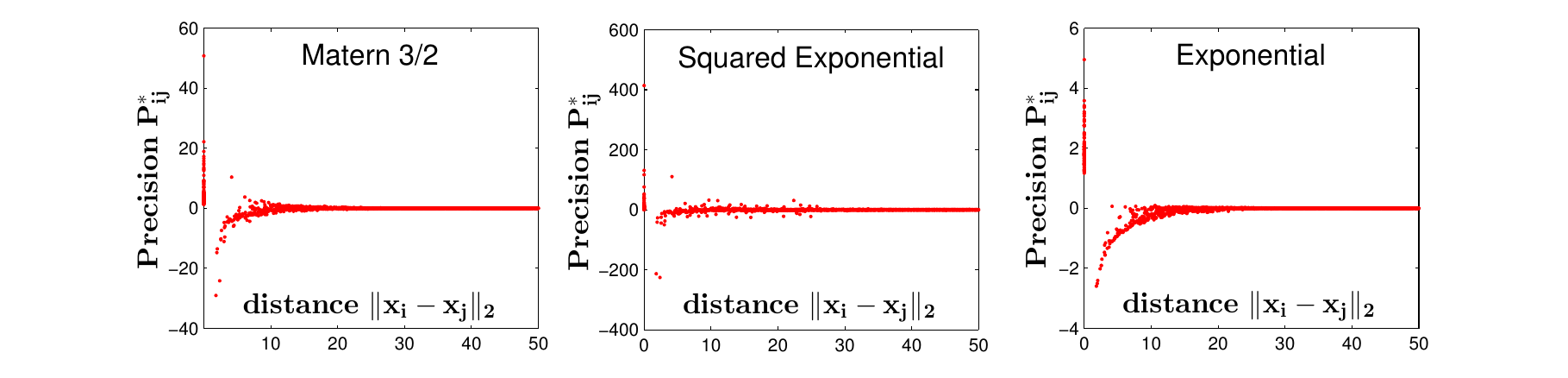}
\vspace{-0.8cm}
  \caption{{\scriptsize Elements of precision matrices from three covariance functions as a function of the Euclidean distance between the data points. The variance, range, and nugget parameters of the covariance functions are 1, 10, and 0, respectively.}} \label{fig:precisionVsDist}
  \vspace*{-0.6cm}
\end{figure}

\noindent \textbf{d)~fourth motivation:} 
Since the spatial locations of the observations are known, i.e., $\cD^x$, these data can be utilized 
to improve the estimation even when the number of realizations at each location is low. 
{As established in Theorem~\ref{thm:decay}, since $|C^*_{ij}|$ decreases to $0$ exponentially as $\norm{\mb{x}_i-\mb{x}_j}_2$ increases, $|P^*_{ij}|$ decays to $0$ exponentially fast as well. In fact, for all stationary covariance functions tested, we observed this behavior -- see Figure~\ref{fig:precisionVsDist}.}
Therefore, 
this information can be utilized for regularizing the likelihood function~(see Section~\ref{subsec:convexProblem}).\vspace*{-0.3cm} 

\section{The SPS algorithm for fitting a GRF model} \label{sec:covSelection}
\vspace*{-0.15cm}
The proposed method for fitting a GRF is composed of two stages: 1) the true precision matrix corresponding to the training data set is approximated with a sparse matrix by solving a convex maximum likelihood 
problem regularized with a weighted $\ell_1$-norm; 2) after inverting the fitted precision matrix from 
the first stage, a least-squares problem is solved to estimate the unknown covariance function parameters. \vspace*{-2mm} 

\subsection{STAGE-I: Estimation of precision matrices} \label{subsec:convexProblem}
{Consider the \emph{unknown} parameter vector} $\th^*=[{\th^*_f}^\top,\theta^*_0]^\top\in\Theta\triangleq\Theta_f\times\reals_+$ {and} suppose $\theta_0\geq 0$ and $\th_f^*\in\mathrm{int}(\Theta_f)$, i.e., $\th^*_\rho\in\mathrm{int}(\Theta_\rho)$, $\th^*_v>0$. Let $C^*=C(\th^*)$ be the covariance matrix of a \emph{zero-mean} GRF corresponding to $\cD^x=\{\bx_i\}_{i\in\cI}\subset\cX$, and $P^*=(C^*)^{-1}$, where $\cI\triangleq\{1,\ldots,n\}$. Fix 
$0\leq a^*\leq b^*\leq\infty$ satisfying $0\leq \frac{1}{b^*}\leq\sigma_{\min}(C^*)\leq\sigma_{\max}(C^*)\leq \frac{1}{a^*}$ where $\sigma_{\min}(\cdot)$ and $\sigma_{\max}(\cdot)$ denote minimum and maximum singular values, respectively; hence, $a^*\leq\sigma_{\min}(P^*)\leq\sigma_{\max}(P^*)\leq b^*$. Given $\mathcal{D} = \{(\mb{x}_i,y^{(r)}_{i}): i\in\cI,\ r=1,...,N\}$, compute the unbiased estimator of $C^*$, 
$S =\frac{1}{N}\sum_{r=1}^N \mb{y}^{(r)}{\mb{y}^{(r)}}^\top\in\mathbb{S}^n$, where $\mathbb{S}^n$ denotes the set of $n\times n$ symmetric matrices, $\mb{y}^{(r)}=[y_i^{(r)}]_{i=1}^n\in\reals^n$. 
Furthermore, we form 
the distance matrix $\tilde{G}\in\mathbb{S}^n$ as follows:
\vspace{-2mm}
{\small
\begin{equation}
\tilde{G}_{ij} = \norm{\mb{x}_i-\mb{x}_j}, \quad \mbox{if $i\neq j$}, \quad 
\tilde{G}_{ii} = \min\{\norm{\mb{x}_i-\mb{x}_j}:\ j\in\cI\setminus\{i\}\}  \label{eq:distMat_original}
\end{equation}
}%
for all $(i,j)\in\cI\times \cI$. Let $\tilde{G}_{\max}\triangleq\max_{i,j}\tilde{G}_{ij}$ and $\tilde{G}_{\min}\triangleq\min_{i,j}\tilde{G}_{ij}$. Next, we define the weight matrix as
\vspace{-2mm}
{\small
\begin{equation}
G_{ij} = \tilde{G}_{ij}/\tilde{G}_{\min}, \quad \mbox{if $i\neq j$}, \quad 
G_{ii} = \tilde{G}_{ii}/\tilde{G}_{\min} \label{eq:distMat}
\end{equation}
}%
To approximate the true precision matrix with a sparse matrix, we propose to solve the following convex problem:
{\small
\begin{equation}\label{eq:convexProgram}
\hat{P} \triangleq \argmin\{\fprod{S,P}-\log\det(P)+\alpha\fprod{G,|P|}:\ a^*\mb{I}\preceq P \preceq b^*\mb{I}\}, \vspace*{-1mm}
\end{equation}
}%
where $|.|$ is the element-wise absolute value operator; hence, the last term is a weighted $\ell_1$-norm with weights equal to the normalized distances $G_{ij}$ -- compare it with \eqref{eq:likelihoodOpt} and \eqref{eq:CSConvex}. The choice of weighted $\ell_1$-norm with weights equal to the pairwise distances between the points is supported by the trends observed in Figure~\ref{fig:precisionVsDist}. Choosing $G_{ii}$ as in \eqref{eq:distMat} controls diagonal elements of $\hat{P}$ compared to off-diagonal elements; otherwise, they might get unreasonably big. 
Note that $\hat{P}$ is always a full rank matrix due to the $\log\det(\cdot)$ term in the objective function. Furthermore, having non-trivial bounds $0<a^*\leq b^*<\infty$ is useful in practice to control the condition number of the estimator, which is also argued for in~\citep{dAspremont2008,Rothman2008sparse}.

If there is no prior information on the process to obtain non-trivial $0<a^*\leq b^*<\infty$, then setting $a^*=0$, and $b^*=\infty$ trivially satisfies the condition on $a^*$ and $b^*$. For this case, \eqref{eq:convexProgram} reduces to $
\min\{\fprod{S,P}-\log\det(P)+\alpha\fprod{G,|P|}:\ P\succ \mathbf{0} \}$. On the other hand, when there is prior information on the process, one can also exploit it to obtain non-trivial bounds $a^*$ and $b^*$. For instance, let $C^*=C(\th^*)=C_f(\th^*_f)+\theta^*_0\mathbf{I}$ be the true covariance matrix corresponding to locations in $\mathcal{D}^x$, 
where $\th^*_f=[{\th^*_\rho}^\top,\theta^*_v]^\top$, $\th_{\rho}^*\in\mathrm{int}(\Theta_\rho)$ and $\theta^*_v>0$ denote the true spatial correlation and variance parameters of the $f$-process. The common structure of the covariance functions implies that $\diag(C_f(\th^*_f))=\theta_v^*\mathbf{1}$, where $\mathbf{1}$ denotes the vector of ones. Therefore, 
$\sigma_{\min}(P^*)\geq1/\Tr(C(\th^*))=\tfrac{1}{n(\theta^*_0+\theta^*_v)}$. Hence, if upper bounds on $\theta^*_0$ and $\theta^*_v$ are known a priori, then one can obtain non-trivial lower bounds. 

In comparison to the 
SCS problem in \eqref{eq:CSConvex}, the proposed 
{formulation} \eqref{eq:convexProgram} penalizes each element of the 
estimator $\hat{P}_{ij}$ for $i\neq j$ with a different weight proportional to $\norm{\mb{x}_i-\mb{x}_j}_2$, i.e., the distance between the corresponding locations.
This model assumes that the off-diagonal precision magnitudes decrease with distance, for which there is empirical evidence as shown in Figure~\ref{fig:precisionVsDist}. {More importantly, Theorem~\ref{thm:decay} shows that this assumption indeed always holds when the covariance 
elements decay with increasing $\norm{\mb{x}_i-\mb{x}_j}_2$}. Moreover, the reason we solve \eqref{eq:convexProgram} using $G$ with strictly positive diagonal is that otherwise the diagonal elements of the precision matrix would not be penalized and they might attain relatively large positive values.

The form of STAGE-I problem in \eqref{eq:convexProgram} is well-studied in the optimization literature. Indeed, for any $0\leq a^*\leq b^*\leq\infty$, one can use the Alternating Direction Method of Multipliers~(ADMM) to generate a sequence of iterates that {\it Q-linearly}\footnote{Let $\{X_\ell\}$ converge to $X^*$ for a given norm $\norm{.}$. The convergence is called $Q$-linear if $\frac{\norm{X_{\ell+1}-X^*}}{\norm{X_{\ell}-X^*}}\leq c$, for some $c\in(0,1)$} converges to $\hat{P}$, where $\hat{P}$ is the unique optimal solution to STAGE-I problem given in \eqref{eq:convexProgram}. For the sake of completeness, in the online-only supplementary material, we provide an ADMM algorithm and state its convergence properties -- see Figure~\ref{alg:admm} and Theorem~\ref{thm:admm} in the online supplement.

\begin{singlespace}
\begin{figure}[!h]
    {\small
    \rule[0in]{6.5in}{1pt}\\
    \textbf{Algorithm SPS}$~(\mathcal{D})$\\
    \rule[0.125in]{6.5in}{0.1mm}
    \vspace{-0.35in}
    \begin{algorithmic}[1]
    \STATE $\mathbf{input:}\ \mathcal{D}=\{(\bx_i,~y_i^{(r)}):\ r=1,\ldots,N,\ i\in\cI\}\subset\cX\times\reals$
    \STATE /* Compute sample covariance and the distance penalty matrices */
    \STATE $\by^{(r)}\gets [y^{(r)}_i]_{i=1}^n$, $S \gets \frac{1}{N}\sum_{r=1}^N\by^{(r)}{\by^{(r)}}^\top$
    \STATE $G_{ij} \gets \norm{\bx_i-\bx_j}_2,\ \forall (i,j)\in\cI\times \cI$,  $G_{ii}\gets \min\{\norm{\bx_i-\bx_j}_2:\ j\in\cI\setminus\{i\}\},\ \forall i\in\cI$
    \STATE  /* Compute the fitted precision matrix -- See Section~\ref{subsec:convexProblem} */
    \STATE $\hat{P} \gets \argmin\left\{\fprod{S,P}-\log\det(P)+\alpha\fprod{G,|P|}:\ a^* \bI \preceq P\preceq b^* \bI \right\}$
    \STATE /* Estimate covariance function parameters -- See Section~\ref{subsec:nonconvexProblem} */
    \STATE $\hat{\th} \gets \argmin_{\th\in\Theta} \norm{C(\boldsymbol{\theta})-{\hat{P}}^{-1}}_F^2$
    \RETURN $\hat{\th}$
    \end{algorithmic}
    \rule[0.25in]{6.5in}{0.1mm}
    }
    \vspace*{-0.5in}
    \caption{{\scriptsize Sparse Precision matrix Selection~(SPS) method}}\label{alg:1}
    \vspace*{-6mm}
\end{figure}
\end{singlespace}

\subsection{STAGE-II: Estimation of covariance function parameters} \label{subsec:nonconvexProblem}
After estimating the precision matrix in the first stage {according to} \eqref{eq:convexProgram}, a least-squares problem is solved in the second stage to fit a parametric covariance function to $\hat{P}$. Although this is a non-convex problem for parametric covariance functions in general, 
our main result, Theorem~\ref{thm:secondstage}, and empirical evidence from our numerical experiments suggest that non-convexity of this problem is much less serious than that of the likelihood function~\eqref{eq:likelihoodOpt}. In STAGE-II, we propose to estimate the covariance parameters by solving \vspace*{-1mm} 
{\small
\begin{equation} \label{eq:nonconvexFrobStationary}
\hat{\boldsymbol{\theta}} \in \argmin_{\boldsymbol{\theta}\in\Theta} \norm{C(\boldsymbol{\theta})-{\hat{P}}^{-1}}_F^2,
\vspace*{-4mm}
\end{equation}
}

\noindent where $\boldsymbol{\theta}=[\boldsymbol{\theta}_{\rho}^\top,\theta_v,\theta_0]^\top$, $\Theta\triangleq\{\th:\ \th_\rho\in\Theta_\rho,~\theta_v\geq 0,~\theta_0\geq 0\}$, and $C(\boldsymbol{\theta})$ is the parametric covariance matrix corresponding to the locations of the training data $\mathcal{D}^x$. {Here }
$\boldsymbol{\theta}_{\rho}\in\reals^q$ denotes the spatial parameters of the \emph{correlation} function, $\theta_v$ is the variance parameter, and $\theta_0$ is 
the nugget, which in some applications is set equal to zero. Indeed, $C_{ij}(\boldsymbol{\theta})=c(\bx_i,\bx_j,\th)$.
The two stage SPS method 
is summarized in Figure~\ref{alg:1}. 

\textbf{Solution to the STAGE-II problem.}
Let $\hat{C}\triangleq\hat{P}^{-1}$, where $\hat{P}$ is defined in \eqref{eq:convexProgram}. Consider sequentially solving 
\eqref{eq:nonconvexFrobStationary}: Fixing $\boldsymbol{\theta}_{\rho}$, the objective in \eqref{eq:nonconvexFrobStationary} is first minimized over $\theta_v$ and $\theta_0$ in closed form 
(inner optimization); hence,
it can be written as a function of $\boldsymbol{\theta}_{\rho}$ \emph{only}. Next, the resulting function is minimized over $\boldsymbol{\theta}_{\rho}$ (outer optimization), i.e., 
{\small
\begin{flalign}\label{eq:app-nonconvexOpt}
\min_{\boldsymbol{\theta}_{\rho}\in\Theta_\rho}\Big\{\min_{\theta_v\geq0,\ \theta_0\geq0}\ \frac{1}{2}\sum_{i,j\in\mathcal{I}} \Big(\theta_v~r(\mb{x}_i,\mb{x}_j,\boldsymbol{\theta}_{\rho})+\theta_0~\delta(\mb{x}_i,\mb{x}_j)-\hat{C}_{ij}\Big)^2\Big\}. \vspace*{-3mm}
\end{flalign}
}%
{where 
$\delta\left(\mb{x}_i,\mb{x}_j\right)=1$ if $\mb{x}_i=\mb{x}_j$, and equals 0 otherwise.}
Consider the inner optimization problem written as follows: \vspace*{-2mm}
{\small
\begin{equation}\label{eq:appB-vectorForm}
f(\th_\rho;\hat{\bc})\triangleq\min_{\theta_0\geq0,~\theta_v\geq0}\ 
\tfrac{1}{2}\norm{\theta_v\mb{r}(\boldsymbol{\theta}_{\rho})+\theta_0\mb{d}-\hat{\mb{c}}}^2, \vspace*{-3mm}
\end{equation}
}%
where $\norm{\cdot}$ denotes the Euclidean norm, $\hat{\mb{c}}$, $\mb{r}$, and $\mb{d}$ are \emph{long vectors} in $\reals^{n^2}$ such that $\hat{\mb{c}}_{ij}=\hat{C}_{ij}$, $\mb{r}_{ij}(\boldsymbol{\theta}_{\rho})=r\left(\mb{x}_i,\mb{x}_j,\boldsymbol{\theta}_{\rho}\right)$, and $\mb{d}_{ij}=\delta\left(\mb{x}_i,\mb{x}_j\right)$ for $(i,j)\in\mathcal{I}$.
We write $\mb{r}(\boldsymbol{\theta}_{\rho})$ as $\mb{r}$ for short
when we do not need to emphasize the dependence of $\br$ on $\boldsymbol{\theta}_{\rho}$.

\begin{theorem} \label{thm:innerOpt}
For any given $\boldsymbol{\theta}_{\rho}\in\Theta_\rho$, the minimization problem in \eqref{eq:appB-vectorForm} has a unique global optimal solution $(\hat{\theta}_v,\hat{\theta}_0)$ that can be computed as
{\small
\vspace*{-2mm}
\begin{equation}\label{eq:innerOptSolution}
(\hat{\theta}_v,\hat{\theta}_0) =
\begin{cases}
(0,~\mb{d}^\top\hat{\mb{c}}/n) & \text{ if }\ \mb{r}^\top\hat{\mb{c}}\leq\mb{d}^\top\hat{\mb{c}}, \\
\left(\frac{\mb{r}^\top\hat{\mb{c}}-\mb{d}^\top\hat{\mb{c}}}{\norm{\mb{r}}^2-n},\frac{\mb{d}^\top\hat{\mb{c}}~\norm{\mb{r}}^2/n-\mb{r}^\top\hat{\mb{c}}}{\norm{\mb{r}}^2-n}\right) & \text{ if }\ \mb{d}^\top\hat{\mb{c}}<\mb{r}^\top\hat{\mb{c}}<\mb{d}^\top\hat{\mb{c}}~\norm{\mb{r}}^2/n, \\
(\mb{r}^\top\hat{\mb{c}}/\norm{\mb{r}}^2,0) & \text{ if }\ \mb{d}^\top\hat{\mb{c}}~\norm{\mb{r}}^2/n\leq \mb{r}^\top\hat{\mb{c}}.
\end{cases}
\vspace*{-4mm}
\end{equation}
}%
\end{theorem}
\vspace*{-1mm}
\begin{corollary}\label{cor:innerOptNoNugget}
In the absence of the nugget parameter $\theta_0$, i.e., $\theta^*_0=0$, $\hat{\theta}_v\triangleq\max\{0, \mb{r}^\top\hat{\mb{c}}/\norm{\mb{r}}^2\}$ is the unique global minimizer to the problem $\min_{\theta_v\geq0}\ 
\tfrac{1}{2}\norm{\theta_v\mb{r}-\hat{\mb{c}}}^2$. 
\end{corollary}
\vspace*{-1mm}

\begin{figure}[H]
  \centering
  \includegraphics[height=4.50cm]{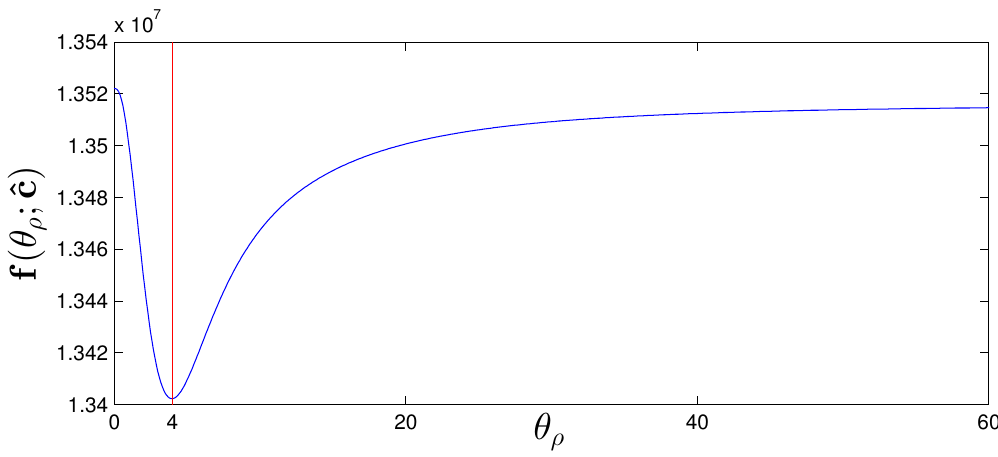}
  \caption{{\scriptsize STAGE-II outer optimization objective $f(\th_\rho;\hat{\bc})$. 
  The red line shows the true parameter $\th_\rho^*=4$}.}
  \label{fig:second_phase}
\end{figure}

Using Theorem~\ref{thm:innerOpt} or Corollary~\ref{cor:innerOptNoNugget}, the solution to the inner problem can be computed as a function of the outer optimization variable, $\boldsymbol{\theta_{\rho}}$, in \eqref{eq:app-nonconvexOpt}. 
In Lemma~\ref{strongConvexityOuterOpt}, we show that under certain conditions, the outer problem objective, $f(\th_\rho;\hat{\bc})$, 
is strongly convex in $\th_\rho$ 
around the global minimum. 
Moreover, for isotropic covariance functions, 
$\hat{\th}_{\rho}=\argmin\{f(\th_\rho;\hat{\bc}):\ \th_{\rho}\in\reals_{+}\}$ can be simplified to a one-dimensional line search 
over $[0, D_{\max}]$, where $D_{\max}$ is an upper bound on $\hat{\th}_{\rho}$. 
This is illustrated in Figure~\ref{fig:second_phase} which displays the STAGE-II objective 
as a function of $\th_{\rho}$. {$f(\th_\rho;\hat{\bc})$ is formed as described in Section~\ref{subsec:nonconvexProblem} using the sample data $\mathcal{D}$ coming from an \emph{isotropic} GRF with true parameters} $(\th_{\rho}^*=4,\theta_{\nu}^*=8,\theta_0^*=4)$ for
a SE covariance function. 
$f(\cdot;\hat{\bc})$ is \emph{unimodal} with global minimum close the true $\th_{\rho}^*$ value. The univariate minimization is performed via bisection; hence, after $\log_2(D_{\max}/\epsilon)$ iterations, the search reaches a point within $\epsilon$-ball 
of $\hat{\th}_\rho$. 

\section{Statistical analysis of the SPS estimator}\label{sec:statProp}
\vspace*{-1mm}
In this section, we focus on the statistical convergence of the parameter estimates obtained by the SPS algorithm displayed in Figure~\ref{alg:1}. 
Given 
$S,G\in\mathbb{S}^{n}$, the SPS estimator of the precision matrix, defined in \eqref{eq:convexProgram}, can be computed 
efficiently using an ADMM algorithm -- see Theorem~\ref{thm:admm} in the online supplement. 
Throughout this section, we assume that non-trivial bounds $0<a^*\leq b^*<\infty$ are given. This same assumption is also made  by~\cite{Rothman2008sparse} to analyze the statistical properties of an estimator related to $\hat{P}$ in~\eqref{eq:convexProgram} -- see Assumptions A2 and A3 in~\citep{Rothman2008sparse}. Moreover, these bounds are useful in practice to control the condition number of the estimator~\citep{dAspremont2008}.

\subsection{Learning the precision matrix}
\begin{theorem}\label{thm:statAnalysis}
Let $\{\by^{(r)}\}_{r=1}^N\subset\reals^n$ be independent realizations of a GRF with zero-mean and stationary covariance function $c(\mb{x},\mb{x}',\boldsymbol{\theta}^*)$ observed over $n$ distinct locations $\{\bx_i\}_{i\in\cI}$ with $\cI\triangleq\{1,...,n\}$; furthermore, let $P^*\triangleq{C^*}^{-1}$ be the corresponding true precision matrix for these observations. Finally, let $\hat{P}$ be the SPS estimator computed as in \eqref{eq:convexProgram} for some $G\in\mS^n$ such that $G_{ij}\geq 0$ for all $(i,j)\in\cI\times\cI$.
Then for any given $M>0$ and $N\geq N_0\triangleq\left\lceil2(M+2)\log n+\log 16\right\rceil$,
we have \vspace*{-1mm}
{\small
\begin{equation*}\label{eq:probBoundGaussProcess}
\mbox{Pr}\Big(\norm{\hat{P}-P^*}_F\leq 2 {b^*}^2\big(n+\norm{G}_F\big)\alpha \Big)\geq 1-n^{-M}, \vspace*{-1mm}
\end{equation*}
}%
for all $b^*$ and $\alpha$ such that $\sigma_{\max}(P^*)\leq b^*$ and $40(\theta^*_v+\theta^*_0)\sqrt{N_0/N}\leq\alpha\leq 40(\theta^*_v+\theta^*_0)$. 
\end{theorem}
\begin{proof}
Through the change of variables $\Delta\triangleq P-P^*$, we can write \eqref{eq:convexProgram} in terms of $\Delta$: 
{\small
\begin{equation*}
\hat{\Delta} = \argmin \{F(\Delta)\triangleq \fprod{S,\Delta+P^*}-\log\det(\Delta+P^*)+\alpha\fprod{G,|\Delta+P^*|}: \Delta\in\cF\},
\end{equation*}}%
where $\cF\triangleq \{\Delta\in\reals^{n\times n}:\ \Delta=\Delta^\top, \ a^*\mb{I}\preceq \Delta+P^* \preceq b^*\mb{I}\}$. Note that $\hat{\Delta}=\hat{P}-P^*$. Define $g(\Delta)\triangleq -\log\det(\Delta+P^*)$ on $\cF$. $g(.)$ is strongly convex over $\cF$ with modulus $1/{b^*}^2$; hence, for any $\Delta\in\cF$, it follows that
$g(\Delta)-g(\mathbf{0}) \geq -\fprod{{P^*}^{-1},\Delta}+\frac{1}{2{b^*}^2}\norm{\Delta}_F^2$.
Let $H(\Delta)\triangleq F(\Delta)-F(\mathbf{0})$ and $S_{\Delta}\triangleq \{\Delta\in\cF: \norm{\Delta}_F > 2{b^*}^2\big(n+\norm{G}_F\big)\alpha\}$.

Under probability event $\Omega=\{|S_{ij}-C^*_{ij}|\leq\alpha,\ \forall(i,j)\in\cI\times\cI\}$, for any $\Delta\in S_{\Delta}\subset\cF$, we have
{\small
\begin{align}
H(\Delta) &\geq \fprod{S,\Delta}-\fprod{{P^*}^{-1},\Delta}+\frac{1}{2{b^*}^2}\norm{\Delta}_F^2+\alpha\fprod{G,|\Delta+P^*|}-\alpha\fprod{G,|P^*|}~\label{eq:probBoundineq1}\\
          &\geq \frac{1}{2{b^*}^2}\norm{\Delta}_F^2+\fprod{\Delta,S-C^*}-\alpha\fprod{G,|\Delta|}\\
          &\geq \frac{1}{2{b^*}^2}\norm{\Delta}_F^2-\alpha\big(n+\norm{G}_F\big)\norm{\Delta}_F> 0,
          \vspace*{-4mm}
\end{align}
}%
where the second inequality follows from the triangle inequality, the third one holds under the probability event $\Omega$ and follows from the Cauchy-Schwarz inequality, and the final strict one follows from the definition of $S_{\Delta}$. Since $F(\mathbf{0})$ is a constant, $\hat{\Delta} = \argmin \{H(\Delta): \Delta\in\cF\}$. Hence, $H(\hat{\Delta})\leq H(\mathbf{0})=0$. Therefore, 
$\hat{\Delta}\not\in S_{\Delta}$ under the probability event $\Omega$. It is important to note that $\hat{\Delta}\in\cF$;
hence, $\hat{\Delta}\not\in S_{\Delta}$ implies $\norm{\hat{\Delta}}_F\leq 2{b^*}^2\big(n+\norm{G}_F\big)\alpha$ whenever the probability event $\Omega$ is true. Thus, 
{\small
\begin{eqnarray*}
\lefteqn{\mbox{Pr}\left(\norm{\hat{P}-P^*}_F\leq 2 {b^*}^2 (n+\norm{G}_F) \alpha\right)}\\
& &\geq \mbox{Pr}\left(|S_{ij}-C^*_{ij}|\leq\alpha,\  \forall(i,j)\in\cI\times\cI\right)  \\
& &= 1-\mbox{Pr}\left(\max_{i,j\in\cI}|S_{ij}-C^*_{ij}|>\alpha\right)\geq 1-\sum_{i,j\in\cI}\mbox{Pr}\left(|S_{ij}-C^*_{ij}|>\alpha\right).
\end{eqnarray*}}%
Recall that $S=\frac{1}{N}\sum_{r=1}^N\by^{(r)}{\by^{(r)}}^\top$ and $\by^{(r)}=[y_i^{(r)}]_{i\in\cI}$ for $r=1,\ldots,N$. Since $y_i^{(r)}/\sqrt{C^*_{ii}}\sim\cN(\mu=0,\sigma=1)$, i.e., standard normal, for all $i$ and $r$,
Lemma~1 in~\citep{ravikumar2011high} implies
$\mbox{Pr}\left(|S_{ij}-C^*_{ij}|>\alpha \right)\leq B_\alpha$
for any $(i,j)\in\cI\times\cI$ and $\alpha\in\left(0,40\max_i C^*_{ii}\right)$, where \[B_\alpha\triangleq 4\exp\left(\frac{-N}{2}\left(\frac{\alpha}{40\max_i C^*_{ii}}\right)^2\right).\]
Hence, given any $M>0$, by requiring 
$N \geq \left(\frac{40\max_i C^*_{ii}}{\alpha}\right)^2 N_0$, we get $B_\alpha\leq \frac{1}{n^2}n^{-M}$. 
Thus, for any $N\geq N_0$, we have $\sum_{i,j\in\cI}\mbox{Pr}\left(|S_{ij}-C^*_{ij}|>\alpha\right)\leq n^{-M}$ for all $40(\theta^*_v+\theta^*_0)\sqrt{\frac{N_0}{N}}\leq\alpha\leq 40(\theta^*_v+\theta^*_0)$ since $C^*_{ii}=\theta^*_v+\theta^*_0$ for all $i$; and this completes the proof.
\end{proof}

After setting $\alpha$ to its lower bound in Theorem~\ref{thm:statAnalysis}, our bound conforms with the one provided by 
\cite{Rothman2008sparse} (Theorem 1) which states $\norm{\hat{P}_\alpha-P^*}_F=\cO\left(\sqrt{\frac{\mathrm{card}(P^*)\log n}{N}}\right)$, where $\hat{P}_\alpha$ is the solution to \eqref{eq:CSConvex}. 
Furthermore, our proof uses the strong convexity property of the objective function, and does not require second-order differentiability of the objective as in~\cite{Rothman2008sparse}; hence, it is more general.

\subsection{Learning the hyperparameters}
To ease the notational burden in the proofs, we assume the nugget parameter $\theta^*_0=0$ in the rest of the discussion. Hence, $\th^*=[{\boldsymbol{\theta}_{\rho}^*}^\top,\theta_v^*]^\top\in\reals^{q+1}$.

Recall that 
$\mb{r}$ in Theorem~\ref{thm:innerOpt} and Corollary~\ref{cor:innerOptNoNugget} {actually} depends on $\boldsymbol{\theta}_{\rho}$, i.e., $\mb{r}(\boldsymbol{\theta}_{\rho})$, where $\boldsymbol{\theta}_{\rho}$ is the decision variable of the outer optimization in \eqref{eq:app-nonconvexOpt}. 
After 
substituting the optimal solution from Corollary~\ref{cor:innerOptNoNugget}, the objective of the outer problem in \eqref{eq:app-nonconvexOpt} 
is obtained by evaluating 
{\small
\begin{equation}\label{eq:objOuterNoNugget}
f(\boldsymbol{\theta}_{\rho};\mb{c}) = \frac{1}{2}\norm{\max\left\{0,~\frac{\mb{r}(\boldsymbol{\theta}_{\rho})^\top\mb{c}}{\norm{\mb{r}(\boldsymbol{\theta}_{\rho})}^2}\right\}\mb{r}(\boldsymbol{\theta}_{\rho})-\mb{c}}^2,
\end{equation}
}
at $\mb{c}=\hat{\mb{c}}$, where $\mb{r}(\boldsymbol{\theta}_{\rho})$ and $\hat{\mb{c}}$ are defined as in \eqref{eq:appB-vectorForm}. 
\begin{lemma}\label{strongConvexityOuterOpt}
Let $\th^*=[{\th_{\rho}^*}^\top,\theta_v^*]^\top\in\mathrm{int}(\Theta_\rho)\times\reals_{++}$ be the true 
parameters, 
and $\bc^*=\theta_v^*\br(\th_{\rho}^*)\in\mR^{n^2}$ be the true covariance matrix $C^*$ in vector form, i.e., 
$\bc^*\in\reals^{n^2}$ such that $\bc^*_{ij}=C^*_{ij}$ for $(i,j)\in\cI\times\cI$.
{Suppose the correlation function $r(\mb{x},\mb{x}',\th_\rho)$ is twice continuously differentiable in $\th_\rho$ over $\Theta_\rho$ for all $\mb{x},\mb{x}'\in\cX$}, then there exists $\gamma^*>0$ such that 
$\grad^2 f(\th_\rho^*;\bc^*)\succeq \gamma^*\mathbf{I}$ if and only if 
$\{\br(\th_{\rho}^*),\br'_1(\th_{\rho}^*),\ldots,\br'_q(\th_{\rho}^*)\}\subset\reals^{n^2}$ are linearly independent, where $\br'_j(\th_{\rho}^*)$ is the $j$-th column of $\bJ\br(\th_{\rho}^*)$, i.e., the Jacobian of $\br:\reals^q\rightarrow\reals^{n^2}$ at $\th_{\rho}^*$.
\end{lemma}
\vspace*{-1mm}
\begin{proof}
Recall $\mb{r}:\reals^q\rightarrow\reals^{n^2}$ such that $\mb{r}_{ij}(\th_\rho)=r(\bx_i,\bx_j,\th_\rho)$ for $(i,j)\in\cI\times\cI$. Let $\th_\rho=[\xi_1,\ldots,\xi_q]^\top$. Define $g:\reals^q\times\reals^{n^2}\rightarrow\reals$ such that 
{\small
\begin{equation}
\label{eq:g-def}
g(\th_\rho;\bc)\triangleq \tfrac{1}{2}\norm{\left(\frac{\br(\th_\rho)^\top\bc}{\norm{\br(\th_\rho)}^2}\right)\br(\th_\rho)-\bc}^2.
\end{equation}}%
Note the objective of the outer problem in \eqref{eq:app-nonconvexOpt}, i.e., $f(\boldsymbol{\theta}_{\rho};\hat{\mb{c}})$ defined in \eqref{eq:objOuterNoNugget}, is equal to $g(\th_\rho;\hat{\bc})$ whenever $\mb{r}(\boldsymbol{\theta}_{\rho})^\top\hat{\mb{c}}\geq 0$. Let $\mb{z}:\reals^q\times\reals^{n^2}\rightarrow\reals^{n^2}$ such that $\bz(\th_\rho;\bc)\triangleq \left(\frac{\br(\th_\rho)^\top\bc}{\norm{\br(\th_\rho)}^2}\right)\br(\th_\rho)-\bc$ and define $p(\bx)\triangleq \tfrac{1}{2}\norm{\bx}^2$, where $\th_\rho$ and $\bc$ are the variable and parameter vectors of function $\bz$, respectively. Hence, 
$g(\th_{\rho};\bc)=p(\bz(\th_{\rho};\bc))$. In the rest, all the derivatives for $\bz$ and $g$ are written with respect to $\th_\rho$ only, not $\bc$. Applying the chain rule we obtain:
\begin{small}
\begin{align}
\label{eq:grad-g}
\grad g(\th_{\rho};\bc)=\bJ\bz(\th_{\rho};\bc)^\top\grad p\left(\bz(\th_{\rho};\bc)\right)= \bJ\bz(\th_{\rho};\bc)^\top\bz(\th_{\rho};\bc), 
\end{align}
\end{small}

\noindent where $\bJ\bz(\th_{\rho};\bc)\in\mR^{n^2\times q}$ denotes the Jacobian matrix, i.e., for $(i,j)\in\cI\times\cI$, and $k\in\{1,\ldots,q\}$, $\left(\bJ\bz(\th_{\rho};\bc)\right)_{(i,j),k}=\frac{\partial}{\partial \xi_k}\bz_{ij}(\th_{\rho};\bc)$. Let $\bH\bz(\th_{\rho};\bc)\in\mR^{n^2q\times q}$ be the matrix of second-order derivatives of $\bz(\th_\rho;\bc)$, i.e., $\bH\bz(\th_{\rho};\bc)=\left[w_{k_1,k_2}\right]_{k_1,k_2\in\{1,\ldots,q\}}$ and $w_{k_1,k_2}=\frac{\partial^2}{\partial\xi_{k_1}\partial\xi_{k_2}}\bz(\th_{\rho};\bc)$ $\in\reals^{n^2}$. Let $\bI_q$ denote $q\times q$ identity matrix. 
Then the Hessian of $g$ can be written as follows: 
\begin{small}
\begin{align}
\grad^2 g(\th_{\rho};\bc) &= \bH\bz(\th_{\rho};\bc)^\top\Big(\bI_q\otimes\grad p\left(\bz(\th_{\rho};\bc)\right)\Big)+\bJ\bz(\th_{\rho};\bc)^\top\grad^2p\left(\bz(\th_{\rho};\bc)\right)\bJ\bz(\th_{\rho};\bc) \nonumber \\
 &=\bH\bz(\th_{\rho};\bc)^\top\Big(\bI_q\otimes\bz(\th_{\rho};\bc)\Big)+\bJ\bz(\th_{\rho};\bc)^\top\bJ\bz(\th_{\rho};\bc). \label{eq:hessian_g}
\end{align}
\end{small}

\noindent Note $\bc^*=\theta^*_v\mb{r}(\boldsymbol{\theta}^*_{\rho})$; 
hence, $\theta^*_v=\frac{\mb{r}(\th^*_{\rho})^\top\mb{c}^*}{\norm{\mb{r}(\th^*_{\rho})}^2}$. Therefore, $\bz(\th_{\rho}^*;\bc^*)=\mathbf{0}$, and the definition of $g$ in \eqref{eq:g-def} implies that $g(\th_{\rho}^*;\bc^*)=0$, and $\grad g(\th_{\rho}^*;\bc^*)=\mathbf{0}$. Thus, 
\begin{small}
\begin{align}\label{eq:gamma_star}
\grad^2 g(\th_{\rho}^*;\bc^*) = \bJ\bz(\th_{\rho}^*;\bc^*)^\top\bJ\bz(\th_{\rho}^*;\bc^*),
\end{align}
\end{small}

\noindent which is clearly a positive semidefinite matrix. Next, we investigate the condition under which $\grad^2 g(\th_{\rho}^*;\bc^*)$ is \emph{positive definite}. $\bJ\bz(\th_{\rho};\bc)$ can be explicitly written as \vspace*{-4mm}
\begin{small}
\begin{equation}
\label{eq:Jz}
\bJ\bz(\th_{\rho};\bc)=\frac{\br(\th_{\rho})}{\norm{\br(\th_{\rho})}^2}\left(\bc-2\frac{\br(\th_{\rho})^\top\bc}{\norm{\br(\th_{\rho})}^2}~\br(\th_{\rho})\right)^\top\bJ\br(\th_{\rho})
+\frac{\br(\th_{\rho})^\top\bc}{\norm{\br(\th_{\rho})}^2}~\bJ\br(\th_{\rho}). \vspace*{-3mm}
\end{equation}
\end{small}

\noindent Plugging in $\th_{\rho}^*$ and $\bc^*$, and using $\bc^*=\theta^*_v\mb{r}(\boldsymbol{\theta}^*_{\rho})$ and $\theta^*_v=\frac{\mb{r}(\th^*_{\rho})^\top\mb{c}^*}{\norm{\mb{r}(\th^*_{\rho})}^2}$, we get \vspace*{-4mm}
\begin{small}
\begin{align*}
\bJ\bz(\th_{\rho}^*;\bc^*)&=\frac{\br(\th_{\rho}^*)}{\norm{\br(\th_{\rho}^*)}^2}\left(\bc^*-2\theta_v^*\br(\th_{\rho}^*)\right)^\top\bJ\br(\th_{\rho}^*)+\theta_v^*\bJ\br(\th_{\rho}^*)\\
 &=-\theta_v^*~\frac{\br(\th_{\rho}^*)}{\norm{\br(\th_{\rho}^*)}^2}\br(\th_{\rho}^*)^\top\bJ\br(\th_{\rho}^*)+\theta_v^*\bJ\br(\th_{\rho}^*)\\
 &=\theta_v^*\left(\bI-\left(\frac{\br(\th_{\rho}^*)}{\norm{\br(\th_{\rho}^*)}}\right)\left(\frac{\br(\th_{\rho}^*)}{\norm{\br(\th_{\rho}^*)}}\right)^\top\right)\bJ\br(\th_{\rho}^*).
 \vspace*{-4mm}
\end{align*}
\end{small}

\noindent Let the $q$ columns of the Jacobian matrix $\bJ\bz(\th_{\rho}^*;\bc^*)$ be denoted by $[\bz'_1,...,\bz'_q]$, and the $q$ columns of $\bJ\br(\th_{\rho}^*)$ be denoted by $[\br'_1,...,\br'_q]$. Define $\br^*\triangleq \br(\th_{\rho}^*)$ and $\tilde{\br}\triangleq \br^*/\norm{\br^*}$, then we have 
{\small
\begin{equation}\label{eq:bz_j}
\bz'_j = \theta_v^*(\bI-\tilde{\br}\tilde{\br}^\top)\br'_j,\quad \forall~j=1,...,q.
\end{equation}}%
For $\grad^2 g(\th_{\rho}^*;\bc^*)$ to be positive definite, the matrix $\bJ\bz(\th_{\rho}^*;\bc^*)$ should be full rank, i.e., $\{\bz'_1,...,\bz'_q\}$ should be linearly independent. 
Note $\{\bz'_1,...,\bz'_q\}$ are linearly dependent if and only if there exists $\b\neq\mathbf{0}$ such that $\sum_{j=1}^q\beta_j\bz'_j=\mathbf{0}$, which is equivalent to the condition 
$\sum_{j=1}^q\beta_j\br'_j=\bar{\b}~\tilde{\br}$ due to \eqref{eq:bz_j}, where $\bar{\b}\triangleq \tilde{\br}^\top\left(\sum_{j=1}^q\beta_j\br'_j\right)$. 
If $\bar{\b}=0$, then the set of vectors $\{\br'_1,\ldots,\br'_q\}$ are linearly dependent; otherwise, $\bar{\b}\neq 0$ implies that $\tilde{\br}=\sum_{j=1}^q(\beta_j/\bar{\b})\br'_j$; thus, $\{\br,\br'_1,\ldots,\br'_q\}$ are linearly dependent. Therefore, $\{\bz'_1,...,\bz'_q\}$ are linearly independent if and only if $\{\br,\br'_1,...,\br'_q\}$ are linearly independent. Finally, the function
{\small
$$s(\th_{\rho},\bc)\triangleq \frac{\br(\th_{\rho})^\top\bc}{\norm{\br(\th_{\rho})}^2}$$}%
is continuous in $(\th_{\rho},\bc)$; hence, the preimage $s^{-1}(\mR_{++})$ is an open set. Moreover, $s(\th_{\rho}^*,\bc^*)=\theta^*_v>0$; hence, $(\th_{\rho}^*,\bc^*)\in s^{-1}(\mR_{++})$. Therefore, there exists $\delta_1>0$ such that $B_{\delta_1}(\th_{\rho}^*,\bc^*)$, the open ball around $(\th_{\rho}^*,\bc^*)$ with radius $\delta_1$, satisfies $B_{\delta_1}(\th_{\rho}^*,\bc^*)\subseteq s^{-1}(\mR_{++})$, and the objective of the outer problem in \eqref{eq:app-nonconvexOpt}, i.e.,  $f(\th_\rho;\bc)$ defined in \eqref{eq:objOuterNoNugget}, is equal to $g(\th_\rho;\bc)$ on $B_{\delta_1}(\th_{\rho}^*,\bc^*)$. Thus, $\grad^2 f(\th_{\rho}^*;\bc^*)$ exists, and it satisfies $\grad^2 f(\th_{\rho}^*;\bc^*)=\grad^2 g(\th_{\rho}^*;\bc^*)\succ\gamma^*\mathbf{I}$.
\end{proof}
\begin{remark}
We would like to comment on the linear independence condition stated in Lemma~\ref{strongConvexityOuterOpt}. 
For instance, consider
the \emph{anisotropic exponential correlation} function $r(\bx,\bx',\th_\rho)=\exp\big(-(\bx-\bx')^\top\diag(\th_\rho)(\bx-\bx')\big)$, where $q=d$, and $\Theta_\rho=\reals^d_{+}$. Let $\cX=[-\beta,\beta]^d$ for some $\beta>0$, and suppose $\{\bx_i\}_{i\in\cI}$ is a set of independent identically distributed \emph{uniform} random samples inside $\cX$. Then it can be easily shown that for the anisotropic exponential correlation function, the condition in Lemma~\ref{strongConvexityOuterOpt} holds with probability 1, i.e., $\{\br(\th_{\rho}^*),\br'_1(\th_{\rho}^*),\ldots,\br'_q(\th_{\rho}^*)\}$ are linearly independent w.p.~1.
\end{remark}

The next result shows the convergence of the SPS estimator as the number of samples per location, $N$, increases. \vspace*{-1mm}
\begin{theorem}\label{thm:secondstage}
Let $\th^*=[{\th_{\rho}^*}^\top,\theta_v^*]^\top\in\mathrm{int}(\Theta_\rho)\times\reals_{++}$ be the true 
parameters.
Suppose $\br:\reals^q\rightarrow\reals^{n^2}$ is twice continuously differentiable, and vectors in $\{\br(\th_{\rho}^*),{\br'_1}(\th_{\rho}^*),\ldots,{\br'_q}(\th_{\rho}^*)\}$ are linearly independent. For any given $M>0$ and $N\geq N_0\triangleq\left\lceil2 (M+2)\log n+\log 16\right\rceil$,
let $\hat{\th}^{(N)}=[\hat{\th}_{\rho}^\top,\hat{\theta}_v]^\top$ be the SPS estimator of $\th^*$, i.e., $\hat{\th}_{\rho}\in\argmin_{\th_\rho\in\Theta_\rho} f(\boldsymbol{\theta}_{\rho};\hat{\mb{c}})$, and $\hat{\theta}_v$ be computed as in Corollary~\ref{cor:innerOptNoNugget}.  
Then for any $\epsilon>0$, there exists $N\geq N_0$ satisfying $N=\cO(N_0/\epsilon^2)$ 
such that setting $\alpha=40\theta_v^*\sqrt{\frac{N_0}{N}}$ in \eqref{eq:convexProgram} implies $\norm{\hat{\th}^{(N)}-\th^*}\leq\epsilon$ with probability at least $1-n^{-M}$; moreover, the STAGE-II function $f(\cdot;\hat{\mb{c}})$ is strongly convex around 
$\hat{\th}_{\rho}$.
\end{theorem}
\vspace*{-2mm}
\begin{proof}
From the hypothesis, $\{\br(\th_{\rho}^*),{\br'_1}(\th_{\rho}^*),...,{\br'_q}(\th_{\rho}^*)\}$ are linearly independent; hence, Lemma~\ref{strongConvexityOuterOpt} implies there exists $\gamma^*>0$ such that $\grad^2 g(\th_\rho^*;\bc^*)\succeq\gamma^*\mathbf{I}$ for $g$ defined in \eqref{eq:g-def} -- throughout the proof, all the derivatives of $g$ are written with respect to $\th_\rho$ only, not $\bc$. Recall from the proof of Lemma~\ref{strongConvexityOuterOpt} that the function $s(\th_{\rho},\bc)\triangleq \frac{\br(\th_{\rho})^\top\bc}{\norm{\br(\th_{\rho})}^2}$ is continuous in $(\th_{\rho},\bc)$;
therefore, there exists $\delta_1>0$ such that $B_{\delta_1}(\th_{\rho}^*,\bc^*)\subseteq s^{-1}(\mR_{++})$. Hence, the objective of the outer problem in \eqref{eq:app-nonconvexOpt}, i.e.,  $f(\th_\rho;\bc)$ defined in \eqref{eq:objOuterNoNugget}, is equal to $g(\th_\rho;\bc)$ on $B_{\delta_1}(\th_{\rho}^*,\bc^*)$. Moreover, since $\th^*_\rho\in\mathrm{int}(\Theta_\rho)$, $\delta_1$ can be chosen to satisfy {$B_{\delta_1}(\th^*_\rho)\subset\mathrm{int}(\Theta_\rho)$}. 

Since $\br(\th_\rho)$ is assumed to be twice continuously differentiable in $\th_\rho$, it follows from \eqref{eq:hessian_g} that $\grad^2 g(\th_\rho;\bc)$ is continuous in $(\th_\rho,\bc)$ on $B_{\delta_1}(\th_{\rho}^*,\bc^*)$. Moreover, eigenvalues of a matrix are continuous functions of matrix entries; hence, $\lambda_{\min}\left(\grad^2 g(\th_\rho;\bc)\right)$ is continuous in $(\th_\rho,\bc)$ on $B_{\delta_1}(\th_{\rho}^*,\bc^*)$ as well. Thus, $f$ is strongly convex around $(\th_{\rho}^*,\bc^*)$. Indeed,
there exists $\delta_2>0$ such that $\delta_2\leq\delta_1$ and $\grad^2 g(\th_\rho;\bc)\succ\frac{\gamma^*}{2}\mathbf{I}$ for all $(\th_{\rho},\bc)\in B_{\delta_2}(\th_{\rho}^*,\bc^*)$. 
Define
$$\cC\triangleq \{\bc:~\norm{\bc-\bc^*}\leq\frac{\delta_2}{\sqrt{2}}\},\qquad \Theta_{\rho}'\triangleq \{\th_\rho\in\Theta_\rho:\ \norm{\th_\rho-\th_\rho^*}\leq\frac{\delta_2}{\sqrt{2}}\};$$
and for all $\bc\in \cC$, let
$$\th_{\rho}(\bc)\triangleq \argmin \{g(\th_{\rho};\bc):\th_{\rho}\in\Theta_{\rho}'\}$$ be the unique minimizer as $g(\cdot;{\bc})$ is strongly convex in $\th_\rho$ over $\Theta_{\rho}'$ for $\bc\in \cC$. 
Furthermore, since $\Theta_{\rho}'$ is a convex compact set and $g(\th_{\rho};\bc)$ is jointly continuous in $(\th_\rho,\bc)$ on $\Theta_{\rho}'\times\cC$, by Berge's Maximum Theorem~\citep{ok2007real}, $\th_{\rho}(\bc)$ is continuous at $\bc^*$ and $\th_{\rho}(\bc^*)=\th_{\rho}^*$. Hence, 
given any $0<\eta<\frac{\delta_2}{\sqrt{2}}$, there exists $\delta(\eta)>0$ such that $\delta(\eta)\leq\frac{\delta_2}{\sqrt{2}}$, and $\norm{\th_{\rho}(\bc)-\th_{\rho}^*}\leq\eta$ for all $\norm{\bc-\bc^*}\leq\delta(\eta)$. 
{It follows from Theorem~\ref{thm:statAnalysis} that} by setting an appropriate $\alpha(\eta)$ in STAGE-I problem \eqref{eq:convexProgram}, it is guaranteed with high probability that $\norm{\hat{\bc}-\bc^*}\leq\delta(\eta)$ -- we will revisit this claim at the end. 
Thus, $\norm{\th_{\rho}(\hat{\bc})-\th_{\rho}^*}\leq\eta<\frac{\delta_2}{\sqrt{2}}$, which implies that $\th_{\rho}(\hat{\bc})=\argmin\{g(\th_{\rho};\hat{\bc}):\ \th_{\rho}\in\Theta_\rho\}$, i.e., it is equal to the solution to the outer problem in \eqref{eq:app-nonconvexOpt} of STAGE-II: $\hat{\th}_{\rho}=\th_{\rho}(\hat{\bc})$.
Hence, it follows that $\norm{\hat{\th}_\rho-\th^*_\rho}^2+\norm{\hat{\bc}-\bc^*}^2<\delta_2^2\leq\delta_1^2$, which implies that $(\hat{\th}_\rho,\hat{\bc})\in B_{\delta_1}(\th_{\rho}^*,\bc^*)\subseteq s^{-1}(\mR_{++})$; and since $\hat{\theta}_v
=\max\{0,s(\hat{\th}_{\rho},\hat{\bc})\}$, we also have $\hat{\theta}_v=s(\hat{\th}_{\rho},\hat{\bc})>0$. Moreover, $\hat{\th}_\rho\in B_{\delta_1}(\th_\rho^*)$ implies $\hat{\th}_\rho\in\mathrm{int}(\Theta_\rho)$. Note that $\hat{\th}_\rho\in\Theta_\rho'$ and $\hat{\bc}\in\cC$; hence, $g(\cdot;\hat{\bc})$ is strongly convex at $\hat{\th}_{\rho}$ with modulus $\frac{\gamma^*}{2}$.

Next we establish a relation between $\delta(\eta)$ and $\eta$ 
by showing $\hat{\th}_{\rho}(\bc)$ is Lipschitz 
around $\bc^*$. Since, for $\bc\in \cC$, $g(\cdot;{\bc})$ is strongly convex in $\th_\rho$ over $\Theta_{\rho}'$ with convexity modulus $\gamma^*/2$, for any $\bc\in \cC$, and $\th_\rho^i\in\Theta_\rho'$ for $i=1,2$, we have 
$$\fprod{\th_\rho^2-\th_\rho^1,~\grad g(\th_\rho^2;\bc)-\grad g(\th_\rho^1;\bc)}\geq\frac{\gamma^*}{2}\norm{\th_\rho^2-\th_\rho^1}^2.$$

\noindent Suppose $\bc^1,\bc^2\in \cC$. Since $\th_\rho(\bc^i)=\argmin\{g(\th_\rho;\bc^i):\ \th_\rho\in\Theta_\rho'\}$ for $i=1,2$, it follows from the first-order optimality conditions that 
{\small
\begin{equation}
\label{eq:opt_cond-g}
\fprod{\th_\rho-\th_\rho(\bc^i),~\grad g(\th_\rho(\bc^i);\bc^i)}\geq 0,\quad \forall~\th_\rho\in\Theta_\rho', \hbox{ and } i=1,2. \vspace*{-2mm}
\end{equation}}%
Strong convexity and \eqref{eq:opt_cond-g} imply $\fprod{\th_\rho(\bc^2)-\th_\rho(\bc^1),~\grad g(\th_\rho(\bc^2);\bc^1)}\geq \frac{\gamma^*}{2}\norm{\th_\rho(\bc^2)-\th_\rho(\bc^1)}^2$. Adding and subtracting $\grad g(\th_\rho(\bc^2);\bc^2)$, and using \eqref{eq:opt_cond-g} again, we get $\frac{\gamma^*}{2}\norm{\th_\rho(\bc^2)-\th_\rho(\bc^1)}^2\leq \fprod{\th_\rho(\bc^2)-\th_\rho(\bc^1),~\grad g(\th_\rho(\bc^2);\bc^1)-\grad g(\th_\rho(\bc^2);\bc^2)}$. Thus, from Cauchy-Schwarz, \vspace*{-2mm}
{\small
\begin{equation}
\label{eq:Lipschitz-g-keyineq}
\norm{\grad g(\th_\rho(\bc^2);\bc^2)-\grad g(\th_\rho(\bc^2);\bc^1)}\geq \frac{\gamma^*}{2}\norm{\th_\rho(\bc^2)-\th_\rho(\bc^1)}. \vspace*{-2mm}
\end{equation}}%
Moreover, \eqref{eq:grad-g} and \eqref{eq:Jz} imply that \vspace*{-2mm}
{\small
\begin{align}
\label{eq:grad-g-explicit}
\grad g(\th_\rho;\bc)&=\frac{1}{\norm{\br(\th_\rho)}^2}\bJ\br(\th_\rho)^\top v(\th_\rho;\bc),\\
v(\th_\rho;\bc)&\triangleq \br(\th_\rho)^\top(\bc-\bz(\th_\rho;\bc))\bz(\th_\rho;\bc)-\frac{(\br(\th_\rho)^\top\bc)(\br(\th_\rho)^\top\bz(\th_\rho;\bc))}{\norm{\br(\th_\rho)}^2}\br(\th_\rho), \nonumber\\
&= \left(\frac{1}{\norm{\br(\th_\rho)}^2}\br(\th_\rho)\br(\th_\rho)^\top-\mathbf{I}\right)\bc\bc^\top\br(\th_\rho),\nonumber
\vspace*{-2mm}
\end{align}}%
where $\mb{z}:\reals^q\times\reals^{n^2}\rightarrow\reals^{n^2}$ is defined as in the proof of Lemma~\ref{strongConvexityOuterOpt}, i.e., $\bz(\th_\rho;\bc)\triangleq \left(\frac{\br(\th_\rho)^\top\bc}{\norm{\br(\th_\rho)}^2}\right)\br(\th_\rho)-\bc$. 
Observe $\norm{\mathbf{I}-\frac{1}{\norm{\br(\th_\rho)}^2}\br(\th_\rho)\br(\th_\rho)^\top}_2=1$ {and $\norm{\br(\th_\rho)}\geq\sqrt{n}$}; therefore, given $\bc^1,\bc^2\in C$, we have
\vspace*{-2mm}
{\small
\begin{align}
\norm{\grad g(\th_\rho;\bc^2)-\grad g(\th_\rho;\bc^1)}
&=\frac{1}{\norm{\br(\th_\rho)}^2}\norm{\bJ\br(\th_\rho)^\top\left(\mathbf{I}-\frac{1}{\norm{\br(\th_\rho)}^2}\br(\th_\rho)\br(\th_\rho)^\top\right)(\bc^1{\bc^1}^\top-\bc^2{\bc^2}^\top)\br(\th_\rho)},\nonumber\\
&\leq \frac{1}{\sqrt{n}}~\norm{\bJ\br(\th_\rho)}_2~(\norm{\bc^1}+\norm{\bc^2})\norm{\bc^2-\bc^1}, \label{eq:Lipschitz-g-final-1}
\vspace*{-2mm}
\end{align}}%
where \eqref{eq:Lipschitz-g-final-1} follows from $\norm{\bc^1{\bc^1}^\top-\bc^2{\bc^2}^\top}_2\leq(\norm{\bc^1}+\norm{\bc^2})\norm{\bc^2-\bc^1}$. Since $\th^*_\rho=\th_\rho(\bc^*)$ and $\hat{\th}_\rho=\th_\rho(\hat{\bc})$, by setting {$\bc^1=\hat{\bc}$ and $\bc^2=\bc^*$} within \eqref{eq:Lipschitz-g-keyineq} and \eqref{eq:Lipschitz-g-final-1}, we get \vspace*{-2mm}
{\small
\begin{equation}
\label{eq:Lipschitz-g-final-ineq}
\norm{\hat{\th}_\rho-\th^*_\rho}\leq\frac{2}{\gamma^*\sqrt{n}}~{\norm{\bJ\br(\th^*_\rho)}_2}~\left(2\norm{\bc^*}+\frac{\delta_2}{\sqrt{2}}\right)\norm{\hat{\bc}-\bc^*}, \vspace*{-2mm}
\end{equation}}%
where we have used the fact that $\norm{\hat{\bc}-\bc^*}\leq\frac{\delta_2}{\sqrt{2}}$. Thus, given any $0<\eta<\frac{\delta_2}{\sqrt{2}}$, for $\delta(\eta)$ chosen as 
{\small
\begin{equation}
\label{eq:delta-eta}
\delta(\eta)\triangleq \min\left\{\frac{\gamma^*\sqrt{n}}{2{\norm{\bJ\br(\th^*_\rho)}_2}}(2\norm{\bc^*}+\frac{\delta_2}{\sqrt{2}})^{-1}\eta,~\frac{\delta_2}{\sqrt{2}}\right\}, \end{equation}
}%
it follows that $\norm{\hat{\bc}-\bc^*}\leq\delta(\eta)$ implies $\norm{\hat{\th}_\rho-\th^*_\rho}\leq\eta$.

Now, we show that $|\hat{\theta}_v-\theta_v^*|$ can be made arbitrarily small.
Let $\bt:\reals^q\rightarrow\reals^{n^2}$ such that $\bt(\th_\rho)=\br(\th_\rho)/\norm{\br(\th_\rho)}^2$; hence,
$\bJ\bt(\th_\rho)=\left(\mathbf{I}-\frac{2}{\norm{\br(\th_\rho)}^2}\br(\th_\rho)\br(\th_\rho)^\top\right)\frac{\bJ\br(\th_\rho)}{\norm{\br(\th_\rho)}^2}$.
Since $\br$ is twice continuously differentiable, there exists $0<U\in\reals$ such that
{\small
{
$$U\triangleq \max\{\norm{\bJ\br(\th_\rho)}_2:\ \norm{\th_\rho-\th^*_\rho}\leq\delta_2/\sqrt{2}\}.$$}}%
Therefore, $\norm{\bJ\bt(\th_\rho)}_2\leq U/\norm{\br(\th_\rho)}^2\leq U/n$ for any $\th_\rho$ belonging to the line segment connecting $\th^*_\rho$ and $\hat{\th}_\rho$.
Furthermore, $\bt(\hat{\th}_\rho)=\bt(\th^*_\rho)+\left(\int_0^1\bJ\bt(\th^*_\rho+\ell~(\hat{\th}_\rho-\th^*_\rho))~ d\ell\right)(\hat{\th}_\rho-\th^*_\rho)$; hence, $\norm{\bt(\hat{\th}_\rho)-\bt(\th^*_\rho)}\leq \frac{U}{n}\norm{\hat{\th}_\rho-\th^*_\rho}\leq\frac{U}{n}\eta$ as $\eta>0$ is chosen as $\eta<\delta_2/\sqrt{2}$.
Since $\hat{\theta}_v=s(\hat{\th}_{\rho},\hat{\bc})>0$ and $\theta^*_v=s(\th^*_{\rho},\bc^*)>0$, it follows 

\begin{small}
\begin{align*}
\vspace*{-5mm}
|\hat{\theta}_v-\theta_v^*| &= \left|\frac{\br(\hat{\th}_{\rho})^\top\hat{\bc}}{\norm{\br(\hat{\th}_{\rho})}^2}-\frac{\br(\th_{\rho}^*)^\top\bc^*}{\norm{\br(\th_{\rho}^*)}^2}\right|\\
 &= \left|\fprod{\frac{\br(\hat{\th}_{\rho})}{\norm{\br(\hat{\th}_{\rho})}^2}-\frac{\br(\th_{\rho}^*)}{\norm{\br(\th_{\rho}^*)}^2},\bc^*}+\fprod{\frac{\br(\hat{\th}_{\rho})}{\norm{\br(\hat{\th}_{\rho})}^2},\hat{\bc}-\bc^*}\right|\\
 &\leq \norm{\bt(\hat{\th}_\rho)-\bt(\th^*_\rho)}\norm{\bc^*}+\frac{1}{\norm{\br(\hat{\th}_\rho)}}\norm{\bc^*-\hat{\bc}}\\
 &\leq \frac{U}{n}\norm{\bc^*}\eta+\frac{1}{\sqrt{n}}\delta(\eta). 
 \vspace*{-4mm}
\end{align*}
\end{small}

\noindent Therefore, using the identity $(a+b)^2\leq2(a^2+b^2)$ for any $a,b\in\reals$, we get 
{\small
\begin{align*}
\norm{\th^*-\hat{\th}}^2 &= \norm{\th_{\rho}^*-\hat{\th}_{\rho}}^2+|\theta_v^*-\hat{\theta}_v|^2 \leq 
\left(\frac{2U^2}{n^2}\norm{\bc^*}^2+1\right)\eta^2+\frac{2}{n}\left(\delta(\eta)\right)^2. \vspace*{-4mm}
\end{align*}}%
From \eqref{eq:delta-eta} and $\norm{\bc^*}=\norm{C^*}_F\leq\Tr(C^*)\leq n\theta^*_v$, let 
{$\kappa\triangleq (\sqrt{2}U\theta^*_v+1)(2n\theta^*_v+\frac{\delta_2}{\sqrt{2}})+\frac{\gamma^*}{\sqrt{2}\norm{\bJ\br(\th^*_\rho)}_2}$}; hence, choosing $\eta_\epsilon\triangleq  \min\left\{ \frac{1}{\kappa}(2\norm{\bc^*}+\frac{\delta_2}{\sqrt{2}})~\epsilon,~\frac{\delta_2}{\sqrt{2}}\right\}$ implies that $\norm{\th^*-\hat{\th}}\leq\epsilon$. Thus, for all sufficiently small $\epsilon>0$, having $\norm{\hat{\bc}-\bc^*}\leq\delta_\epsilon$ implies that $\norm{\th^*-\hat{\th}}\leq\epsilon$, where {$\delta_\epsilon\triangleq \delta(\eta_\epsilon)=\frac{\gamma^*\sqrt{n}}{2\norm{\bJ\br(\th^*_\rho)}_2}~\frac{\epsilon}{\kappa}$}.
Next, according to Theorem~\ref{thm:statAnalysis}, given $\epsilon>0$ and $M>0$, for all $N$ such that {$\sqrt{N/N_0}\geq160\frac{\kappa}{\gamma^*}\norm{\bJ\br(\th^*_\rho)}_2\left(\frac{b^*}{a^*}\right)^2(n+\norm{G}_F)\theta^*_v \frac{1}{\epsilon}$}, choosing $\alpha=40\theta^*_v\sqrt{\frac{N_0}{N}}$ in \eqref{eq:convexProgram} guarantees 
\begin{small}
\begin{align}\label{eq:ineq_series}
\norm{\hat{\bc}-\bc^*} = \norm{\hat{C}-C^*}_F \leq \sqrt{n}\norm{\hat{C}-C^*}_2 \leq\frac{\sqrt{n}}{{a^*}^2}\norm{\hat{P}-P^*}_2 \leq\frac{\sqrt{n}}{{a^*}^2}\norm{\hat{P}-P^*}_F\leq\delta_\epsilon,
\vspace*{-4mm}
\end{align}
\end{small}%
with probability at least $1-n^{-M}$, where the equality follows from the definitions of $\bc^*$ and $\hat{\bc}$, the second inequality follows from $\hat{C}=\hat{P}^{-1}$, $C^*={P^*}^{-1}$, and the fact that $P\mapsto P^{-1}$ is Lipschitz continuous on $P\succeq a^*\mathbf{I}$. This completes the proof.
\end{proof}

\subsubsection{Tighter bounds through exploiting the decay property} \label{sec:new_sec}
In the GRF setting considered in this paper $\mathrm{card}(P^*)=n^2$; hence, the bound obtained in Theorem~\ref{thm:statAnalysis} may seem loose. On the other hand, Theorem~\ref{thm:decay} shows that the elements of $P^*$ exhibit a fast decay similar to the elements of the covariance matrix. Therefore, although $\mathrm{card}(P^*)=n^2$, significant amount of these elements are close to 0. In this section, we argue that through exploiting this property, the $\cO(1)$ constant in Theorem~\ref{thm:statAnalysis} can be significantly improved. More precisely, the main reason we get $\sqrt{n^2\log n/N}$ bound on $\norm{\hat{P}-P^*}_F$ is that the analysis provided for Theorem~\ref{thm:statAnalysis} is similar to \cite{Rothman2008sparse} (Theorem 1)  and it does not exploit the fast decay in the elements of $P^*$, leading us to $\sqrt{\frac{\mathrm{card}(P^*)\log n}{N}}$ and $\mathrm{card}(P^*)=n^2$. Instead in the new version in Corollary~\ref{cor:probBound4JaffardCov},  we were able to replace $\mathrm{card}(P^*)$ with the cardinality of some set $S_{\bar{\epsilon}}\subset\{(i,j)\in\cI\times\cI:\ |P^*_{ij}|\geq \bar{\epsilon}\}$, defined in \eqref{eq:def_S_e} for some $\bar{\epsilon}>0$. Clearly, $|S_{\bar{\epsilon}}|\leq \mathrm{card}(P^*)$ for all $\bar{\epsilon}\geq 0$.

The fast decay of the elements in inverse of the covariance matrix (Theorem~\ref{thm:decay}) is the motivation behind sparse approximation of the inverse covariance matrix in the STAGE-I of SPS as Jaffard's decay algebra clearly applies to the inverse covariance matrix. Proposed sparse approximation of $P^*$ leads to $|S_{\bar{\epsilon}}|$ term in the bound on $\norm{\hat{P}-P^*}_F$. More precisely, intuitively, given a reasonable threshold value $\bar{\epsilon}>0$, due to fast decay seen in the elements of $P^*$, one expects $|S_{\bar{\epsilon}}|\ll n^2$. Unfortunately, even for very simple deterministic designs such as a $d$-dimensional lattice, computing $|S_{\bar{\epsilon}}|$ in closed-form is a hard combinatorial problem -- indeed, we spend quite some time computing it in the closed form for lattice designs only to get some partial results. Therefore, we numerically investigated how $|S_{\bar{\epsilon}}|$ compares to $n$ for different values of $\bar{\epsilon}>0$ and it turns out that $|S_{\bar{\epsilon}}|$ behaves similar to $\cO(n)$, rather than $\cO(n^2)$ -- In Remark~\ref{rem:cor12}, we plotted $|S_{\bar{\epsilon}}|/n$ and compared it to $n^2/n$ for (squared) exponential and Matern covariance functions and for different values of $\bar{\epsilon}$ values ranging between $10^{-1}$ and $10^{-15}$.

In the rest of this section, suppose that the true covariance matrix $C^*$ belongs to the class $\cE_{\gamma}$ for some $\gamma>0$; hence, it follows from Theorem~\ref{thm:decay} that 
$P^*$ belongs to the class $\cE_{\gamma'}$ for some $\gamma'>0$. Thus, $\forall (i,j)\in\cI\times\cI$, $|P_{ij}^*|\leq K_{\gamma'}\exp(-\gamma' d(\bx_i,\bx_j))$, where $d(\bx_i,\bx_j)$ is the Euclidean distance between $\bx_i$ and $\bx_j$.
Given $\bar{\epsilon}>0$, we define
\begin{equation}
q(\bar{\epsilon})\triangleq\tfrac{1}{\gamma'}\log(K_{\gamma'}/\bar{\epsilon}).
\end{equation}
If $d(\bx_i,\bx_j)\geq q(\bar{\epsilon})$, then we have $|P_{ij}^*|\leq K_{\gamma'}\exp(-\gamma'd(\bx_i,\bx_j))\leq\bar{\epsilon}$. Next, we define
\begin{equation}\label{eq:def_S_e}
S_{\bar{\epsilon}}\triangleq \{(i,j)\in\cI\times\cI: d(\bx_i,\bx_j)\leq q(\bar{\epsilon})\} \subseteq  \{(i,j)\in\cI\times\cI: |P^*_{ij}|\geq \bar{\epsilon}\}
\end{equation}
and let $S_{\bar{\epsilon}}^c$ be its complement, i.e., $S_{\bar{\epsilon}}^c\triangleq\cI\times\cI\backslash S_{\bar{\epsilon}}$.

Next, in Corollary~\ref{cor:probBound4JaffardCov} we establish $\cO(\log(1/\bar{\epsilon})\sqrt{|S_{\bar{\epsilon}}|\log n/N})$ bound on $\norm{\hat{P}-P^*}_F$ and show that $N=\cO(1/\epsilon^2)$ rate result of Theorem~\ref{thm:secondstage} still holds for all $\epsilon>0$.

\begin{corollary}\label{cor:probBound4JaffardCov}
Let $\{\by^{(r)}\}_{r=1}^N\subset\reals^n$ be independent realizations of a GRF with zero-mean and stationary covariance function $c(\mb{x},\mb{x}',\boldsymbol{\theta}^*)$ observed over $n$ distinct locations $\{\bx_i\}_{i\in\cI}$ with $\cI\triangleq\{1,...,n\}$. Suppose there exists some $K_\gamma,\gamma>0$ such that $c(\mb{x},\mb{x}',\boldsymbol{\theta}^*)\leq K_\gamma {\rm exp}(-\gamma \norm{\bx-\bx'})$ for $\bx,\bx'\in\cX$.
Then, under the premise of Theorem~\ref{thm:secondstage}, for any given $\epsilon>0$ and $M>0$, there exists $\bar{\epsilon}>0$ and $N_{\epsilon}=\cO(N_0/\epsilon^2)$
such that with probability $1-n^{-M}$, we have
\begin{equation}\label{eq:probBoundJaffard}
\norm{\hat{P}-P^*}_F \leq \cO\Big(\log(1/\bar{\epsilon})\sqrt{|S_{\bar{\epsilon}}|\log n / N_\epsilon}\Big)=\cO(\epsilon),
\end{equation}
where $\hat{P}$ is the SPS estimator of $P^*$ computed as in \eqref{eq:convexProgram} with $\alpha=40\theta^*_v\sqrt{N_0/N_\epsilon}$. Moreover,  \eqref{eq:probBoundJaffard} implies that the STAGE-II solution satisfies $\norm{\hat{\th}^{(N_\epsilon)}-\th^*}\leq\epsilon$.
\end{corollary}

\begin{proof}
Given any $\Delta\in\reals^{n\times n}$, we define $\Delta_{S_{\bar{\epsilon}}}$ and $\Delta_{S_{\bar{\epsilon}}^c}$ as follows: $\Delta_{S_{\bar{\epsilon}}}(i,j)$ is equal to $\Delta(i,j)$ if $(i,j)\in S_{\bar{\epsilon}}$ and zero otherwise ($\Delta_{S_{\bar{\epsilon}}^c}$ is defined \emph{similarly}).

For any given $\bar{\epsilon}>0$, define $T\triangleq \{\Delta\in\cF: \norm{\Delta}_F > \alpha{b^*}^2(1+q(\bar{\epsilon})/G_{\min})|S_{\bar{\epsilon}}|^{1/2}+\Big(\big(\alpha{b^*}^2(1+q(\bar{\epsilon})/G_{\min})|S_{\bar{\epsilon}}|^{1/2}\big)^2+4\alpha{b^*}^2\norm{G}_{\infty}\sum_{(i,j)\in S_{\bar{\epsilon}}^c}|P^*_{ij}|\Big)^{1/2}\}$ where $\cF$ is defined as in the proof of Theorem~\ref{thm:statAnalysis}. From \eqref{eq:probBoundineq1}, for any $\bar{\epsilon}>0$ and $\Delta\in T$, we have
{\scriptsize
\begin{align}
H(\Delta) &\geq \frac{1}{2{b^*}^2}\norm{\Delta}_F^2+\fprod{\Delta,S-C^*}+\alpha\fprod{G,|\Delta+P^*|-|P^*|}, \nonumber\\
		& \geq \frac{1}{2{b^*}^2}\norm{\Delta}_F^2+\fprod{\Delta,S-C^*}
-\alpha\sum_{(i,j)\in S_{\bar{\epsilon}}}G_{ij}|\Delta_{ij}|+\alpha\sum_{(i,j)\in S_{\bar{\epsilon}}^c}G_{ij}(|\Delta_{ij}|-2|P^*_{ij}|),~\label{eq:corJaffard-2} \\	
		&  \geq \frac{1}{2{b^*}^2}\norm{\Delta}_F^2-\alpha\norm{\Delta_{S_{\bar{\epsilon}}}}_1-\alpha\norm{\Delta_{S_{\bar{\epsilon}^c}}}_1-\alpha\frac{q(\bar{\epsilon})}{G_{\min}}\norm{\Delta_{S_{\bar{\epsilon}}}}_1+\alpha \frac{G_{\min}}{G_{\min}}\norm{\Delta_{S_{\bar{\epsilon}}^c}}_1-2\alpha\sum_{(i,j)\in S_{\bar{\epsilon}}^c}G_{ij}|P^*_{ij}|,~\label{eq:corJaffard-3} \\
		& = \frac{1}{2{b^*}^2}\norm{\Delta}_F^2-\alpha(1+\frac{q(\bar{\epsilon})}{G_{\min}})\norm{\Delta_{S_{\bar{\epsilon}}}}_1
-2\alpha\sum_{(i,j)\in S_{\bar{\epsilon}}^c}G_{ij}|P^*_{ij}|,~\label{eq:corJaffard-4} \\
		& \geq \frac{1}{2{b^*}^2}\norm{\Delta}_F^2-\alpha(1+\frac{q(\bar{\epsilon})}{G_{\min}})|S_{\bar{\epsilon}}|^{1/2}\norm{\Delta}_F-2\alpha\norm{G}_\infty\sum_{(i,j)\in S_{\bar{\epsilon}}^c}|P^*_{ij}|>0,~\label{eq:corJaffard-5}
\end{align}
}%
where \eqref{eq:corJaffard-2} follows from the triangle inequality, 
 \eqref{eq:corJaffard-3} follows under the probability event $\Omega$ as defined in the proof of Theorem~\ref{thm:statAnalysis} and the definition of $q(\bar{\epsilon})$ in~\eqref{eq:def_S_e} and $G_{ij}=\tilde{G}_{ij}/G_{\min}\geq 1$, as defined in~\eqref{eq:distMat}, and finally \eqref{eq:corJaffard-5} holds 
 for any $\Delta\in S_{\Delta}$. Hence, following a similar argument within the proof of Theorem~\ref{thm:statAnalysis}, under the probability event $\Omega$, we have
\begin{equation}
\scriptsize
\norm{\hat{\Delta}}_F \leq \alpha{b^*}^2(1+q(\bar{\epsilon})/G_{\min})|S_{\bar{\epsilon}}|^{1/2}+\Big(\big(\alpha{b^*}^2(1+q(\bar{\epsilon})/G_{\min})|S_{\bar{\epsilon}}|^{1/2}\big)^2+4\alpha{b^*}^2\norm{G}_{\infty}\sum_{(i,j)\in S_{\bar{\epsilon}}^c}|P^*_{ij}|\Big)^{1/2}.
\end{equation}
Now, suppose the following inequality holds:
\begin{align}
\scriptsize
4{b^*}^{-2}\norm{G}_{\infty}(1+q(\bar{\epsilon})/G_{\min})^{-2} \frac{1}{|S_{\bar{\epsilon}}|} \sum_{(i,j)\in S_{\bar{\epsilon}}^c}|P^*_{ij}| & \leq \alpha \label{eq:alpha_lb}
\end{align}
which is equivalent to
\begin{align*}
\scriptsize
\Big(4\alpha{b^*}^2\norm{G}_{\infty}\sum_{(i,j)\in S_{\bar{\epsilon}}^c}|P^*_{ij}|\Big)^{1/2} &\leq  \alpha{b^*}^2(1+q(\bar{\epsilon})/G_{\min})|S_{\bar{\epsilon}}|^{1/2}.
\end{align*}
Therefore, provided that \eqref{eq:alpha_lb} holds, we have
\begin{equation}\label{eq:bound_exact}
\norm{\hat{P}-P^*}_F\leq (1+\sqrt{2}){b^*}^2(1+q(\bar{\epsilon})/G_{\min})|S_{\bar{\epsilon}}|^{1/2}\alpha.
\end{equation}%
It follows from the proof of Theorem~\ref{thm:statAnalysis} that setting $\alpha=40\theta^*_v\sqrt{N_0/N}$ implies
\begin{equation}\label{eq:cor12_bound}
\norm{\hat{P}-P^*}_F\leq \cO\Big(\log(1/\bar{\epsilon}) |S_{\bar{\epsilon}}|^{1/2}\sqrt{N_0/N}\Big)
\end{equation}
with high probability (i.e., $1-n^{-M}$). Hence, using the definition of $N_0$ and by taking $N\geq N_0$, then with probability $1-n^{-M}$, we get the bound in \eqref{eq:probBoundJaffard}. Provided the sufficient condition \eqref{eq:alpha_lb},
from \eqref{eq:ineq_series} in the proof of Theorem~\ref{thm:secondstage}, we get $\norm{\hat{\th}^{(N)}-\th^*}\leq\epsilon$ whenever
\begin{equation*}
\frac{\sqrt{n}}{{a^*}^2}\norm{\hat{P}-P^*}_F\leq\delta_{\epsilon}\triangleq\frac{\gamma^*\sqrt{n}\epsilon}{2\kappa\norm{\bJ\br(\th^*_{\rho})}_2},
\end{equation*}
which is equivalent to
\begin{equation}
\label{eq:eps_condition}
\norm{\hat{P}-P^*}_F\leq\frac{\gamma^*{a^*}^2\epsilon}{2\kappa\norm{\bJ\br(\th^*_{\rho})}_2}.
\end{equation}
Therefore, using the bound for $\norm{\hat{P}-P^*}_F$ provided in \eqref{eq:bound_exact} for $\alpha=40\theta^*_v\sqrt{N_0/N}$, the condition in \eqref{eq:eps_condition} holds for all large $N$ such that
\begin{equation}\label{eq:ineq_1}
(1+\sqrt{2}){b^*}^2(1+q(\bar{\epsilon})/G_{\min})|S_{\bar{\epsilon}}|^{1/2}40\theta^*_\nu\sqrt{\frac{N_0}{N}} \leq \frac{\gamma^*{a^*}^2}{2\kappa\norm{\bJ\br(\th^*_{\rho})}_2}\epsilon.
\end{equation}
Define $N_\epsilon$ such that \eqref{eq:ineq_1} holds with equality. Next, we discuss how $\bar{\epsilon}>0$ should be set so that the sufficient condition in \eqref{eq:alpha_lb} holds for $\alpha_\epsilon\triangleq 40\theta^*_v\sqrt{N_0/N_\epsilon}$, i.e.,
{\footnotesize
\begin{align}
\label{eq:explicit_condition}
4{b^*}^{-2}\norm{G}_{\infty}(1+q(\bar{\epsilon})/G_{\min})^{-2} \frac{1}{|S_{\bar{\epsilon}}|} \sum_{(i,j)\in S_{\bar{\epsilon}}^c}|P^*_{ij}| & \leq \alpha_\epsilon =
 \frac{(a^*/b^*)^2}{2(1+\sqrt{2})}\frac{\gamma^*}{\kappa\norm{\bJ\br(\th^*_{\rho})}_2}\frac{(1+q(\bar{\epsilon})/G_{\min})^{-1}}{\sqrt{|S_{\bar{\epsilon}}|}}\epsilon
\end{align}
}%
where the equality follows from the definitions of $N_\epsilon$ and $\alpha_\epsilon$.
Since $\kappa=\cO(n)$ and $\norm{\bJ\br(\th^*_{\rho})}_2=\cO(n)$, we define
\begin{equation}\label{eq:g_n}
g_n(\bar{\epsilon})\triangleq \frac{n^2\sum_{(i,j)\in S_{\bar{\epsilon}}^c}|P^*_{ij}| }{|S_{\bar{\epsilon}}|^{1/2}(1+q(\bar{\epsilon})/G_{\min})},
\end{equation}
which is clearly an increasing function of $\bar{\epsilon}$. Hence, for any $\epsilon>0$, there exists an $\bar{\epsilon}>0$ small enough such that
\begin{equation}\label{eq:condition}
\frac{g_n(\bar{\epsilon})}{\epsilon} \leq \frac{{a^*}^2\gamma^*}{8(1+\sqrt{2})\norm{G}_\infty}.
\end{equation}
Therefore, it follows from \eqref{eq:explicit_condition} that for $\bar{\epsilon}$ satisfying \eqref{eq:condition}, the sufficient condition in \eqref{eq:alpha_lb} is true for $\alpha_\epsilon=40\theta^*_\nu\sqrt{\frac{N_0}{N_\epsilon}}$, where $N_\epsilon=\cO(1/\epsilon^2)$.
This completes the proof.
\end{proof}

\begin{figure}[htpb]
  \centering
  \includegraphics[scale=0.6]{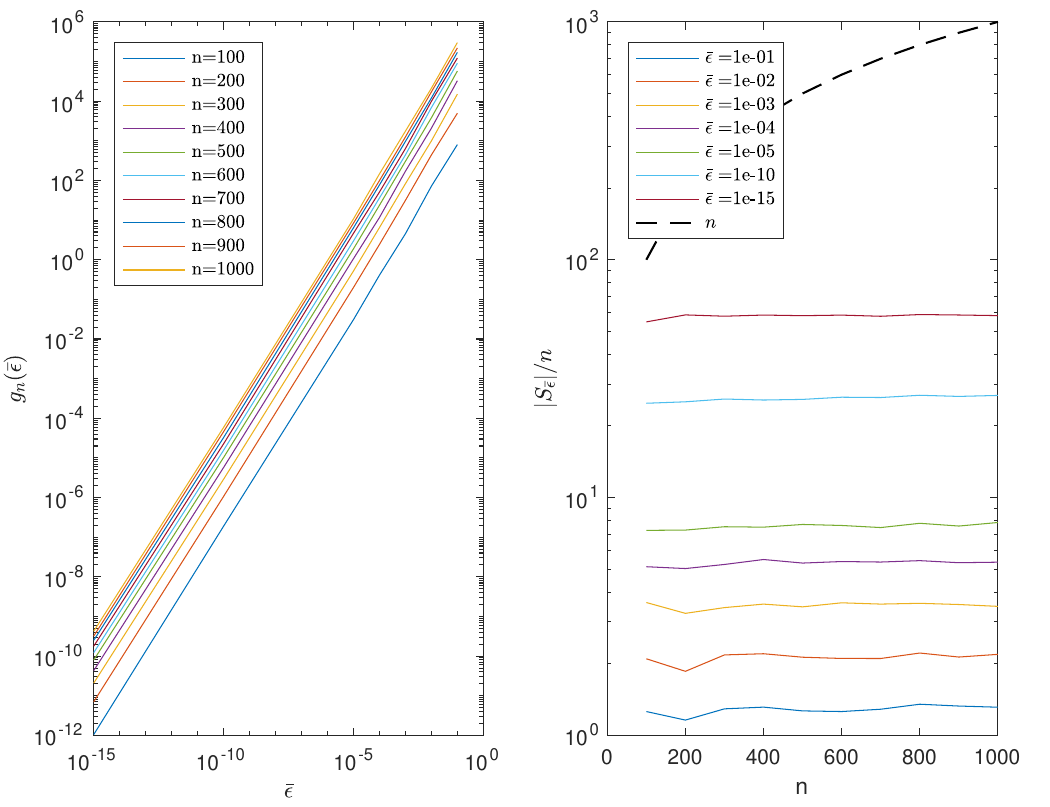}
  \caption{Exponential covariance function}
  \label{fig:ass_exp}
\end{figure}

\begin{figure}[htpb]
  \centering
  \includegraphics[scale=0.6]{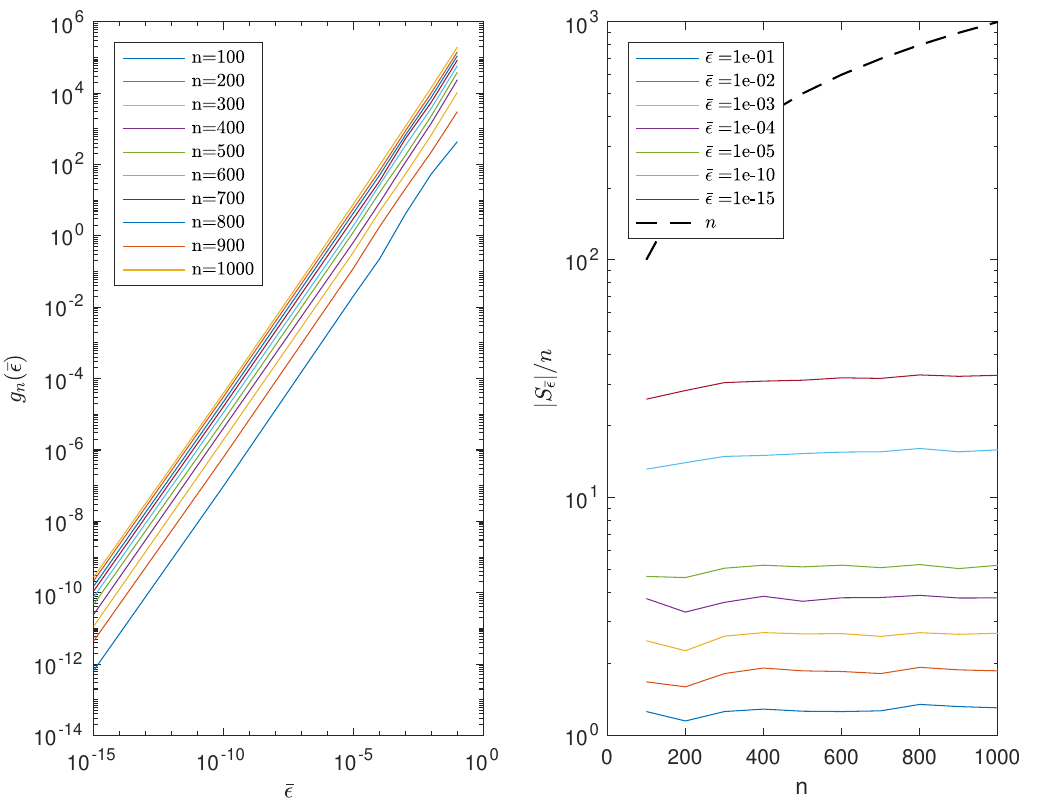}
  \caption{Matern covariance function}
  \label{fig:ass_matern}
\end{figure}

\begin{figure}[htpb]
  \centering
  \includegraphics[scale=0.6]{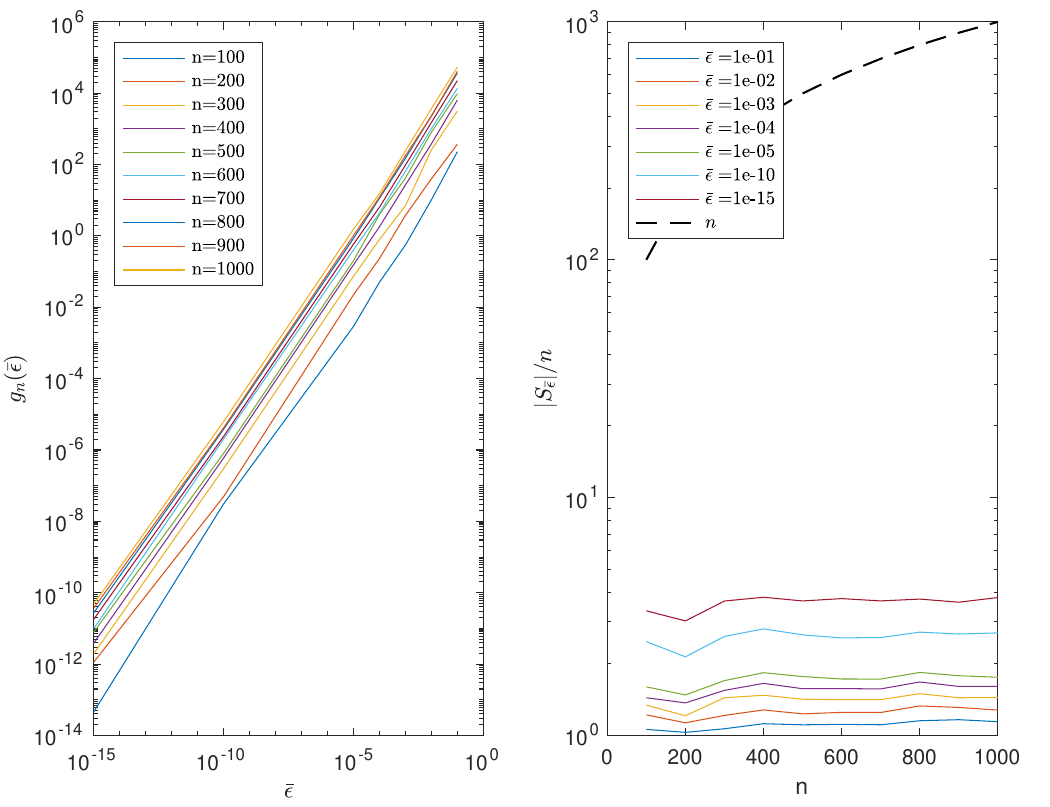}
  \caption{Squared exponential covariance function}
  \label{fig:ass_sq_exp}
\end{figure}

\begin{remark} \label{rem:cor12}
Note that $g_n(\bar{\epsilon})$ in \eqref{eq:condition} is an increasing function of $\bar{\epsilon}$ since $|S_{\bar{\epsilon}}|$ is nonincreasing (hence $|S^c_{\bar{\epsilon}}|$ is nondecreasing), and $q(\bar{\epsilon})$ is a decreasing function of $\bar{\epsilon}$ -- see also the left plots in Figures~\ref{fig:ass_exp}, \ref{fig:ass_matern}, and \ref{fig:ass_sq_exp}. Hence, for any $\epsilon>0$ (the STAGE-II bound), there exists $\bar{\bar{\epsilon}}>0$ such that the condition \eqref{eq:condition} holds for all $\bar{\epsilon}\in(0,\leq\bar{\bar{\epsilon}}]$. The plots on the left in Figures~\ref{fig:ass_exp}, \ref{fig:ass_matern}, and \ref{fig:ass_sq_exp} show decay of $g_n(\bar{\epsilon})$ with decreasing $\bar{\epsilon}$ for exponential, Matern ($\nu=3/2$), and squared exponential covariance functions over a  two-dimensional uniform design, respectively. The corresponding plots on the right illustrate the growth of $|S_{\bar{\epsilon}}|$ in $n$ for different $\bar{\epsilon}$ values. More precisely, to visually compare the growth of $|S_{\bar{\epsilon}}|$ and $n^2$, we plotted $|S_{\bar{\epsilon}}|/n$ and $n$ in log-scale against $n$ in the x-axis. Note $|S_{\bar{\epsilon}}|/n$ behaves as a constant that depends on $\bar{\epsilon}$; hence, $|S_{\bar{\epsilon}}|$ scales as $\cO(n)$ compared to $\mathrm{card}(P^*)=\cO(n^2)$ appearing in the bound provided in Theorem~\ref{thm:statAnalysis} -- see also Section~\ref{sec:num4cor12}.
\end{remark}

\begin{figure}[t]
	\centering
	\includegraphics[scale=0.6]{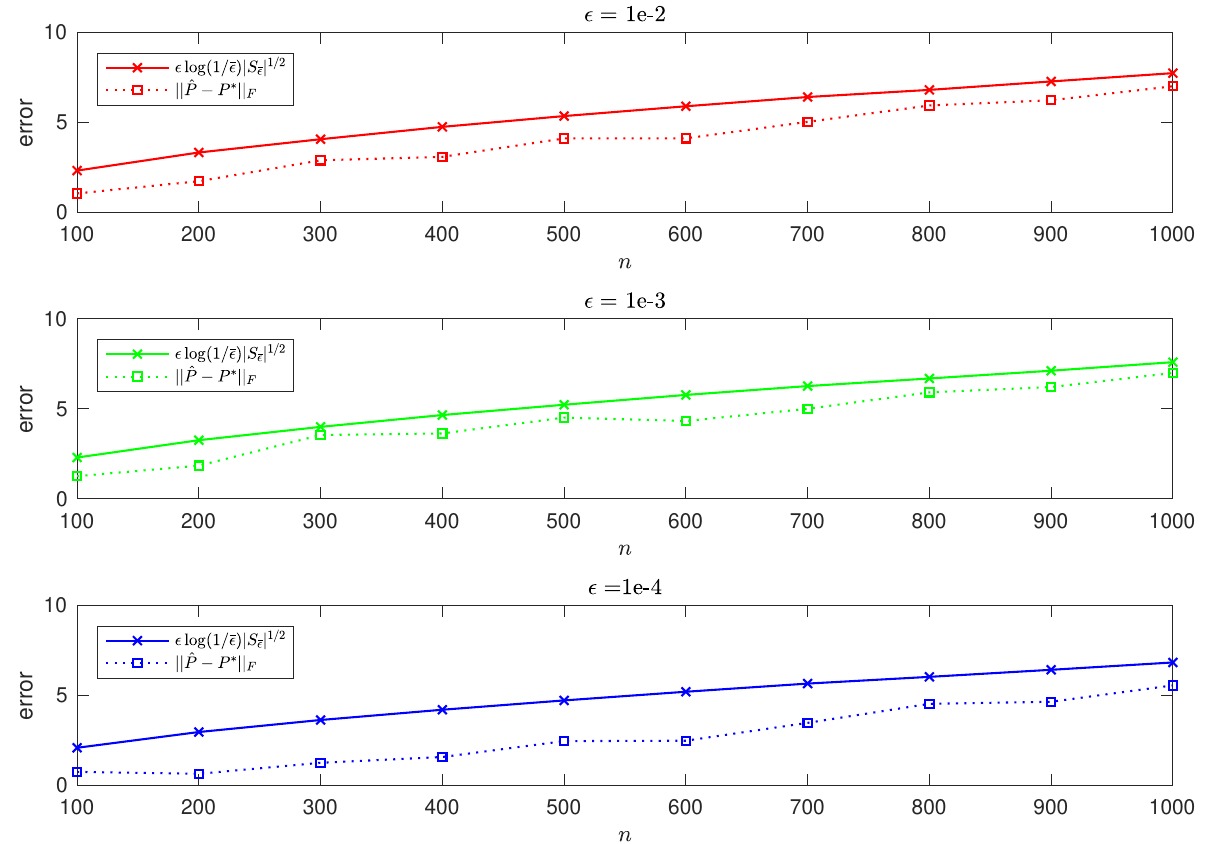}
	\caption{Actual error $\norm{\hat{P}-P^*}_F$ versus the theoretical bound $\epsilon\log(1/\bar{\epsilon}) |S_{\bar{\epsilon}}|^{1/2}$}
	\label{fig:4cor12_unscaled}
\end{figure}

\section{Numerical performance of the proposed SPS method} \label{sec:numericals}
In this section, the performance of the proposed algorithms is reported for both simulated and real data sets. To solve the STAGE-I problem, the sparsity parameter $\alpha$ in \eqref{eq:convexProgram} 
was set to {$c\sqrt{\log(n)/N}$ (except for Section~\ref{sec:num4cor12} 
) with $c=10^{-3}$}. Furthermore, to solve STAGE-I problem, we used a particular ADMM implementation displayed in Figure~\ref{alg:admm} of the online supplement, for which the penalty sequence $\{\rho_{\ell}\}$ is set
to a geometrically increasing sequence $\rho_{\ell+1}=1.05\rho_{\ell}$ with $\rho_0=n$. In the simulation studies $R$ denotes the number of simulation replications. {The numerical tests were carried on computer with an Intel Xeon x5650 CPU and 12.0 GB memory.} 

\subsection{
Actual error versus the theoretical bound based on Corollary~\ref{cor:probBound4JaffardCov}}\label{sec:num4cor12}
We simulate an isotropic GRF with exponential covariance function with $\th^*_\rho=10$, $\theta^*_\nu=1$, and $\theta^*_0=0$ in a 2-dimensional square $\cX=[0,100]^2$ over a uniform design with $n=100,200,...,1000$ points, and calculate $C^*$ and $P^*$ accordingly. Next, we sample $N=N_0/\epsilon^2$ realizations from the resulting GRFs for $\epsilon\in\{1e-2,1e-3,1e-4\}$, where we set $M=1$ in the definition of $N_0$.

To obtain the \emph{numerical error}, the STAGE-I problems are solved using $\alpha=40\theta^*_\nu\sqrt{N_0/N}$ with $N=N_0/\epsilon^2$, i.e., $\alpha=40\theta^*_\nu\epsilon$ to obtain $\hat{P}$. The numerical error is then calculated as $\norm{\hat{P}-P^*}_F$. The \emph{theoretical error bound} is computed based on \eqref{eq:probBoundJaffard} of Corollary~\ref{cor:probBound4JaffardCov}; in particular, using \eqref{eq:cor12_bound} in the proof Corollary~\ref{cor:probBound4JaffardCov}, i.e., for any fixed $\epsilon>0$ we set the bound to $\log(1/\bar{\epsilon}) |S_{\bar{\epsilon}}|^{1/2}\sqrt{N_0/N}$ for $N=N_0/\epsilon^2$ and $\bar{\epsilon}>0$ satisfying \eqref{eq:condition}; hence, it is given as $\epsilon\log(1/\bar{\epsilon}) |S_{\bar{\epsilon}}|^{1/2}$. Given $\epsilon$, to compute $\bar{\epsilon}$ satisfying \eqref{eq:condition}, one needs $a^*=\lambda_{\min}(P^*)$ and $\gamma^*$, which is the strong convexity modulus of the $g(\cdot)$ function defined in \eqref{eq:g-def} and is calculated using \eqref{eq:gamma_star}.
For any $\epsilon\in\{1e-2,1e-3,1e-4\}$, to calculate $g_n(\bar{\epsilon})$ small enough satisfying \eqref{eq:condition}, we start from $\bar{\epsilon}=1$ and keep dividing by 10 until the condition is satisfied. The first $\bar{\epsilon}$ that satisfies \eqref{eq:condition} is then used to calculate the theoretical error bound. Results are displayed in Figure~\ref{fig:4cor12_unscaled}, which shows that 
the numerical error nicely matches with the theoretical bound for all values of $\epsilon\in\{1e-2,1e-3,1e-4\}$.

\subsection{SPS vs ML parameter estimates}
\vspace*{-3mm}
An \emph{anisotropic} zero-mean GRF with a squared exponential correlation function, i.e., \eqref{eq:anisCorrFunc} with $M(\th_\rho)=\diag(\th_{\rho}^{-2})$ and $\th_{\rho}\in\Theta_\rho=\reals^d_{+}$, was simulated $R$ times in a hypercube domain $\cX=[0,10]^d$ where the variance and nugget parameters are fixed at $\theta_{\nu}^*=1$, $\theta_0^*=0.1$, and $\th^*_{\rho}\in\mR^d$ is sampled randomly from the intersection of a hypersphere having radius $10$ within the positive quadrant in each replication -- $\th^*_{\rho,l}$ denotes the true correlation parameter vector for the $l$-th replication. 
Table~\ref{tbl:SPSvsMLE} compares the quality of the SPS 
and ML parameter estimates in terms of the mean and standard deviation of $\{\norm{\hat{\th}_l-\th_l^*}\}_{l=1}^R$ for $R=5$ repeated model fits as the dimension $d$, the numbers of locations $n$, and 
process realizations $N$ change, where $\th_l^*=[{\th_{\rho,l}^*}^\top,\theta_v^*,\theta_0^*]^\top$ and $\hat{\th}_l$ is the SPS estimate for the $l$-th replication. To deal with the nonconcavity of the likelihood function, the ML method is initialized from 1, 10, and 100 random starting points, and the \emph{stationary} solutions with the best objective values among those corresponding to 1, 10, and 100 initializations are denoted as MLE-1, MLE-10, and MLE-100, respectively. To avoid numerical inconsistencies when solving the likelihood problem, the numerically stable approach mentioned in \cite{Lophaven02} is used. Table~\ref{tbl:SPSvsMLE} also includes the timing comparison between the two methods in seconds. Since the run times do not change much as $N$ changes, we reported the run times for each $(d,n)$ setting averaged over changing values of $N$. We highlighted in the blue color the regimes where SPS performs better than \emph{all} MLE methods in the mean.
According to these empirical findings, \emph{generally} in higher dimensions, e.g., $d\in\{5,10\}$ and specifically for $N\geq5$, SPS works as well as (or even better than) all MLE methods. Comparing the estimation times, SPS almost always beats MLE-10 and finishes an order of magnitude sooner than MLE-100. In the lower dimensional case ($d=2$), SPS has a reasonable estimation performance, better than MLE-1; but is dominated by MLE-10 and MLE-100. 
On the other hand, it is worth emphasizing the fact that the better performance of MLE-10 and MLE-100 comes at the cost of considerably longer computation time.
\begin{table}[h!]
    \begin{center}
    \tiny
    \begin{threeparttable}
    \begin{tabular}[t]{lllrrrrr}
    \toprule
     &  &  &\multicolumn{4}{c}{N} & \\
     \cmidrule(r){4-7}
     d & n & Method & \multicolumn{1}{c}{1} & \multicolumn{1}{c}{5} & \multicolumn{1}{c}{10} & \multicolumn{1}{c}{40} & Time in seconds \\
    \cmidrule(r){1-1} \cmidrule(r){2-2} \cmidrule(r){3-3} \cmidrule(r){4-7} \cmidrule(r){8-8}
    \multirow{12}{*}{2} & \multirow{4}{*}{100} & SPS & \textbf{2.9} (2.1)&\textbf{2.5} (1.7)&\textbf{1.6} (0.9)&\textbf{1.1} (0.6)& \blue{\textbf{6.3} (0.9)} \\
      &  & MLE-1 & \textbf{2.6} (1.7)&\textbf{2.1} (1.5)&\textbf{1.4} (0.9)&\textbf{1.0} (0.7)&\textbf{1.2} (0.2)  \\
      &  & MLE-10 & \textbf{2.6} (1.7)&\textbf{2.1} (1.5)&\textbf{1.4} (0.9)&\textbf{1.0} (0.7)&\blue{\textbf{11.1} (1.1)}  \\
      &  & MLE-100 & \textbf{2.6} (1.7)&\textbf{2.1} (1.5)&\textbf{1.4} (0.9)&\textbf{1.0} (0.7)&\blue{\textbf{119.2} (7.1)}  \\
      \cmidrule(r){2-8}
      & \multirow{4}{*}{500} & SPS & \textbf{2.7} (1.8)&\textbf{1.8} (1.1)&\textbf{1.5} (0.9)&\textbf{1.0} (0.6)&\blue{\textbf{274.3} (8.1)} \\
      &  & MLE-1 & \textbf{2.3} (1.5)&\textbf{1.4} (1.0)&\textbf{1.0} (0.6)&\textbf{1.0} (0.4)&\textbf{49.7} (25.1) \\
      &  & MLE-10 & \textbf{2.3} (1.5)&\textbf{1.4} (1.0)&\textbf{1.0} (0.6)&\textbf{1.0} (0.4)&\blue{\textbf{407.6} (72.2)}  \\
      &  & MLE-100 & \textbf{2.3} (1.5)&\textbf{1.4} (1.0)&\textbf{1.0} (0.6)&\textbf{1.0} (0.4)&\blue{\textbf{3931.5} (419.4)}  \\
      \cmidrule(r){2-8}
      & \multirow{4}{*}{1000} & SPS & \textbf{2.1} (1.4)&\textbf{1.6} (0.9)&\textbf{1.3} (0.6)&\textbf{0.9} (0.3)&\blue{\textbf{1793.9} (41.6)}  \\
      &  & MLE-1 & \textbf{1.9} (1.3)&\textbf{1.2} (0.8)&\textbf{0.9} (0.5)&\textbf{0.8} (0.2)&\textbf{405.3} (342.4) \\
      &  & MLE-10 & \textbf{1.9} (1.3)&\textbf{1.2} (0.8)&\textbf{0.9} (0.5)&\textbf{0.8} (0.2)&\blue{\textbf{2858.6} (434.9)}  \\
      &  & MLE-100 & \textbf{1.9} (1.3)&\textbf{1.2} (0.8)&\textbf{0.9} (0.5)&\textbf{0.8} (0.2)&\blue{\textbf{24771.1} (2413.1)} \\
    \cmidrule(r){1-8}
    \multirow{12}{*}{5} & \multirow{4}{*}{100} & SPS & \textbf{3.2} (1.6)&\blue{\textbf{2.3} (1.4)}&\blue{\textbf{1.8} (1.2)}&\blue{\textbf{1.3} (0.8)}&\blue{\textbf{7.6} (2.4)}  \\
      &  & MLE-1 & \textbf{3.1} (1.9)&\blue{\textbf{2.5} (1.7)}&\blue{\textbf{2.1} (1.6)}&\blue{\textbf{1.8} (1.4)}&\textbf{1.9} (0.4)  \\
      &  & MLE-10 & \textbf{3.1} (1.9)&\blue{\textbf{2.5} (1.7)}&\blue{\textbf{1.9} (1.6)}&\blue{\textbf{1.7} (1.4)}&\blue{\textbf{20.7} (0.8)}  \\
      &  & MLE-100 & \textbf{2.9} (1.5)&\textbf{2.3} (1.6)&\blue{\textbf{1.9} (1.5)}&\blue{\textbf{1.6} (1.2)}&\blue{\textbf{233.7} (68.6)} \\
      \cmidrule(r){2-8}
      & \multirow{4}{*}{500} & SPS & \textbf{2.9} (1.7)&\textbf{1.9} (1.4)&\textbf{1.8} (1.1)&\textbf{1.3} (0.8)&\blue{\textbf{359.3} (45.3)} \\
      &  & MLE-1 & \textbf{2.8} (1.7)&\textbf{1.9} (1.6)&\textbf{1.8} (1.5)&\textbf{1.6} (1.4)&\textbf{72.3} (12.2)  \\
      &  & MLE-10 & \textbf{2.8} (1.7)&\textbf{1.8} (1.6)&\textbf{1.8} (1.5)&\textbf{1.3} (1.1)&\blue{\textbf{782.0} (116.1)} \\
      &  & MLE-100 & \textbf{2.5} (1.7)&\textbf{1.7} (1.4)&\textbf{1.6} (1.4)&\textbf{1.2} (1.0)&\blue{\textbf{7924.2} (1720.7)}  \\
      \cmidrule(r){2-8}
      & \multirow{4}{*}{1000} & SPS & \textbf{2.8} (1.6)&\textbf{1.6} (1.1)&\textbf{1.3} (0.7)&\blue{\textbf{0.9} (0.4)}&\blue{\textbf{2050.8} (137.7)}  \\
      &  & MLE-1 & \textbf{2.7} (1.4)&\textbf{1.8} (1.3)&\textbf{1.5} (1.0)&\blue{\textbf{1.1} (0.9)}&\textbf{520.9} (133.4)  \\
      &  & MLE-10 & \textbf{2.5} (1.4)&\textbf{1.7} (1.3)&\textbf{1.3} (1.0)&\blue{\textbf{1.0} (0.7)}&\blue{\textbf{4485.7} (686.0)}  \\
      &  & MLE-100 & \textbf{2.3} (1.3)&\textbf{1.5} (1.3)&\textbf{1.2} (0.9)&\blue{\textbf{1.0} (0.6)}&\blue{\textbf{49587.4} (1099.7)} \\
   \cmidrule(r){1-8}
    \multirow{12}{*}{10} & \multirow{4}{*}{100} & SPS & \blue{\textbf{5.8} (2.3)}&\blue{\textbf{4.1} (1.6)}&\blue{\textbf{3.2} (1.4)}&\blue{\textbf{1.9} (1.0)}&\blue{\textbf{9.8} (5.3)} \\
      &  & MLE-1 & \blue{\textbf{6.6} (2.6)}&\blue{\textbf{5.1} (2.4)}&\blue{\textbf{5.0} (2.4)}&\blue{\textbf{4.3} (2.1)}&\textbf{3.9} (2.2) \\
      &  & MLE-10 & \blue{\textbf{6.2} (2.3)}&\blue{\textbf{5.1} (2.1)}&\blue{\textbf{4.9} (1.9)}&\blue{\textbf{4.8} (1.9)}&\blue{\textbf{48.9} (10.7)}  \\
      &  & MLE-100 & \blue{\textbf{6.2} (2.1)}&\blue{\textbf{5.0} (2.0)}&\blue{\textbf{4.4} (1.9)}&\blue{\textbf{3.9} (1.8)}&\blue{\textbf{532.1} (121.8)} \\
      \cmidrule(r){2-8}
      & \multirow{4}{*}{500} & SPS & \textbf{4.9} (2.0)&\blue{\textbf{3.9} (1.5)}&\blue{\textbf{2.9} (1.3)}&\blue{\textbf{1.6} (0.9)}&\blue{\textbf{284.6} (8.7)} \\
      &  & MLE-1 & \textbf{5.8} (2.4)&\blue{\textbf{4.9} (2.2)}&\blue{\textbf{4.3} (2.0)}&\blue{\textbf{3.7} (1.7)}&\textbf{230.4} (154.7)  \\
      &  & MLE-10 & \textbf{5.0} (2.3)&\blue{\textbf{4.8} (2.0)}&\blue{\textbf{3.9} (1.9)}&\blue{\textbf{3.6} (1.7)}&\blue{\textbf{2873.7} (1042.7)}  \\
      &  & MLE-100 & \textbf{4.8} (2.0)&\blue{\textbf{4.3} (1.9)}&\blue{\textbf{3.4} (1.8)}&\blue{\textbf{2.8} (1.5)}&\blue{\textbf{42739.0} (21786.0)}  \\
      \cmidrule(r){2-8}
      & \multirow{4}{*}{1000} & SPS & \textbf{4.8} (1.9)&\blue{\textbf{3.6} (1.3)}&\blue{\textbf{2.6} (1.3)}&\blue{\textbf{1.4} (0.8)}&\blue{\textbf{3544.1} (171.2)} \\
      &  & MLE-1 & \textbf{5.8} (2.3)&\blue{\textbf{4.8} (2.0)}&\blue{\textbf{4.1} (1.7)}&\blue{\textbf{3.6} (1.6)}&\textbf{644.8} (180.3)  \\
      &  & MLE-10 & \textbf{4.8} (2.1)&\blue{\textbf{4.5} (1.8)}&\blue{\textbf{3.7} (1.4)}&\blue{\textbf{3.4} (1.3)}&\blue{\textbf{5535.6} (926.5)} \\
      &  & MLE-100 & \textbf{--} (--)&\blue{\textbf{--} (--)}&\blue{\textbf{--} (--)}&\blue{\textbf{--} (--)}&\blue{\textbf{--} (--)} \\
    \bottomrule
    \end{tabular}
    \end{threeparttable}
    \end{center}
    \vspace*{-0.5cm}
     \caption{{\scriptsize{SPS vs MLE methods. The reported numbers are the mean (standard deviation) of $\{\norm{\hat{\th}_l-\th_l^*}\}_{l=1}^5$.} Blue color is used to highlight the regimes where the proposed method performs better than the MLE counterparts in the mean. For $d=10$ and $n=1000$, one replicate of MLE-100 was not finished in 24hrs.}}
    \label{tbl:SPSvsMLE}
\end{table}

\begin{figure} [htbp]
  \centering
  \includegraphics[height=6cm,width=6cm]{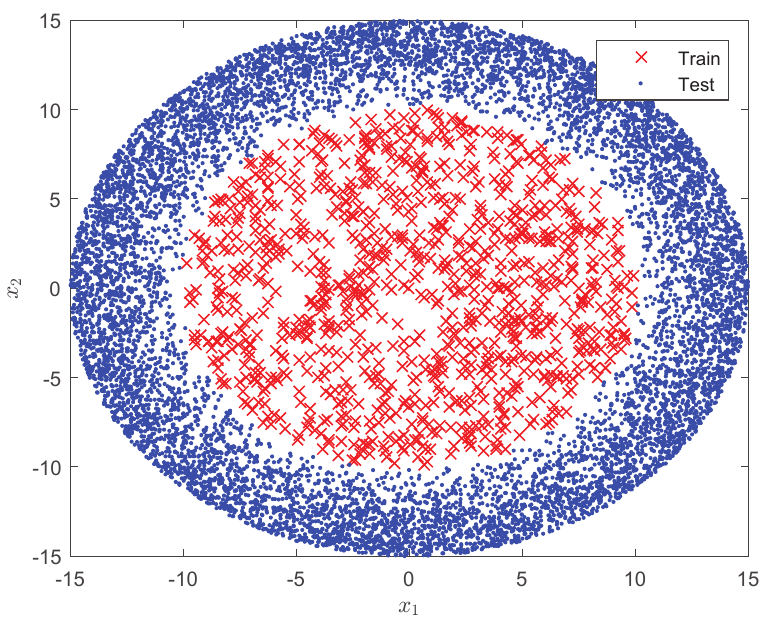}
  \hspace{1cm}
  \includegraphics[height=6cm,width=6cm]{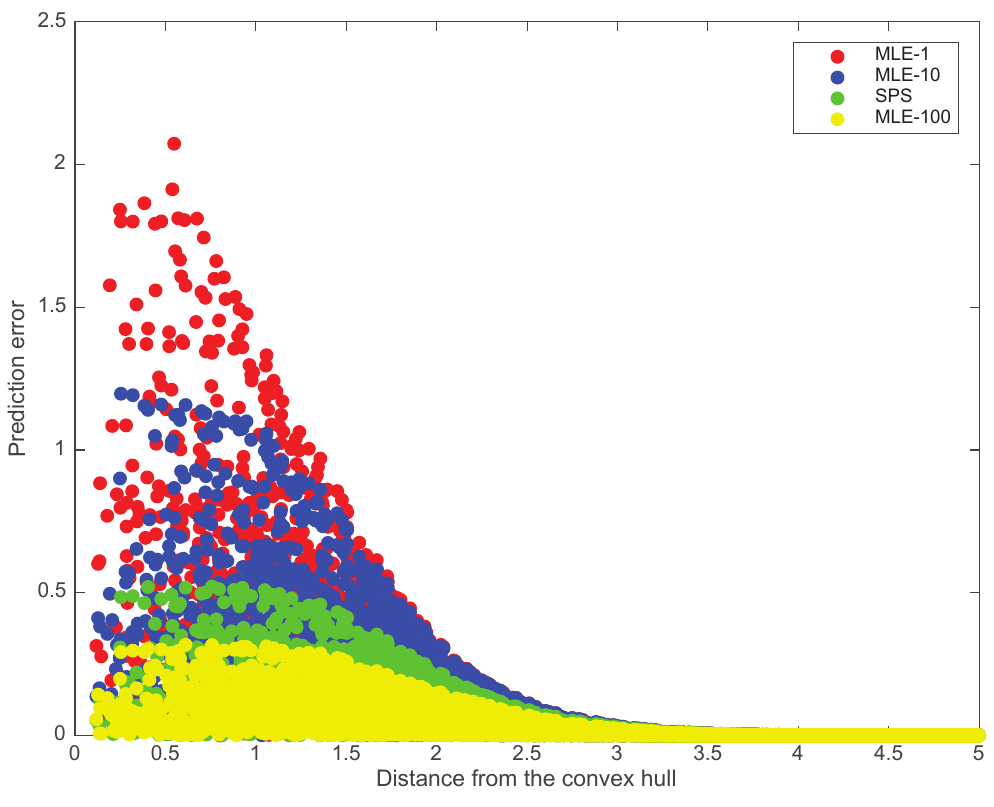}
  \vspace*{-2mm}
  \caption{{\scriptsize 
  \textbf{Left:} Hyper-sphere design locations projected on the $x_1-x_2$ plane. \textbf{Right:} Prediction performance}}
  \label{fig:f6}
\end{figure}

Next, we discuss the effect of parameter estimation quality on the process predictions. When $n$ locations are dense in the domain, interpolating predictions may still be adequate even if the parameter estimates are 
biased; however, when the location density is low, poor estimates will result in weak prediction performance. This issue is further aggravated in \emph{extrapolation} scenarios. To show the extrapolation behavior, we sample $n=1000$ training data within a 10-dimensional hypersephere with radius 10 from a zero-mean isotropic GRF with variance, nugget, and range parameters equal to 1, 0.1, and $\sqrt{2}$, respectively. Next, we sample 10,000 test data of which distance to the center is 
between 10 and 15, i.e., from a hyper-ring. The left graph on Figure~\ref{fig:f6} shows the design locations projected on the $x_1-x_2$ plane. The training data is used to fit {GRF models} 
{using} MLE-1, MLE-10, MLE-100~{(see the paragraph above for their definitions)}, and SPS. The graph on the right of Figure~\ref{fig:f6} shows the prediction errors as a function of the distance between the test point and the convex hull of the training data set. Prediction performance of SPS is better than MLE-1 and MLE-10; but slightly worse than MLE-100, while estimation time is greatly in favor of SPS compared to the MLE-100 method. \vspace*{-2mm}

\subsection{Dealing with the ``big $n$" problem}
\label{sec:blocking}
To solve {the} STAGE-I problem efficiently for large $n$, we propose to segment the set of training locations $\mathcal{D}^x\triangleq\{\mb{x}_i\}_{i=1}^n\subset\cX$ into $K$ segments (or blocks) $\{\mathcal{D}^x_k\}_{k=1}^K$ of size $n_k\triangleq\mbox{card}(\cD^x_k)$ such that the number of data points in each segment, $n_k$, is less than $n_B$ (in our experiments we set $n_B$ to $1000$). If $n\leq n_B$, then segmentation is not needed; otherwise, we used the following two segmentation schemes in our numerical experiments.

\textbf{Spatial Segmentation (SS):} This scheme is based on partitioning the spatial domain $\mathcal{X}$ into $K$ non-overlapping hypercubes, and considering the training data points in each hypercube $\cX_k$ as one block. Let $\cX_k\triangleq\{\bx\in\reals^d:\ \bar{\ell}_k\leq \bx\leq \bar{u}_k\}$ for some $\{\bar{\ell}_k,~\bar{u}_k\}_{k=1}^K$ such that $\cup_{k=1}^K\cX_k=\cX$ and $\cX_{k_1}\cap \cX_{k_2}=\emptyset$ for all $k_1\neq k_2$. Then we define $\cD^x_k\triangleq\cD^x\cap\cX_k$ for all $k$. Assuming that data point locations, $\cD^x$, are uniformly distributed within $\mathcal{X}$, each block $\cD^x_k$ will contain $\frac{n}{K}$ observations in expectation.

\textbf{Random Segmentation (RS):} The set of training data locations $\cD^x$ is partitioned \emph{uniformly at random} into $K$ blocks $\{\cD^x_k\}_{k=1}^K$ such that the first $K-1$ blocks contain $\lfloor\frac{n}{K}\rfloor$ data locations and the last block contains $n-(K-1)\lfloor\frac{n}{K}\rfloor$ many. 
Let $\mathcal{D}_k=\{(\mb{x}_i,y^{(r)}_i):\ r=1,\ldots,N,~i\in \mathcal{I}_k\}$ denote the subset of training data corresponding to the $k$-th block, where the index set $\mathcal{I}_k$ is defined as $\cI_k\triangleq\{1\leq i\leq n:\ \mb{x}_i\in\cD^x_k\}$. Hence, $n_k=|\cI_k|$. Note that the SS {segmentation} scheme, but not the RS scheme, can handle non-stationary GRFs.

For both segmentation schemes, we solve STAGE-I problem {\em for each segment $k$} separately using the sample covariance $S_k$ and the matrix of pairwise distances $G_k$ corresponding to observations in segment $k$, 
, i.e., 
$\hat{P}_k = \argmin_{P\succ 0} \fprod{S_k,P}-\log\det(P)+\alpha_k\fprod{G_k,|P|}$, 
for $k=1,\ldots,K$, for some $\alpha_k>0$. 
Under the assumption that the underlying stochastic process is second-order stationary, i.e., the covariance function parameters are fixed across the domain $\mathcal{X}$ \citep{SteinBook}, one can fit a single covariance function for the whole domain. In this case, for both schemes 
we propose to estimate the covariance function parameters by solving the following least squares problem:
$\hat{\boldsymbol{\theta}} \in \argmin_{\boldsymbol{\theta}\in\Theta} \sum_{k=1}^K\norm{{\hat{P}_k}^{-1}-C_k(\boldsymbol{\theta})}_F^2$.
Note that this method generates a predicted surface with {\em no discontinuities} along the boundary between segments. This is in contrast to other methods that partition large datasets for fitting a GRF but require further formulation to achieve continuity, see \cite{Park2011}. Finally, in case the process cannot be assumed to be stationary, the second stage optimization is solved separately for each segment $k$, which has its own covariance parameter estimates. These estimates are computed by solving
$\hat{\boldsymbol{\theta}}_k \in \argmin_{\boldsymbol{\theta}_k\in\Theta} \norm{{\hat{P}_k}^{-1}-C_k(\boldsymbol{\theta}_k)}_F^2$, for each segment $k=1,\ldots,K$.


\vspace{0.2cm}

\textbf{{SPS-fitted GRF for small and large data sets.}} We simulated two data sets of sizes $n=1,000$ and $n=64,000$ points from a GRF with zero mean and isotropic SE covariance function with parameters: range $\th^*_{\rho}=4$, variance $\theta^*_v=8$, and nugget $\theta^*_0=4$ over a square domain $\cX=[0,~100]\times[0,~100]$. In the simulation with $n=1000$, the results are based on R (number of simulation replications) equal to 100,
while for the simulation with $n=64,000$, given that 
each run of the simulation takes around 3-4 hours, results are given for 
$R=5$ replicates. {The number of realizations $N$ is set to 1 in these simulations. Let $\boldsymbol{\hat{\theta}}_{l}$ denote the covariance parameter estimates obtained in the $l$-th replication. In all the tables, $\boldsymbol{\bar{\theta}}\triangleq\sum_{l=1}^{R}\boldsymbol{\hat{\theta}}_l/R$ and $\mathbf{stdev}_{\boldsymbol{\theta}}\triangleq\sqrt{\tfrac{1}{R}\sum_{l=1}^R(\boldsymbol{\hat{\theta}}_l-\boldsymbol{\bar{\theta}})^2}$ denote the sample mean and the standard deviation of the parameter estimates, respectively.
For benchmarking, we 
compare our estimates with those obtained by \emph{Domain Decomposition} method (DDM) of \cite{Park2011} using a rectangular mesh. The number of control points on the boundaries and the number of constraining degrees of freedom equal to 3. DDM is selected since it shows the best performance among the other big-n methods considered in Section~\ref{sec:comparison}.}
\begin{table} [h!]
    \renewcommand{\arraystretch}{1.2}
    \centering
    \scriptsize
    \begin{adjustbox}{max width=\textwidth}
    \begin{tabular}{l|cc|cc}
    \hline
     & \multicolumn{2}{c|}{$n$=1000 (R=100 replicates)} & \multicolumn{2}{c}{$n$=64000 (R=5 replicates)} \\
    \hline
    Method & $\overline{\boldsymbol{\theta}}=(\overline{\th}_\rho,\overline{\theta}_v,\overline{\theta}_0)$ & $\mathbf{stdev}_{\boldsymbol{\theta}}$ &
    $\overline{\boldsymbol{\theta}}=(\overline{\th}_\rho,\overline{\theta}_v,\overline{\theta}_0)$
    & $\mathbf{stdev}_{\boldsymbol{\theta}}$  \\
    \hline
    SPS-SS & (3.98, 7.77, 4.87) & (0.41, 1.01, 0.76)
    & (4.01, 8.16, 4.75) & (0.43, 0.78, 0.29)  \\
    SPS-RS & {(4.01, 8.11, 4.22)} & {(0.90, 1.16, 0.85)}
    & (3.98, 7.97, 4.83) & (0.06, 0.11, 0.11) \\
    \hline
    {DDM} & (1.73, 7.34, 6.11) & (1.23, 1.98, 2.67)
    & (1.43, 8.67, 3.65) & (0.87, 0.95, 0.53) \\
     \hline
    \end{tabular}
    \end{adjustbox}
    \vspace*{-2mm}
    \caption{{\scriptsize {Parameter estimate of SPS under the two segmentation schemes vs. DDM (Domain Decomposition method of \cite{Park2011}) for simulated data with $N$=1 realization. The covariance function is squared-exponential and the true parameter values are $\th^*_\rho=4$, $\theta^*_v=8$, and $\theta^*_0=4$.}}}
    \label{tbl:sim1}
\end{table}

Both 
{segmentation} schemes were used for comparison. When $n=1000$, for the SS {segmentation} scheme, the domain was split into $3\times 3=9$ equal size square segments; for the RS {segmentation} scheme $\cD^x$ was randomly partitioned into 9 equal cardinality sets. Similarly, when $n=64,000$, for the SS {segmentation} scheme, the domain was split into $8\times 8=64$ equal size square segments; for the RS 
scheme, $\cD^x$ was randomly partitioned into 64 equal cardinality sets. Table \ref{tbl:sim1} shows the model fitting results.

{When $n=1000$, the parameter estimates using either {segmentation} scheme appear unbiased.
Computing $\hat{\boldsymbol{\theta}} \in \argmin_{\boldsymbol{\theta}\in\Theta} \sum_{k=1}^K\norm{{\hat{P}_k}^{-1}-C_k(\boldsymbol{\theta})}_F^2$, we explicitly ignore the correlation of process values for any two points in different blocks.
Over the fixed domain when $n$ is large, i.e., $n=64,000$, the larger data location density results in more observations close to boundaries; hence, correlations between blocks for the SS scheme may not be ignored anymore. Empirical results show lower $\mathbf{stdev}_{\boldsymbol{\theta}}$ under the RS 
scheme when $n$ is large. DDM underestimates the range parameter in all scenarios.}

\textbf{Effect of range and nugget parameters.} {To analyze the effect of the range, $\th^*_{\rho}$, and nugget, $\theta^*_0$, parameters on the performance of the proposed method, we setup another simulation with $n$=64,000 points with results shown in Table~$\ref{tbl:sim2}$. 
When the range parameter increases, the standard deviations of the parameter estimates increase under both {segmentation} schemes. Furthermore, the RS 
scheme appears to be less sensitive to changes in the nugget parameter. In general, for big-n scenarios and given the high point density, the RS scheme {results in} more robust parameter estimates. DDM highly underestimates the range parameter, especially when $\th^*_\rho=30$.}
\begin{table} [htbp]
    \centering
    \scriptsize
    \begin{adjustbox}{max width=\textwidth}
    \renewcommand{\arraystretch}{1.25}
    \begin{tabular}{c|l|cc|cc}
    \hline
     & & \multicolumn{2}{c|}{$\th^*_\rho=4$} & \multicolumn{2}{c}{$\th^*_\rho=30$} \\
    \hline
    Nugget & {Method} & $\overline{\boldsymbol{\theta}}=(\overline{\th}_\rho,\overline{\theta}_v,\overline{\theta}_0)$ & $\mathbf{stdev}_{\boldsymbol{\theta}}$ &
    $\overline{\boldsymbol{\theta}}=(\overline{\th}_\rho,\overline{\theta}_v,\overline{\theta}_0)$
    & $\mathbf{stdev}_{\boldsymbol{\theta}}$ \\
    \hline
    \multirow{3}{*}{$\theta^*_0=4$} & SPS-SS & (4.01, 8.16, 4.75) & (0.43, 0.78, 0.29)
    & (29.03, 7.94, 4.80) & (1.15, 1.76, 0.11) \\

     & SPS-RS & (3.98, 7.97, 4.83) & (0.06, 0.11, 0.11)
    & (29.24, 7.95, 4.79) & (1.93, 0.26, 0.19) \\

    & {DDM} & (1.43, 8.67, 3.65) & (0.87, 0.95, 0.53)
    & (19.32, 8.65, 3.69) & (0.97, 0.64, 0.73) \\
    \hline
    \multirow{3}{*}{$\theta^*_0=8$} & SPS-SS & (4.03, 8.08, 8.77) & (0.47, 0.86, 0.36)
    & (28.39, 7.97, 8.77) & (1.35, 1.89, 0.13)  \\

     & SPS-RS & (3.98, 7.98, 8.83) & (0.07, 0.12, 0.15)
    & (28.93, 7.87, 8.78) & (1.65, 0.31, 0.18) \\

    & {DDM} & (1.27, 7.35, 9.10) & (0.08, 0.08, 0.25)
    & (11.45, 7.35, 10.10) & (1.89, 0.70, 0.49) \\
    \hline
    \end{tabular}
    \end{adjustbox}
    \caption{{\scriptsize {SPS vs. DDM (Domain Decomposition method of \cite{Park2011}) estimates for simulated data sets with $n=64,000$, $N=1$ and $R=5$. The covariance function is squared-exponential with variance parameter $\theta^*_v=8$.}}}
    \label{tbl:sim2}
\end{table}

\subsection{SPS vs state-of-the-art for fitting GRFs to big data sets}
\label{sec:comparison}
We compare the SPS method against the \emph{Partial Independent Conditional} (PIC) method by \cite{snelson2007}, the \emph{Domain Decomposition} (DDM) method of \cite{Park2011}, and the \emph{Full Scale covariance Approximation} (FSA) of \cite{SangHuang2012}. \cite{Park2011} provided computer codes for PIC and DDM, and we coded the FSA method. 
We simulated a data set of size $n=64,000$ generated from a zero mean GRF with isotropic Squared-Exponential (SE) covariance function with $N=1$ realization using the following parameter values: $\th^*_{\rho}$=4, 
$\theta^*_v$=8, 
and $\theta^*_0$=4. 

{For each replication, 90\% of the simulated data was allocated for training (i.e., for estimating the parameters), and 10\% for testing (prediction). The Mean Square Prediction Error~(MSPE) is computed on the \emph{test data}. The MSPE corresponding to the $l$-th replication is computed as follows:
$\mathrm{MSPE}_{l} \triangleq \frac{1}{n_{l}^t}\norm{\mb{y}_{l}^t-\hat{\mb{y}}_{l}^t}^2$,
where $n_{l}^t$ denotes the number of \emph{test data} points in the $l$-th replication, $\mb{y}_{l}^t\in\reals^{n_{l}^t}$ and $\hat{\mb{y}}_{l}^t\in\reals^{n_{l}^t}$ are vectors of true and predicted function values, respectively. Since the true parameter values are known for the simulated data set, the true function values $\mb{y}^t$ are taken to be the predictions obtained via \eqref{eq:predDist} using the \emph{true} parameter values. Calculating the MSPE this way shows the specific error due to the discrepancies between the estimated and true parameter values. Finally, $\overline{\rm{MSPE}}$ and $\mathbf{stdev}_{\rm{MSPE}}$ are defined similar to $\overline{\boldsymbol{\theta}}$ and $\mathbf{stdev}_{\boldsymbol{\theta}}$.}

In the SPS method, we used SS segmentation scheme with 64 equal-size blocks as described in Section~\ref{sec:blocking}. In the PIC method, the number of local regions was set to 64, and the number of pseudo inputs was set to 100. In the DDM method, a rectangular mesh was selected with both the number of control points on the boundaries and the number of constraining degrees of freedom equal to 3. In the FSA method, the number of knots was set to 50 on a regular grid, the tapering function used was spherical with taper range set to 10. These settings are based on the guidance provided in the corresponding papers. In the training phases for PIC, DDM, and FSA methods, the initial values for each covariance function parameters were randomly selected from the uniform distribution over $(0,10]$ in each replicate. The reason is that these methods attempt to solve non-convex problem in \eqref{eq:likelihoodOpt}; hence, the local minima generated by the optimization solvers highly depend on the initial point. 
Therefore, to be fair to these methods, we run them starting from many randomly generated initial solutions for each replicate.
The mean and standard deviation of MSPE and parameter estimates for $R=5$ replications are reported in Table \ref{tbl:comparison}.

\begin{figure} [htbp]
  \centering
\includegraphics[height=5.5cm,width=6.5cm]{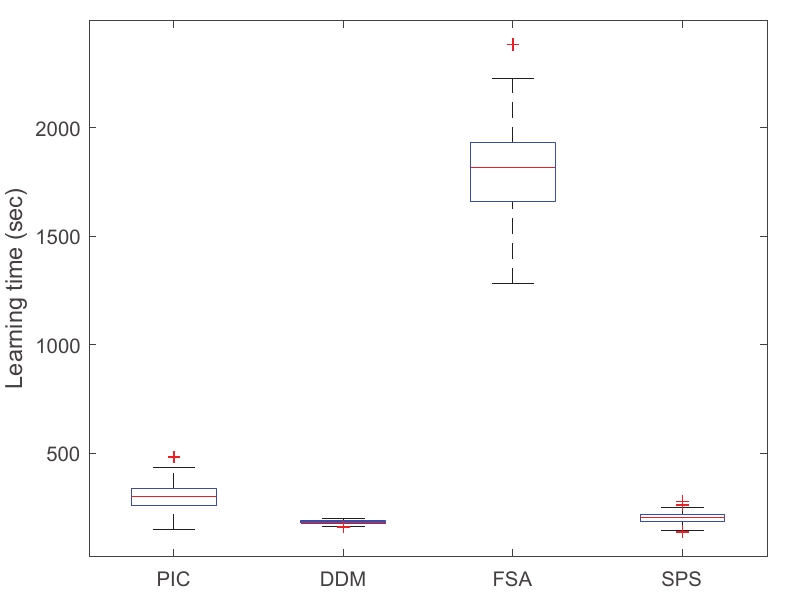}
  \hspace{0.5cm}
\includegraphics[height=5.5cm,width=6.5cm]{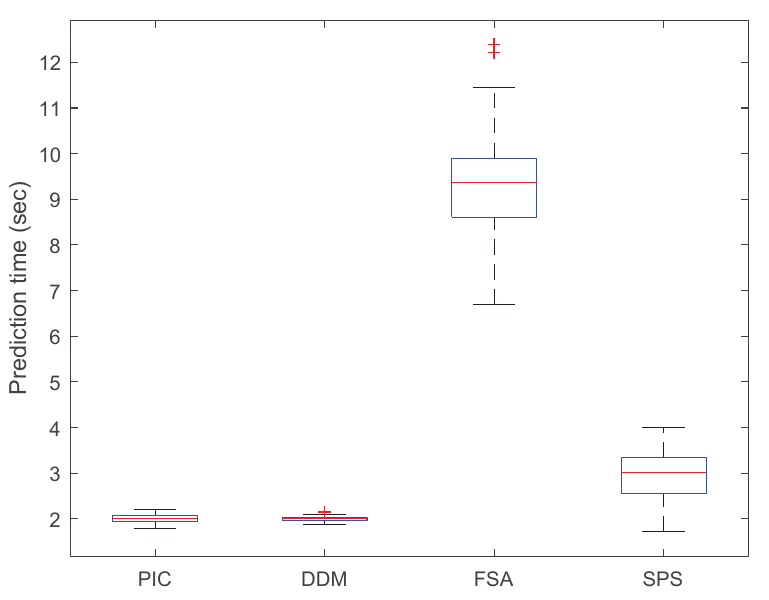}
  \vspace*{-2mm}
  \caption{
  \textbf{Left:} Learning times \textbf{Right:} Prediction times of PIC, DDM, FSA, and SPS methods.}
  \label{fig:duration}
\end{figure}

\begin{table} [htbp]
    \renewcommand{\arraystretch}{1.2}
    \centering
    \scriptsize
    \begin{tabular}{lcccc}
    \hline
    Method & $\overline{\boldsymbol{\theta}}=(\overline{\th}_\rho,\overline{\theta}_v,\overline{\theta}_0)$ & $\mathbf{stdev}_{\boldsymbol{\theta}}$ & $\overline{\rm{MSPE}}$ & $\mathbf{stdev}_{\rm{MSPE}}$ \\
    \hline
    PIC & (5.11, 6.22, 5.01) & (2.03, 2.82, 2.532) &  2.87 & 1.03 \\
    DDM & (0.83, 8.23, 4.03) & (0.09, 0.96, 0.73) &  2.23 &  0.44  \\
    FSA & (3.31, 2.58, 0.65) & (2.97, 0.76, 0.14) &  4.47  &  1.35 \\
    SPS & (4.14, 7.82, 4.64) & (0.65, 1.06, 0.57) &  0.42  &  0.35 \\
    \hline
    \end{tabular}
    \caption{{\scriptsize Comparison of the SPS method against PIC, DDM, and FSA for $n=64,000$, $N=1$ and $R=5$ on data sets generated from a GRF with zero mean and SE covariance function with true parameters $\boldsymbol{\theta}^*=(\th_\rho^*,\theta^*_v,\theta^*_0)=(4,8,4)$.}}
    \label{tbl:comparison}
\end{table}
The SPS method provides the least biased estimates for all three covariance parameters, with the degree of bias provided by the other methods being much more substantial. Furthermore, the mean MSPE for the SPS method is considerably lower than that of the other alternatives (one order of magnitude less), and {has the least variability}. The CPU times required by each method in the learning and prediction stages are displayed in Figure~\ref{fig:duration}.

For both learning and prediction the DDM method is the fastest, and the FSA method the slowest. {However, note that while PIC, DDM and SPS use segmentation (64 blocks in this example), there is no segmentation in FSA but instead one should select the number of knots. Hence computing times of FSA against the other three methods are not completely comparable.} Of the remaining two methods contrasted, PIC is faster than SPS in the prediction phase but SPS is faster than PIC in learning. In view of the prediction performance of all the methods compared, the slight speed advantage of DDM over SPS is not a demerit of our method: DDM is unable to provide an unbiased estimate of the range parameter $\th^*_{\rho}$, crucial in spatial modeling, and this naturally results in considerably worse predictions.

\subsection{Implementation of the SPS method for real data sets}
Finally, we tested the SPS method on two real data sets. The first data set contains ozone and air quality data as recorded by the Total Ozone Mapping Spectrometer (TOMS) instrument on board the Nimbus-7 satellite. The data set contains 48,331 Total Column Ozone (TCO) measurements over the globe on October 1, 1988 and is available at NASA's website\footnote{$\mbox{http://ozoneaq.gsfc.nasa.gov/nimbus7Ozone.md}$}. The second data set is the Day/Night Cloud Fraction (CF) from January to September 2009 (size $n=64,800$ points) collected by the Moderate Resolution Imaging Spectrometer (MODIS) instrument on board the Terra satellite, a data set also available at NASA's website\footnote{http://gdata1.sci.gsfc.nasa.gov/daac-bin/G3/gui.cgi?instance\_id=MODIS\_MONTHLY\_L3}. A wrapper Matlab function which can read the TCO data with its specific format and produce the input/output matrices is available with our software package at our lab's website \url{http://sites.psu.edu/engineeringstatistics/computer-codes/}.
\begin{table} [h!]
    \renewcommand{\arraystretch}{1.2}
    \centering
    \scriptsize
    \begin{tabular}{lcccccc}
    \hline
    Data & Covariance Function & {Segmentation} & $\overline{\boldsymbol{\theta}}=(\overline{\th}_\rho,\overline{\theta}_v,\overline{\theta}_0)$ & $\mathbf{stdev}_{\boldsymbol{\theta}}$ & $\overline{\rm{MSPE}}$ & $\mathbf{stdev}_{\rm{MSPE}}$ \\
    \hline
    TCO & Matern & RS & (12.20, 1098, 0.00) & (0.01, 0.54, 0.00) &  4.5361 & 0.0623 \\
    \hline
    CF & Exponential & RS & (10.07, 0.05, 0.82) & (0.04, 0.00, 0.00) &  0.0044 &  0.0000  \\
    \hline
    \end{tabular}
    \caption{{\scriptsize Implementation of the proposed SPS method for two real data sets, TCO and CF, with performance measures computed over R=5 replicated cross-validations (10\% of data randomly sampled in each replication for testing).}}
    \label{tbl:realData}
\end{table}

The mean and standard deviations of the MSPEs and the parameter estimates for $R=5$ replicates are reported in Table \ref{tbl:realData}. In each replication, 10\% of the data is randomly selected for testing, and the remaining 90\% is used for learning the covariance parameters. {Both {segmentation} schemes were implemented and RS {segmentation resulted in} 
better prediction results for both 
data sets; {therefore, we only report RS results}. 
The RS {segmentation} scheme is adopted using a random partition of $\cD^x$ in each replication}. The type of covariance function 
was selected based on the best MSPE values obtained. Since this is real data, we cannot make any judgment about the quality of parameter estimates; however, the standard deviations are quite small relative to the parameter estimate magnitudes for both data sets. \vspace*{-2mm}

\section{Conclusions and further research} \label{sec:concludings}
A new two-stage method to estimate the parameters of Gaussian Random Field (GRF) models is presented and its theoretical error bound is established. This method, which we named Sparse Precision matrix Selection (SPS), first finds a sparse estimate of the precision (inverse covariance) matrix of the underlying GRF by solving a nonsmooth convex optimization problem, and then estimates the parameters of the GRF model by solving a least-square problem. Numerical studies confirming our theoretical findings are presented; these include numerical comparisons with MLE which requires solving a nonconvex problem. In higher dimensional regimes, especially with higher number of realizations, the SPS method performs better than its MLE counterparts. Moreover, the computational time of SPS scales much better with the number of locations, replicated observations per location, and especially, with the dimension.

The sparse estimation of the precision matrix of the GRF model is well-motivated and theoretically supported by Jaffard's decay algebra \citep{jaffard1990proprietes}. 
Indeed, we were able to bound the error of the sparse estimate of the true precision matrix, which is dense but its elements decay rapidly to zero in magnitude. This bound then allowed us to establish the error bound of the parameter estimates in Corollary~\ref{cor:probBound4JaffardCov}.

The following are some possible future research directions:
\textbf{a)} The established error bound for the stage-I problem that involves sparse estimation of the true dense precision matrix includes the cardinality of some set $|S_{\bar{\epsilon}}|$ given in \eqref{eq:def_S_e} for some $\bar{\epsilon}>0$. Clearly, $|S_{\bar{\epsilon}}|\leq \mathrm{card}(P^*)$ for all $\bar{\epsilon}\geq 0$; however, theoretically quantifying the cardinality of this set for even simple designs, e.g. d-dimensional lattice, is a combinatorial challenge and is left for future works.
\textbf{b)} Our numerical motivations at the beginning of the paper suggest that the decay rate of the precision matrix increases with the density of the points (infill asymptotics). We did not investigate this phenomenon \emph{theoretically} nor did we utilize it in the proposed parameter estimation algorithm, and it is left as a matter for future research.


\bibliographystyle{authordate1} 
\bibliography{spsRefs-1}


\newpage
\section{Appendix}
\subsection{Proof of Theorem~\ref{thm:innerOpt}}
\label{sec:sup_proof_thm_inner}
\begin{proof}
For any given $\boldsymbol{\theta}_\rho\in\Theta_\rho$, note that $\mb{d}$ is not parallel to $\mb{r}=\mb{r}(\boldsymbol{\theta}_\rho)$, i.e., $\mb{d}\nparallel\mb{r}$. Let $h$ denote the objective function in \eqref{eq:appB-vectorForm}, i.e., $h(\theta_v,\theta_0)\triangleq \tfrac{1}{2}\norm{\theta_v\mb{r}
+\theta_0\mb{d}-\hat{\mb{c}}}^2$, it satisfies \vspace*{-2mm}
{\small
\begin{subequations}
\begin{align}
\frac{\partial h(\theta_v,\theta_0)}{\partial \theta_v}&=\mb{r}^\top(\theta_v\mb{r}+\theta_0\mb{d}-\hat{\mb{c}}), \label{eq:partialTheta1}\\
\frac{\partial h(\theta_v,\theta_0)}{\partial \theta_0}&=\mb{d}^\top(\theta_v\mb{r}+\theta_0\mb{d}-\hat{\mb{c}}). \label{eq:partialTheta0}
\vspace*{-4mm}
\end{align}
\end{subequations}}%
The Hessian of $h$ in 
\eqref{eq:appB-vectorForm} is
{\small
$\grad^2 h=\begin{bmatrix} \mb{r}^\top\mb{r} & \mb{r}^\top\mb{d} \\ \mb{r}^\top\mb{d} & \mb{d}^\top\mb{d} \end{bmatrix}$}.
Note $\mb{r}^\top\mb{r}>n>0$, and $\det(\grad^2 h)=\norm{\mb{r}}^2\norm{\mb{d}}^2-(\mb{r}^\top\mb{d})^2>0$ by Cauchy-Schwartz and the fact that $\mb{r}\nparallel\mb{d}$; thus, $\grad^2 h$ is positive definite. Therefore, 
for any given $\boldsymbol{\theta}_\rho\in\Theta_\rho$, $h$ is strongly convex jointly in $\theta_v$ and $\theta_0$.

From the definitions of $\mb{d}$ and $\mb{r}$, we have $\norm{\mb{d}}^2=n$, $\mb{d}^\top\mb{r}=n$, and $\norm{\mb{r}}^2>n$ (because $\mb{r}(\bx,\bx,\boldsymbol{\theta}_\rho)=1$ for any $\bx$ and $\boldsymbol{\theta}_\rho\in\Theta_\rho$). Necessary and sufficient KKT conditions imply 
\begin{small}
\begin{subequations}
\begin{align}
&\grad h(\theta_v,\theta_0) \geq \mb{0}, \label{eq:dualFs} \\
&\theta_v\geq 0,\quad \theta_0\geq 0, \label{eq:primalFs} \\
&\frac{\partial h(\theta_v,\theta_0)}{\partial \theta_v}~\theta_v = 0, \quad 
\frac{\partial h(\theta_v,\theta_0)}{\partial \theta_0}~\theta_0 = 0. \label{eq:csTheta}
\end{align}
\end{subequations}
\end{small}

Below, we consider four possible scenarios for problem \eqref{eq:appB-vectorForm}:
\begin{enumerate}
\item $(\theta_v=0, \theta_0=0)$ -- This solution is optimal if and only if $\mb{r}^\top\hat{\mb{c}}\leq 0$ 
    and $\mb{d}^\top\hat{\mb{c}}\leq 0$ (from \eqref{eq:partialTheta1} \eqref{eq:partialTheta0}, and \eqref{eq:dualFs}). However, since $\hat{C}=\hat{P}^{-1}$ is positive definite, its diagonal elements are strictly positive, $\mb{d}^\top\hat{\mb{c}}> 0$. Hence, this scenario is not possible.\vspace*{-3mm}
\item $(\theta_v=0, \theta_0>0)$ -- From \eqref{eq:dualFs}, \eqref{eq:csTheta} and \eqref{eq:partialTheta0} follows that $(\theta_v=0, \theta_0=\mb{d}^\top\hat{\mb{c}}/n)$ is the optimal solution if and only if $\mb{r}^\top\hat{\mb{c}}\leq\mb{d}^\top\hat{\mb{c}}$. \vspace*{-3mm} 
\item $(\theta_v>0, \theta_0>0)$ -- From \eqref{eq:partialTheta1}, \eqref{eq:partialTheta0}, \eqref{eq:csTheta}, and \eqref{eq:dualFs} 
    \begin{small}
    \begin{align*}
    (\theta_v,\theta_0)=\left(\frac{\mb{r}^\top\hat{\mb{c}}-\mb{d}^\top\hat{\mb{c}}}{\norm{\mb{r}}^2-n},
    \frac{(\mb{d}^\top\hat{\mb{c}})\norm{\mb{r}}^2/n-\mb{r}^\top\hat{\mb{c}}}{\norm{\mb{r}}^2-n}\right)
    \end{align*}
    \end{small}

\noindent is the optimal solution if and only if $(\mb{d}^\top\hat{\mb{c}})\norm{\mb{r}}^2/n>\mb{r}^\top\hat{\mb{c}}>\mb{d}^\top\hat{\mb{c}}$. \vspace*{-3mm}
\item $(\theta_v>0, \theta_0=0)$ -- From \eqref{eq:dualFs}, \eqref{eq:csTheta} and \eqref{eq:partialTheta1} follows that $(\theta_v=\mb{r}^\top\hat{\mb{c}}/\norm{\mb{r}}^2,\theta_0=0)$ is the optimal solution if and only if $\mb{r}^\top\hat{\mb{c}}\geq(\mb{d}^\top\hat{\mb{c}})\norm{\mb{r}}^2/n$. \vspace*{-3mm}
\end{enumerate}
\end{proof}

\newpage
\section{Online Supplementary Material}
\subsection{An ADMM Algorithm for solving Stage-I problem~\eqref{eq:convexProgram}}
\label{sec:ADMM}
\begin{theorem}
\label{thm:admm}
Let $0\leq a^*\leq b^*\leq\infty$. Given arbitrary $Z_0,W_0\in\mathbb{S}^n$ and $\rho>0$, let 
$\rho_\ell=\rho$ for $\ell\geq 0$, and $\{P_\ell,Z_\ell\}_{\ell\geq 1}$ denote the iterate sequence generated by \textbf{ADMM}$(S,G,\alpha,a^*,b^*)$ as shown in Figure~\ref{alg:admm}. Then $\{P_\ell\}$ converges
$Q$-linearly\footnote{Let $\{X_\ell\}$ converge to $X^*$ for a given norm $\norm{.}$. The convergence is called $Q$-linear if $\frac{\norm{X_{\ell+1}-X^*}}{\norm{X_{\ell}-X^*}}\leq c$, for some $c\in(0,1)$; and $R$-linear if $\norm{X_{\ell}-X^*}\leq c_\ell$, for some $\{c_\ell\}$ converging to 0 $Q$-linearly.} to $\hat{P}$, and $\{Z_\ell\}$ converges $R$-linearly to $\hat{P}$, where $\hat{P}$ is the unique optimal solution to STAGE-I problem given in \eqref{eq:convexProgram}. 
\end{theorem}
\begin{singlespace}
\begin{figure}[!h]
    {\small
    \rule[0in]{6.5in}{1pt}\\
    \textbf{Algorithm ADMM}$~(S,G,\alpha,a^*,b^*)$\\
    \rule[0.125in]{6.5in}{0.1mm}
    \vspace{-0.35in}
    \begin{algorithmic}[1]
    \STATE $\mathbf{input:}\ Z_0,W_0\in\mathbb{S}^n$, $\{\rho_\ell\}_{\ell\geq 0}\subset\reals_{++}$, $0\leq a^*\leq b^*\leq\infty$
    \STATE \textbf{if} $a^*>0$ \textbf{and} $b^*<\infty$ \textbf{then} $a\gets a^*$, $b\gets b^*$
    \STATE \textbf{if} $a^*=0$ \textbf{and} $b^*<\infty$ \textbf{then} $a\gets \min\{b^*, \frac{1}{\norm{S}_2+\alpha\norm{G}_F}\}$, $b\gets b^*$
    \STATE \textbf{if} $a^*>0$ \textbf{and} $b^*=\infty$ \textbf{then} $a\gets a^*$, $b\gets \frac{n a^*}{\alpha G_{\min}}\max\{\norm{S}_2+\alpha\norm{G}_F,~1/a^*\}$
    \STATE \textbf{if} $a^*=0$ \textbf{and} $b^*=\infty$ \textbf{then} $a\gets (\norm{S}_2+\alpha\norm{G}_F)^{-1}$, $b\gets n/(\alpha G_{\min})$
    \WHILE{$\ell\geq 0$}
    \STATE $P_{\ell+1}\gets \argmin_{P\in\mathbb{S}^{n}}\{\fprod{S,P}-\log\det(P)+\frac{\rho_\ell}{2}\norm{P-Z_\ell+\tfrac{1}{\rho_\ell}W_\ell}_F^2:\ a\mb{I}\preceq P \preceq b\mb{I}\}$ \label{algeq:admm_P}
    \STATE $Z_{\ell+1}\gets \argmin_{Z\in\mathbb{S}^{n}}\{\alpha\fprod{G,|Z|}+\frac{\rho_\ell}{2}\norm{Z-P_{\ell+1}-\tfrac{1}{\rho_\ell}W_\ell}_F^2:\ \diag(Z)\geq\mathbf{0}\}$ \label{algeq:admm_Z}
    \STATE $W_{\ell+1}\gets W_\ell+\rho_\ell(P_{\ell+1}-Z_{\ell+1})$
    \ENDWHILE
    \end{algorithmic}
    \rule[0.25in]{6.5in}{0.1mm}
    }
    \vspace*{-0.5in}
    \caption{{\scriptsize ADMM algorithm for STAGE-I}}\label{alg:admm}
\end{figure}
\end{singlespace}

\noindent \textbf{Remark.} As shown in the proof of Theorem~\ref{thm:admm}, when $a^*=0$ and/or $b^*=\infty$, the choice of $a,b$ in Figure~\ref{alg:admm} satisfies $a\leq \sigma_{\min}(\hat{P})\leq\sigma_{\max}(\hat{P})\leq b$ for $\hat{P}$, defined in \eqref{eq:convexProgram}. This technical condition makes sure that the ADMM iterate sequence converges linearly.

The 
algorithm 
is terminated at the end of iteration $\ell$ when both primal and dual residuals $(r_\ell, s_\ell)$ are below a given tolerance value, where $r_\ell\triangleq P_{\ell+1}-Z_{\ell+1}$ and $s_\ell\triangleq \rho_\ell(Z_{\ell+1}-Z_\ell)$. From the necessary and sufficient optimality conditions for
Step~\ref{algeq:admm_P} and Step~\ref{algeq:admm_Z} in Figure~\ref{alg:admm}, $r_\ell=s_\ell=0$ implies $P_{\ell+1}=Z_{\ell+1}=\hat{P}$, i.e., the unique optimal solution to \eqref{eq:convexProgram}.
In practice, ADMM converges to an acceptable accuracy within a few tens of iterations, which was also the case in our numerical experiments.

Typically, in ADMM algorithms~\citep{boyd2011}, 
the penalty parameter is held constant, i.e., $\rho_\ell = \rho> 0$ for all $\ell\geq 1$, for some $\rho>0$. Although the convergence 
is guaranteed for all $\rho> 0$, the empirical performance 
critically depends on the choice of 
$\rho$ -- it deteriorates rapidly if the penalty is set too large or too small~\citep{Kontogiorgis98}. 
Moreover, 
\cite{Lions79} discuss that there exists a $\rho^*>0$ which optimizes the convergence rate bounds for the constant penalty ADMM scheme; however, estimating $\rho^*$ is difficult in practice. 
In our experiments, we used an increasing penalty sequence $\{\rho_\ell\}_{\ell\geq 1}$. For details on the convergence of variable penalty ADMM, see \citep{he2002new,Aybat14} 
in addition to the references above.

Next, we show that Steps~\ref{algeq:admm_P} and \ref{algeq:admm_Z} of ADMM, displayed in Figure~\ref{alg:admm}, can be computed efficiently. Given a convex function $f:\mathbb{S}^n\rightarrow\reals\cup\{+\infty\}$ and $\lambda>0$, the proximal mapping $\prox{\lambda f}:\mathbb{S}^n\rightarrow\mathbb{S}^n$ is defined as $\prox{\lambda f}(\bar{P})\triangleq\argmin_{P\in\mathbb{S}^n}\lambda f(P)+\tfrac{1}{2}\norm{P-\bar{P}}_F^2$; and given a set $\cQ\subset\mathbb{S}^{n}$, let $\mathbf{1}_{\cQ}(\cdot)$ denote the indicator function of $\cQ$, i.e., $\mathbf{1}_{\cQ}(P)=0$ for $P\in\cQ$; otherwise equal to $+\infty$. {For the proof of Lemma~\ref{lem:prox_psi}, see~\citep{Yuan2012}. The result of Lemma~\ref{lem:prox_phi} for off-diagonal indices follows from the typical soft-thresholding from the lasso solution, see~\cite{friedman2007pathwise}. \vspace*{-2mm}}
\begin{lemma}
\label{lem:prox_psi}
Let $\Psi(P)\triangleq\fprod{S,P}-\log\det(P)+\mathbf{1}_{\cQ}(P)$, and $\cQ\triangleq\{P\in\mathbb{S}^{n}:\ a\mb{I}\preceq P \preceq b\mb{I}\}$. In generic form, Step~\ref{algeq:admm_P} of ADMM can be written as $\prox{\Psi/\rho}(\bar{P})$ for some $\bar{P}\in\mathbb{S}^{n}$ and $\rho>0$. Suppose $\bar{P}-\tfrac{1}{\rho}S$ has eigen-decomposition $U\diag(\bar{\lambda})U^\top$. Then $\prox{\Psi/\rho}(\bar{P})=U\diag(\lambda^*)U^\top$, where\vspace*{-3mm}
{
\begin{small}
\begin{equation}
\label{eq:opt_singular}
\lambda^*_i=\max\Big\{\min\Big\{\frac{\bar{\lambda}_i+\sqrt{\bar{\lambda}_i^2+4/\rho}}{2},\ b\Big\},\ a\Big\},\quad i=1,\ldots,n. \vspace*{-3mm}
\end{equation}
\end{small}}
\end{lemma}
\vspace*{-1.5mm}
\begin{lemma}
\label{lem:prox_phi}
Let $\Phi(P)\triangleq\alpha\fprod{G,|P|}+\mathbf{1}_{\cQ'}(P)$, and $\cQ'\triangleq\{P\in\mathbb{S}^n:\ \diag(P)\geq\mathbf{0}\}$. In generic form, Step~\ref{algeq:admm_Z} of ADMM can be written as $\prox{\Phi/\rho}(\bar{P})$ for some $\bar{P}\in\mathbb{S}^{n}$ and $\rho>0$, which can be computed as follows: \vspace*{-2mm}
\begin{small}
\begin{subequations}
\label{eq:phi_prox}
\begin{align}
(\prox{\Phi/\rho}(\bar{P}))_{ij}&=\sgn\left(\bar{P}_{ij}\right)\max\left\{|\bar{P}_{ij}|-\tfrac{\alpha}{\rho}G_{ij}, 0\right\}, \quad\forall (ij)\in\cI\times\cI \mbox{ s.t. } i\neq j,\\
(\prox{\Phi/\rho}(\bar{P}))_{ii}&=\max\left\{\bar{P}_{ii}-\tfrac{\alpha}{\rho}G_{ii}, 0\right\}, \quad\forall i\in\cI. \vspace*{-3mm}
\end{align}
\end{subequations}
\end{small}
\vspace*{-0.75cm}
\end{lemma}
The proofs of Theorem~\ref{thm:admm}, Lemma~\ref{lem:prox_psi} and Lemma~\ref{lem:prox_phi} follow from the existing results in the literature. 
For the sake of completeness, these proofs are provided in the supplementary material.
\subsubsection{Proof of Theorem~\ref{thm:admm}}
\label{sec:sup_proof_admm_thm}
Consider a more generic problem of the following form: \vspace*{-3mm}
\begin{small}
\begin{equation}
\min_{P\in\mathbb{S}^n}\Psi(P)+\Phi(P), \label{eq:generic_convex_problem}
\vspace*{-3mm}
\end{equation}
\end{small}

\noindent where $\Psi:\mathbb{S}^n\rightarrow\reals\cup\{+\infty\}$ and $\Phi:\mathbb{S}^n\rightarrow\reals\cup\{+\infty\}$ are proper closed convex functions, and $\mathbb{S}^n$ denotes the vector space of n-by-n symmetric matrices.
By introducing an auxiliary variable $Z\in\mathbb{S}^n$, \eqref{eq:generic_convex_problem} can be equivalently written as $\min\{\Psi(P)+\Phi(Z):\ P=Z,~P,Z\in\mathbb{S}^n\}$. For a given penalty parameter $\rho>0$, the augmented Lagrangian function is defined as 
\begin{small}
\begin{equation} \label{eq:augmentedLagrangian1_generic}
\cL_{\rho}(P,Z,W) \triangleq  \Psi(P)+\Phi(Z) + \fprod{W,P-Z}+\tfrac{\rho}{2}\norm{P-Z}_F^2,
\vspace*{-4mm}
\end{equation}
\end{small}

\noindent where $W\in\mathbb{S}^n$ is the dual multiplier for the linear constraint $P - Z = \mathbf{0}$.
Given an initial primal-dual point $Z^1,W^1\in\mathbb{S}^n$, when the ADMM algorithm~\citep{boyd2011} is implemented on \eqref{eq:generic_convex_problem}, it generates a sequence of iterates $\{P_\ell,Z_\ell\}_{\ell\geq 1}$ according to: 
\begin{small}
\begin{subequations}
\begin{align}
P_{\ell+1} & = \argmin_{P\in\mathbb{S}^n} \cL_{\rho}(P,Z_\ell,W_\ell)\ \ \ =\prox{\Psi/\rho}\left(Z_\ell-\tfrac{1}{\rho}W_\ell\right), \label{eq:ADMM-Pk_gen} \\
Z_{\ell+1} & = \argmin_{Z\in\mathbb{S}^n} \cL_{\rho}(P_{\ell+1},Z,W_\ell)=\prox{\Phi/\rho}\left(P_{\ell+1}+\tfrac{1}{\rho}W_\ell\right), \label{eq:ADMM-Zk_gen} \\
W_{\ell+1} & = W_\ell + \rho~(P_{\ell+1} - Z_{\ell+1}).
\vspace*{-8mm}
\end{align}
\label{eq:ADMM}
\vspace*{-3mm}
\end{subequations}
\end{small}

\noindent For all $\rho> 0$, convergence of the ADMM iterate sequence $\{P_\ell,Z_\ell\}_{\ell\geq 1}$ is guaranteed. 
In particular, $\lim_{\ell\geq 1}Z_\ell=\lim_{\ell\geq 1}P_\ell$; moreover, any limit point of $\{P_\ell\}$ is a minimizer of \eqref{eq:generic_convex_problem}.
Recently, \cite{Deng12} showed that the 
ADMM iterate sequence converges linearly if $\Psi$ is strongly convex and has a Lipschitz continuous gradient. In particular, $\{P_\ell, W_\ell\}$ converges\footnote{Q-linear and R-linear convergence were defined in Section~\ref{sec:ADMM}.}  $Q$-linearly to a primal-dual optimal pair $(P^{\rm opt},W^{\rm opt})$, where $P^{\rm opt}$ is the unique primal optimal solution, and $\{Z_\ell\}$ converges $R$-linearly to $P^{\rm opt}$.

Returning to the SPS method, note that the precision matrix estimation problem in \eqref{eq:convexProgram} immediately fits into the ADMM framework by setting $\Psi(P)=\fprod{S,P}-\log\det(P)+\mathbf{1}_{\cQ}(P)$ and $\Phi(P)=\alpha\fprod{G,|P|}$, where $\cQ\triangleq \{P\in\mathbb{S}^{n}:\ a^* \mathbf{I}\preceq P\preceq b^*\mathbf{I}\}$ and $\mathbf{1}_{\cQ}(\cdot)$ is the indicator function of $\cQ$, i.e., $\mathbf{1}_{\cQ}(P)=0$ if $P\in\cQ$; and it is equal to $+\infty$, otherwise. Therefore, both $\Psi$ and $\Phi$ are closed convex functions. When $0<a^*\leq b^*<+\infty$, Theorem~\ref{thm:admm} immediately follows from the convergence properties of ADMM discussed above.

Now consider the case $a^*=0$ and $b^*=+\infty$. For this scenario, $\Psi$ is strictly convex and differentiable on $\cQ$ with $\grad\Psi(P)=S-P^{-1}$; however, note that $\grad\Psi(P)$ is not Lipschitz continuous on $\cQ$. Therefore, this choice of $\Psi$ and $\Phi$ do not satisfy the assumptions in \citep{Deng12}. On the other hand, following the discussion in \citep{dAspremont2008}, we will show that by selecting a slightly different $\cQ$, one can obtain an equivalent problem to \eqref{eq:convexProgram} which does satisfy the ADMM convergence assumptions in \citep{Deng12}; hence, 
linear convergence rate for stage I of the SPS method can be obtained. Noting that $|t|=\max\{ut:\ |u|\leq 1\}$, one can write \eqref{eq:convexProgram} equivalently as follows: 
\begin{small}
\begin{equation}\label{eq:convexProgram_equiv1}
\hat{F}\triangleq \min_{P\succ 0}\quad \max_{\{U\in\mathbb{S}^n:\ |U_{ij}|\leq\alpha G_{ij}\}} \cL(P,U)\triangleq \fprod{S+U,P}-\log\det(P),
\vspace*{-4mm}
\end{equation}
\end{small}

\noindent where $\hat{F}\triangleq \fprod{S,\hat{P}}-\log\det(\hat{P})+\alpha\fprod{G,|\hat{P}|}$, and $\hat{P}$ is the solution to \eqref{eq:convexProgram}, i.e., $\hat{P}=\argmin\{\fprod{S,P}-\log\det(P)+\alpha\fprod{G,|P|}:\ P\succ\mathbf{0}\}$. Since $\cL$ is convex in $P$, linear in $U$, and $\{U\in\mathbb{S}^n:\ |U_{ij}|\leq\alpha G_{ij}\}$ is compact, the strong min-max property holds: 
\begin{small}
\begin{equation}
\hat{F}=\max_{\{U\in\mathbb{S}^n:\ |U_{ij}|\leq\alpha G_{ij}\}} \quad \min_{P\succ 0}\cL(P,U) 
=\max_{\{U\in\mathbb{S}^n:\ |U_{ij}|\leq\alpha G_{ij}\}} \quad n-\log\det\left((S+U)^{-1}\right), \label{eq:convexProgram_equiv2}
\end{equation}
\end{small}

\noindent where \eqref{eq:convexProgram_equiv2} follows from the fact that for a given $U\in\mathbb{S}^n$, $\hat{P}(U)=(S+U)^{-1}$ minimizes the inner problem if $S+U\succ\mathbf{0}$; otherwise, the inner minimization problem is unbounded from below. Therefore, we conclude that $\hat{P}$ is the optimal solution to \eqref{eq:convexProgram} if and only if there exists $\hat{U}\in\mathbb{S}^n$ such that $\hat{P}=(S+\hat{U})^{-1}\succ\mathbf{0}$, $|\hat{U}_{ij}|\leq\alpha G_{ij}$ for all $(i,j)\in\cI$, and $\fprod{S,\hat{P}}+\alpha\fprod{G,|\hat{P}|}=n$. Since $S,\hat{P}\succeq\mathbf{0}$, we have $\fprod{S,\hat{P}}\geq 0$; hence, $\fprod{G,|\hat{P}|}\leq n/\alpha$. Hence, we can derive the desired bounds, similar to those derived in~\citep{dAspremont2008}: \vspace*{-4mm} 
\begin{small}
\begin{align}
a&\triangleq \frac{1}{\norm{S}_2+\alpha\norm{G}_F}\leq\frac{1}{\norm{S}_2+\norm{\hat{U}}_F}\leq\frac{1}{\norm{S+\hat{U}}_2}
=\sigma_{\min}(\hat{P}), \label{eq:ak}\\
b&\triangleq \frac{n}{\alpha~G_{\min}}\geq\frac{\fprod{G,|\hat{P}|}}{G_{\min}}\geq\sum_{i,j}|\hat{P}_{ij}|\geq\norm{\hat{P}}_F\geq\norm{\hat{P}}_2=\sigma_{\max}(\hat{P}), \label{eq:bk}
\vspace*{-4mm}
\end{align}
\end{small}

\noindent where $G_{\min}\triangleq \min\{G_{ij}:\ (i,j)\in\cI\times\cI, i\neq j\}>0$. Therefore, 
\eqref{eq:convexProgram} is equivalent to 
\begin{small}
\begin{equation}\label{eq:convexProgram_equiv}
\hat{P} = \argmin \{\fprod{S,P}-\log\det(P)+\alpha\fprod{G,|P|}:\ a\mb{I}\preceq P \preceq b\mb{I}\},
\end{equation}
\end{small}

\noindent for $a$ and $b$ defined in \eqref{eq:ak} and \eqref{eq:bk}, respectively. Going back to the convergence rate discussion, when ADMM is applied to \eqref{eq:convexProgram_equiv} we can guarantee that the primal-dual iterate sequence converges linearly. In particular, we apply ADMM on \eqref{eq:generic_convex_problem} with 
\begin{small}
\begin{align}
\Psi(P)&=\fprod{S,P}-\log\det(P)+\mathbf{1}_{\tilde{\cQ}}(P),\quad \tilde{\cQ}\triangleq \{P\in\mathbb{S}^{n}:\ a\mb{I}\preceq P \preceq b\mb{I}\}, \label{eq:psi}\\
\Phi(P)&=\alpha\fprod{G,|P|}+\mathbf{1}_{\cQ'}(P), \quad \cQ'\triangleq \{P\in\mathbb{S}^n:\ \diag(P)\geq\mathbf{0}\}.\label{eq:phi}
\end{align}
\end{small}

\noindent Since $\tilde{\cQ}\subset\mathbb{S}^{n}_+\subset\cQ'$, the term $\mathbf{1}_{\cQ'}(.)$ in the definition of $\Phi$ appears redundant. However, defining $\Phi$ this way will restrict the sequence $\{Z_\ell\}$ to lie in $\cQ'$ rather than in $\mathbb{S}^n$, which leads to faster convergence to feasibility in practice. By resetting $\cQ$ to $\tilde{\cQ}$ as in \eqref{eq:psi}, we ensure that $\Psi$ is strongly convex with constant $1/b^2$ and $\grad\Psi$ is Lipschitz continuous with constant $1/a^2$. Indeed, the Hessian of $\Psi$ is a quadratic form on $\mathbb{S}^n$ such that $\grad^2\Psi(P)[H,H]=\Tr(P^{-1}HP^{-1}H)$, which implies $\tfrac{1}{b^2}\norm{H}_F^2\leq\grad^2\Psi(P)[H,H]\leq \tfrac{1}{a^2}\norm{H}_F^2$.

The values of $a>0$ and $b<+\infty$ in the definition of $\tilde{\cQ}\triangleq \{P\in\mathbb{S}^{n}:\ a\mb{I}\preceq P \preceq b\mb{I}\}$ for the other cases, i.e., $(a^*=0, b^*<+\infty)$ and $(a^*>0, b^*=+\infty)$ are given in Figure~\ref{alg:admm}; these bounds can also be proven very similarly; thus, their proofs are omitted.

\subsubsection{Proof of Lemma~\ref{lem:prox_psi}}
\label{sec:sup_proof_prox_psi_lem}
The $ \prox{\Psi/\rho}$ map can be equivalently written as 
\begin{small}
\begin{equation}
\label{eq:psi_prox_equiv}
\prox{\Psi/\rho}(\bar{P})=\argmin_{P\in\mathbb{S}^{n}}\{-\log\det(P)+\tfrac{\rho}{2}\norm{P-(\bar{P}-\tfrac{1}{\rho}S)}_F^2:\ a\mb{I}\preceq P \preceq b\mb{I}\}. 
\end{equation}
\end{small}

\noindent Let $U\diag(\bar{\lambda})U^\top$ be the eigen-decomposition of $\bar{P}-\tfrac{1}{\rho}S$. Fixing $U\in\mathbb{S}^{n}$, and by restricting the variable $P\in\mathbb{S}^{n}$ in \eqref{eq:psi_prox_equiv} to have the form $U\diag(\lambda)U^\top$ for some $\lambda\in\reals^{n}$, we obtain the optimization problem \eqref{eq:restricted_problem} over the variable $\lambda\in\reals^{n}$: 
\begin{small}
\begin{equation}
\label{eq:restricted_problem}
\min_{\lambda\in\reals^{n}}\left\{-\sum_{i=1}^{n}\log(\lambda_i)+\tfrac{\rho}{2}(\lambda_i-\bar{\lambda}_i)^2:\ a\leq\lambda_i\leq b,~ i=1,\ldots,n\right\}. 
\end{equation}
\end{small}

\noindent For a given $\bar{t}\in\reals$, and $a,b,\gamma>0$, the unique minimizer of $\min_{t\in\reals}\{-\log(t)+\tfrac{\rho}{2}|t-\bar{t}|^2:\ a\leq t\leq b\}$ can be written as $\max\left\{\min\left\{\frac{\bar{t}+\sqrt{\bar{t}^2+4/\rho}}{2},\ b\right\},\ a\right\}$. Hence, $\lambda^*\in\reals^{n}$ given in \eqref{eq:opt_singular} is the unique minimizer of \eqref{eq:restricted_problem}. Let $h:\reals^{n}\rightarrow\reals\cup\{+\infty\}$ be defined as $h(\lambda)\triangleq -\sum_{i=1}^{n}\log(\lambda_i)+\mathbf{1}_{\cH}(\lambda)$, where $\cH\triangleq \{\lambda\in\reals^{n}:\ a\mathbf{1}\leq\lambda\leq b\mathbf{1}\}$ 
and $\lambda^*=\argmin_{\lambda\in\reals^{n}}\{h(\lambda)+\tfrac{\rho}{2}\norm{\lambda-\bar{\lambda}}_2^2\}$. From the first-order optimality conditions, it follows that $\bar{\lambda}-\lambda^*\in\tfrac{1}{\rho}~\partial h(\lambda)|_{\lambda=\lambda^*}$.

Let $H:\mathbb{S}^{n}\rightarrow\reals\cup\{+\infty\}$ be such that $H(P)=-\log\det(P)+\mathbf{1}_{\cQ}(P)$. Definition of $\prox{\Psi/\rho}(\bar{P})$ implies that $\left(\bar{P}-\tfrac{1}{\rho}S-\prox{\Psi/\rho}(\bar{P})\right)\in\tfrac{1}{\rho}\partial H(P)|_{P=\prox{\Psi/\rho}(\bar{P})}$. In the rest of the proof, $\sigma:\mathbb{S}^{n}\rightarrow\reals^{n}$ denotes the function that returns the singular values of its argument. Note that $H(P)=h(\sigma(P))$ for all $P\in\mathbb{S}^{n}$. Since $h$ is \emph{absolutely symmetric}, Corollary~2.5 in~\cite{lewis-convex-unitarily} implies that $P^{\rm prox}=\prox{\Psi/\rho}(\bar{P})$ if and only if $\sigma(\bar{P}-\tfrac{1}{\rho}S-P^{\rm prox})\in\tfrac{1}{\rho}~\partial h(\lambda)|_{\lambda=\sigma(P^{\rm prox})}$ and there exists a simultaneous singular value decomposition of the form $P^{\rm prox}=U\diag(\sigma(P^{\rm prox}))U^\top$ and $\bar{P}-\tfrac{1}{\rho}S-P^{\rm prox}=U\diag\left(\sigma\left(\bar{P}-\tfrac{1}{\rho}S_k-P^{\rm prox}\right)\right)U^\top$. Hence, $\prox{\Psi/\rho}(\bar{P})=U\diag(\lambda^*)U^\top$  follows from 
$\bar{\lambda}-\lambda^*\in\tfrac{1}{\rho}~\partial h(\lambda)|_{\lambda=\lambda^*}$.
\vspace*{-4mm}
\subsubsection{Proof of Lemma~\ref{lem:prox_phi}}
\label{sec:sup_proof_prox_phi_lem}
From the definition of $\prox{\Phi/\rho}$, we have \vspace*{-4mm}
{\small
\begin{equation}
\label{eq:phi_prox_separable}
\prox{\Phi/\rho}=\argmin_{P\in\mathbb{S}^{n}}\Big\{\sum_{(i,j)\in\cI\times\cI}\tfrac{\alpha}{\rho}G_{ij}|P_{ij}|+\tfrac{1}{2}|P_{ij}-\bar{P}_{ij}|^2:\ \diag(P)\geq \mathbf{0}\Big\}. \vspace*{-2mm}
\end{equation}}%
For a given $\bar{t}\in\reals$, and $\gamma>0$, the unique minimizer of $\min_{t\in\reals}\gamma|t|+\tfrac{1}{2}|t-\bar{t}|^2$ can be written as $\sgn(\bar{t})\max\{|\bar{t}|-\gamma,~0\}$; and the minimizer of $\min_{t\in\reals}\{\gamma t+\tfrac{1}{2}|t-\bar{t}|^2:\ t\geq 0\}$ can be written as $\max\{\bar{t}-\gamma,~0\}$. Hence, \eqref{eq:phi_prox} follows from the separability of the objective in \eqref{eq:phi_prox_separable}. 

\subsection{Additional Numerical Results}
In this section, the importance of STAGE-I in the SPS algorithm is shown numerically. Notice that the STAGE-II can be directly implemented for the sample covariance matrix $S$. This could be interpreted as direct estimation of the covariance function parameters by fitting the covariogram \citep{CressieBook}. For this purpose, the covariance function parameters are estimated from $N$ realizations of a zero-mean GRF simulated over $n=100$ randomly selected locations over a square domain $\cX=[0,\beta]\times[0,\beta]$ with a Matern covariance function with smoothness parameter $3/2$ and the parameter vector $\th^*=[{\th^*_\rho}^\top,\theta^*_v,\theta^*_0]^\top=[15,8,1]^\top$. The SPS and covariogram methods are then compared based on $R=100$ simulation replications (every time $n=100$ locations are randomly resampled). Table~\ref{tbl:SPSvsCorMeanAll} shows the mean and standard error of the parameter estimates, respectively. With increasing $N$, we see faster convergence of the SPS parameter estimates to their true values. Compared to the covariogram method, the SPS mean parameter estimates are almost always closer to their true parameter values and their standard errors are lower. The importance of  STAGE-I in the SPS algorithm is more evident when $N\ll N_0$. STAGE-I zooms into the region in the parameter space of the nonconvex objective function where the global minimum lies, and this results in better covariance parameter estimates. As expected, increasing the domain size ($\beta$) results in a lower point density in the domain, and this deteriorates the performance of both methods. 

\begin{landscape}
\begin{table}
    \caption{Mean (standard deviation) of parameter estimates from $R=100$ replications when the process is sampled $N$ times at each of $n=100$  randomly chosen locations in the domain $\cX=[0,\beta]\times[0,\beta]$}
    \centering
    \tiny
    \begin{tabular}[t]{rrrrr|rrrrr}
    \toprule
     & \multicolumn{4}{c}{SPS} & \multicolumn{4}{c}{Covariogram}\\
    \cmidrule(r){1-1} \cmidrule(r){2-5} \cmidrule(r){6-9}
    $\beta\setminus N$ & \multicolumn{1}{c}{\textbf{1}} & \multicolumn{1}{c}{\textbf{5}} & \multicolumn{1}{c}{\textbf{20}} & \multicolumn{1}{c}{\textbf{40}} & \multicolumn{1}{c}{\textbf{1}} & \multicolumn{1}{c}{\textbf{5}} & \multicolumn{1}{c}{\textbf{20}} & \multicolumn{1}{c}{\textbf{40}} \\
    \hline
    \multirow{3}{*}{25}
    &\textbf{10.22}~(0.08)&\textbf{12.81}~(0.71)&\textbf{13.95}~(0.34)&\textbf{14.81}~(0.25)&\textbf{4.14}~(1.15)&\textbf{14.03}~(0.73)&\textbf{14.15}~(0.33)&\textbf{14.63}~(0.25)\\
    &\textbf{8.03}~(0.27)&\textbf{8.10}~(0.31)&\textbf{7.87}~(0.16)&\textbf{8.13}~(0.13)&\textbf{4.01}~(0.74)&\textbf{7.22}~(0.34)&\textbf{7.56}~(0.16)&\textbf{7.99}~(0.13)\\
    &\textbf{0.83}~(0.07)&\textbf{0.92}~(0.08)&\textbf{0.91}~(0.05)&\textbf{0.92}~(0.03)&\textbf{1.05}~(0.09)&\textbf{1.62}~(0.07)&\textbf{1.38}~(0.04)&\textbf{1.09}~(0.03)\\
    \hline
    \multirow{3}{*}{50}
    &\textbf{11.84}~(0.14)&\textbf{14.30}~(0.71)&\textbf{14.64}~(0.37)&\textbf{15.19}~(0.25)&\textbf{7.01}~(2.04)&\textbf{14.39}~(0.83)&\textbf{15.63}~(0.39)&\textbf{15.17}~(0.26)\\
    &\textbf{7.93}~(0.31)&\textbf{7.93}~(0.21)&\textbf{8.05}~(0.13)&\textbf{8.02}~(0.09)&\textbf{6.07}~(0.45)&\textbf{7.57}~(0.23)&\textbf{7.92}~(0.13)&\textbf{7.94}~(0.09)\\
    &\textbf{1.13}~(0.08)&\textbf{1.14}~(0.11)&\textbf{1.04}~(0.07)&\textbf{1.05}~(0.05)&\textbf{1.02}~(0.13)&\textbf{1.67}~(0.10)&\textbf{1.54}~(0.07)&\textbf{1.22}~(0.05)\\
    \hline
    \multirow{3}{*}{75}
    &\textbf{13.73}~(2.14)&\textbf{15.28}~(1.06)&\textbf{15.37}~(0.34)&\textbf{15.11}~(0.21)&\textbf{21.76}~(2.50)&\textbf{17.65}~(1.19)&\textbf{15.63}~(0.35)&\textbf{15.22}~(0.21)\\
    &\textbf{7.39}~(0.41)&\textbf{7.72}~(0.20)&\textbf{7.88}~(0.10)&\textbf{7.96}~(0.07)&\textbf{7.77}~(0.42)&\textbf{7.69}~(0.20)&\textbf{7.97}~(0.10)&\textbf{8.04}~(0.07)\\
    &\textbf{1.27}~(0.14)&\textbf{1.64}~(0.12)&\textbf{1.64}~(0.08)&\textbf{1.20}~(0.05)&\textbf{1.37}~(0.16)&\textbf{1.29}~(0.13)&\textbf{1.07}~(0.08)&\textbf{0.98}~(0.05)\\
    \hline
    \multirow{3}{*}{100}
    &\textbf{19.82}~(3.36)&\textbf{14.87}~(0.45)&\textbf{15.62}~(0.27)&\textbf{15.12}~(0.14)&\textbf{31.56}~(3.70)&\textbf{16.57}~(0.58)&\textbf{15.91}~(0.29)&\textbf{15.11}~(0.15)\\
    &\textbf{7.18}~(0.32)&\textbf{8.21}~(0.17)&\textbf{7.77}~(0.09)&\textbf{7.90}~(0.06)&\textbf{6.88}~(0.34)&\textbf{8.09}~(0.17)&\textbf{7.87}~(0.09)&\textbf{8.01}~(0.06)\\
    &\textbf{1.38}~(0.18)&\textbf{1.57}~(0.11)&\textbf{1.76}~(0.07)&\textbf{1.30}~(0.04)&\textbf{1.91}~(0.21)&\textbf{1.22}~(0.11)&\textbf{1.14}~(0.07)&\textbf{1.01}~(0.04)\\
    \hline
    \multirow{3}{*}{125}&
    \textbf{25.28}~(4.23)&\textbf{15.27}~(0.76)&\textbf{15.74}~(0.27)&\textbf{15.03}~(0.15)&\textbf{38.68}~(4.50)&\textbf{16.63}~(0.67)&\textbf{15.88}~(0.26)&\textbf{15.01}~(0.15)\\
    &\textbf{7.70}~(0.38)&\textbf{7.97}~(0.18)&\textbf{7.67}~(0.10)&\textbf{7.95}~(0.07)&\textbf{7.77}~(0.38)&\textbf{7.88}~(0.18)&\textbf{7.79}~(0.10)&\textbf{8.09}~(0.07)\\
    &\textbf{1.90}~(0.24)&\textbf{1.73}~(0.13)&\textbf{1.90}~(0.08)&\textbf{1.28}~(0.05)&\textbf{1.77}~(0.24)&\textbf{1.25}~(0.12)&\textbf{1.25}~(0.08)&\textbf{0.95}~(0.05)\\
    \hline
    \multirow{3}{*}{150}
    &\textbf{41.45}~(5.74)&\textbf{17.18}~(0.73)&\textbf{15.26}~(0.17)&\textbf{15.49}~(0.14)&\textbf{46.96}~(7.52)&\textbf{15.82}~(0.54)&\textbf{15.15}~(0.16)&\textbf{15.57}~(0.14)\\
    &\textbf{6.78}~(0.35)&\textbf{7.75}~(0.19)&\textbf{7.85}~(0.08)&\textbf{7.87}~(0.06)&\textbf{6.77}~(0.39)&\textbf{7.83}~(0.19)&\textbf{7.71}~(0.08)&\textbf{7.71}~(0.06)\\
    &\textbf{2.42}~(0.27)&\textbf{1.39}~(0.13)&\textbf{1.09}~(0.06)&\textbf{1.13}~(0.06)&\textbf{2.57}~(0.30)&\textbf{1.91}~(0.13)&\textbf{1.77}~(0.06)&\textbf{1.50}~(0.05)\\
    \hline
    \multirow{3}{*}{175}
    &\textbf{47.10}~(7.40)&\textbf{16.32}~(0.70)&\textbf{15.22}~(0.16)&\textbf{14.85}~(0.10)&\textbf{79.66}~(10.52)&\textbf{15.52}~(0.67)&\textbf{15.16}~(0.16)&\textbf{14.92}~(0.10)\\
    &\textbf{6.86}~(0.38)&\textbf{7.76}~(0.17)&\textbf{8.07}~(0.08)&\textbf{8.06}~(0.06)&\textbf{5.87}~(0.46)&\textbf{7.80}~(0.17)&\textbf{7.93}~(0.8)&\textbf{7.88}~(0.07)\\
    &\textbf{2.36}~(0.27)&\textbf{1.28}~(0.13)&\textbf{1.01}~(0.06)&\textbf{0.95}~(0.05)&\textbf{3.67}~(0.35)&\textbf{1.86}~(0.13)&\textbf{1.69}~(0.07)&\textbf{1.35}~(0.05)\\
    \hline
    \multirow{3}{*}{200}
    &\textbf{64.18}~(9.77)&\textbf{17.69}~(2.01)&\textbf{15.10}~(0.15)&\textbf{15.02}~(0.10)&\textbf{86.13}~(11.73)&\textbf{17.68}~(2.70)&\textbf{15.04}~(0.14)&\textbf{15.11}~(0.10)\\
    &\textbf{6.12}~(0.38)&\textbf{7.76}~(0.17)&\textbf{7.90}~(0.08)&\textbf{7.99}~(0.06)&\textbf{5.50}~(0.42)&\textbf{7.82}~(0.18)&\textbf{7.75}~(0.08)&\textbf{7.80}~(0.06)\\
    &\textbf{3.01}~(0.29)&\textbf{1.28}~(0.13)&\textbf{1.05}~(0.06)&\textbf{1.00}~(0.05)&\textbf{4.07}~(0.35)&\textbf{1.88}~(0.14)&\textbf{1.75}~(0.06)&\textbf{1.42}~(0.05)\\
    \hline
    \multirow{3}{*}{250}
    &\textbf{101.44}~(13.90)&\textbf{22.33}~(4.78)&\textbf{14.97}~(0.14)&\textbf{14.92}~(0.09)&\textbf{144.40}~(16.46)&\textbf{18.42}~(3.42)&\textbf{14.88}~(0.14)&\textbf{14.99}~(0.09)\\
    &\textbf{5.46}~(0.40)&\textbf{7.99}~(0.21)&\textbf{8.10}~(0.09)&\textbf{8.17}~(0.06)&\textbf{4.73}~(0.43)&\textbf{8.17}~(0.20)&\textbf{7.97}~(0.09)&\textbf{7.97}~(0.06)\\
    &\textbf{3.47}~(0.33)&\textbf{1.09}~(0.15)&\textbf{0.95}~(0.07)&\textbf{0.89}~(0.05)&\textbf{4.71}~(0.37)&\textbf{1.56}~(0.16)&\textbf{1.63}~(0.07)&\textbf{1.33}~(0.05)\\
    \hline
    \multirow{3}{*}{300}
    &\textbf{172.48}~(9.07)&\textbf{35.91}~(8.96)&\textbf{15.14}~(0.14)&\textbf{15.15}~(0.12)&\textbf{266.24}~(19.87)&\textbf{39.57}~(9.77)&\textbf{15.11}~(0.15)&\textbf{15.22}~(0.12)\\
    &\textbf{4.50}~(0.44)&\textbf{7.51}~(0.22)&\textbf{8.01}~(0.08)&\textbf{7.97}~(0.07)&\textbf{2.92}~(0.41)&\textbf{7.52}~(0.25)&\textbf{7.86}~(0.09)&\textbf{7.77}~(0.07)\\
    &\textbf{4.78}~(0.39)&\textbf{1.47}~(0.21)&\textbf{1.00}~(0.07)&\textbf{1.04}~(0.06)&\textbf{6.89}~(0.38)&\textbf{2.13}~(0.24)&\textbf{1.69}~(0.08)&\textbf{1.49}~(0.06)\\
    \hline
    \multirow{3}{*}{400}
    &\textbf{319.69}~(22.19)&\textbf{109.62}~(20.91)&\textbf{26.19}~(5.51)&\textbf{15.03}~(0.12)&\textbf{451.89}~(26.09)&\textbf{119.02}~(20.11)&\textbf{20.55}~(7.75)&\textbf{15.06}~(0.11)\\
    &\textbf{2.62}~(0.30)&\textbf{6.29}~(0.36)&\textbf{7.87}~(0.14)&\textbf{8.01}~(0.08)&\textbf{1.36}~(0.36)&\textbf{6.23}~(0.34)&\textbf{7.86}~(0.15)&\textbf{7.83}~(0.07)\\
    &\textbf{6.18}~(0.28)&\textbf{2.76}~(0.35)&\textbf{1.18}~(0.12)&\textbf{1.03}~(0.07)&\textbf{8.09}~(0.32)&\textbf{3.51}~(0.32)&\textbf{1.74}~(0.14)&\textbf{1.47}~(0.06)\\
    \hline
    \multirow{3}{*}{500}
    &\textbf{483.79}~(19.57)&\textbf{227.05}~(33.13)&\textbf{36.48}~(13.62)&\textbf{15.05}~(0.12)&\textbf{645.88}~(30.76)&\textbf{286.89}~(31.08)&\textbf{43.31}~(11.85)&\textbf{15.06}~(0.12)\\
    &\textbf{1.81}~(0.24)&\textbf{5.15}~(0.42)&\textbf{7.48}~(0.20)&\textbf{8.02}~(0.08)&\textbf{0.66}~(0.32)&\textbf{4.71}~(0.39)&\textbf{7.33}~(0.18)&\textbf{7.85}~(0.08)\\
    &\textbf{7.19}~(0.25)&\textbf{4.03}~(0.41)&\textbf{1.49}~(0.19)&\textbf{0.97}~(0.08)&\textbf{9.00}~(0.32)&\textbf{5.15}~(0.39)&\textbf{2.20}~(0.17)&\textbf{1.41}~(0.07)\\
    \bottomrule
    \end{tabular}
    \label{tbl:SPSvsCorMeanAll}
\end{table}
\end{landscape}

\end{document}